%% file: main.tex
\pdfoutput=1

\documentclass[11pt]{article}

\usepackage[preprint]{acl}

\usepackage{times}
\usepackage{latexsym}
\usepackage[T1]{fontenc}
\usepackage[utf8]{inputenc}
\usepackage{microtype}
\usepackage{inconsolata}
\usepackage{graphicx}
\usepackage{adjustbox}
\usepackage{array}
\usepackage{amsmath}
\usepackage{color}
\usepackage[table]{xcolor}
\usepackage{xspace}
\usepackage{cleveref}
\usepackage{enumitem}
\usepackage{multirow}
\usepackage{subcaption}
\usepackage{booktabs}
\usepackage{pifont}
\usepackage{url}
\usepackage{wrapfig}
\usepackage{ragged2e}
\usepackage{multicol}
\usepackage{arydshln}
\usepackage[most]{tcolorbox}

\definecolor{pastelred}{RGB}{255,179,179}
\definecolor{pastelblue}{RGB}{179,209,255}

\title{LLMs Anchor on Chief Complaint and Fail to Integrate Evidence in Sequential Clinical Triage}

\author{Dipankar Srirag\textsuperscript{1}\quad Haokai Zhao\textsuperscript{1}\quad Ashutosh Kumar\textsuperscript{2}\\
\textbf{Eleanor Hopper}\textsuperscript{3,4}\quad\textbf{Michael Dalton}\textsuperscript{3}\quad\textbf{Quoc Dung Nguyen\textsuperscript{1,5}}\\
\textbf{Aditya Joshi\textsuperscript{1}\quad Salil S. Kanhere\textsuperscript{1}\quad Padmanesan Narasimhan\textsuperscript{1}}\\
\textsuperscript{1}University of New South Wales, Sydney\\
  \textsuperscript{2}Independent Researcher\quad
  \textsuperscript{3}St Vincent's Hospital, Sydney\\
  \textsuperscript{4}Royal Prince Alfred Hospital, Sydney\quad
  \textsuperscript{5}Campbelltown Hospital, Sydney\\
\href{mailto:d.srirag@unsw.edu.au}{\texttt{d.srirag@unsw.edu.au}}
}

\begin{document}
\maketitle
\input{sections/00-abstract}
\input{sections/01-introduction}
\input{sections/03-evaluation-methodology}
\input{sections/04-experiments}

\input{sections/05-results}
\input{sections/02-related-works}
\input{sections/06-discussion}

\input{sections/07-limitations}

\bibliography{references}

\appendix
\input{appendix/A-esi}
\input{appendix/B-prompts}

\input{appendix/D-numerical-results}
\input{appendix/E-clinician-corp}
\input{appendix/C-annot-exercise}

\end{document}

%% file: sections/00-abstract.tex
\begin{abstract}
Triage in the emergency department (ED) is a sequential decision process that unfolds turn by turn. Existing evaluations of large language models (LLMs) for triage use completed retrospective records and report performance close to that of physicians. We implement a methodology for evaluating LLMs on \textit{sequential triage}, the task of predicting a triage acuity label from a growing prefix of a nurse-patient conversation. We evaluate six LLMs at five sequential checkpoints on two corpora: 425 LLM-generated and 50 clinician-authored conversations, both labelled under the Emergency Severity Index (ESI). Every model, measured by quadratic weighted kappa (QWK), degrades from moderate-to-substantial agreement on completed records to fair-to-moderate agreement at every sequential checkpoint. Controlled perturbations show that the label at every checkpoint is \textbf{anchored on the chief-complaint} exchanges, and prompting interventions fail to lift this plateau. Models extract clinically relevant content from later turns, yet the surprisal of the true label rises across the checkpoints. So the model \textbf{fails to integrate the evidence}. Three expert clinicians on the same conversations reach a QWK of 0.887--0.929, while the best model reaches 0.295. Predictions concentrate at mid-acuity labels, and models agree with each other more than with the ground truth. Deploying LLMs for ED triage based on offline benchmarks alone misses this sequential failure.
\end{abstract}


%% file: sections/01-introduction.tex
\section{Introduction}
\label{sec:intro}

Triage is the first phase in a patient's journey through the emergency department (ED) of a hospital. Triage in the real world proceeds as a conversation: the triage nurse elicits the chief complaint (CC), gathers vitals and history during a brief patient interview. Finally, the nurse records their notes and assigns an ordinal acuity level that reflects the urgency of the presentation. Therefore, triage is sequential (by virtue of conversation) and time-constrained (depending on the criticality of the patient's health condition). Under the Emergency Severity Index (ESI; \citealp{esi-handbook}), the acuity level ranges from 1 (highest urgency) to 5 (lowest). Following the prescribed algorithm for ESI, CC is elicited during the intial phase of the interaction, while post-CC utterances help narrow the working acuity assessment through vitals and history. {Detailed description of the ESI algorithm is provided in Appendix~\ref{sec:a-esi-algorithm}.}

Existing evaluations of large language models (LLMs) on clinical triage operate retrospectively on records completed by triage nurses \citep{Masanneck2024, williams2024llm, lu-etal-2024-triageagent, Brodeur2026}. These records are \textit{an outcome} of the interaction, and not reflective of the evidence available to the nurse \textit{during} the interaction \citep{srirag2026triagedischargesurveynlp}. This distinction is important, and evaluation of LLM-based triage systems should reflect the sequential nature of evidence elicitation during triage. We define \textit{sequential clinical triage} as predicting an acuity label from a growing prefix of the nurse-patient conversation. We evaluate 6 LLMs on sequential clinical triage across two settings: (\textsc{offline} where the full structured EHR is available, and \textsc{sequential} which utilises incremental conversational evidence based on conversation prefixes at five checkpoints starting with the chief-complaint exchange). We evaluate the sequential setting on 425 \textsc{simulated} (LLM-generated) conversations in addition to 50 \textsc{clinician}-authored conversations.

\begin{figure*}[t!]
    \centering
    \includegraphics[width=\linewidth]{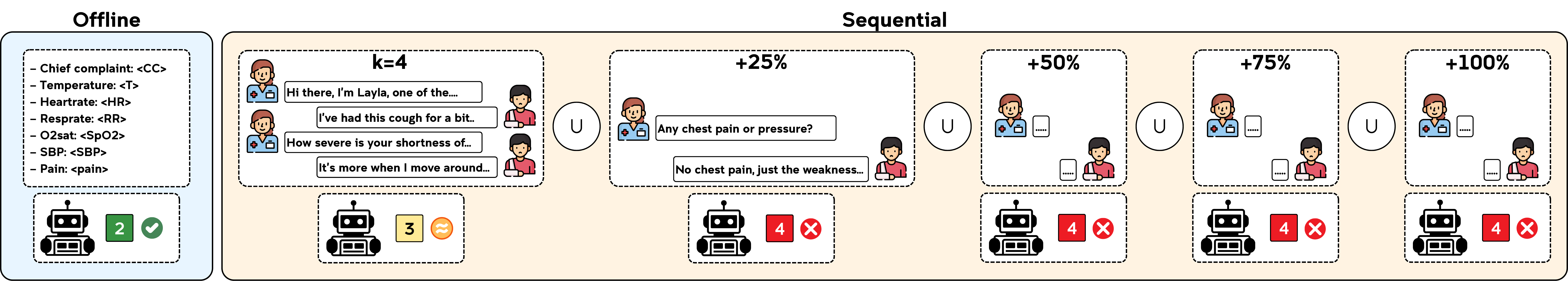}
    \caption{Input format for the two information settings: \textsc{offline} and \textsc{sequential}. The \textsc{offline} input serialises the structured EHR record into text. The \textsc{sequential} input consists of conversation prefixes at five checkpoints indexed by \textit{k}, with utterances accumulating at subsequent checkpoints ($\bigcup$). We evaluate the model at every checkpoint.}
    \label{fig:framework}
\end{figure*}

Through our evaluation, we aim to answer four research questions: (1) How well do models perform sequential clinical triage?; (2) Where in the conversation does the model commit its triage decision (\textit{i.e., the model determines the acuity level})?; (3) How does a model's chosen commitment to an acuity level compare to a clinician's under the same evidence?; (4) Why does more evidence not improve sequential clinical triage?

Our results show that every model degrades from moderate-to-substantial offline agreement with the ground truth to fair-to-moderate agreement at every sequential checkpoint on both corpora. Additionally, we perform perturbations to input conversations, such as reordering and removing utterances. We find that such perturbations to the chief-complaint utterances alter the label at every checkpoint, while perturbations to later turns do not. This localises the predictive signal to the chief-complaint exchange. We also compare model behaviour with three expert clinicians on the same 50 conversations from \textsc{simulated}, where clinicians wait for later turns and reach QWK 0.887--0.929 while the best model reaches 0.295. Finally, we show that models extract the clinically relevant content from the transcript, yet the posterior on the true label moves away from the ground truth as more evidence arrives. \textbf{The model fails to integrate the evidence.}


%% file: sections/03-evaluation-methodology.tex
\section{Evaluation Methodology}
\label{sec:methodology}
Our objective is to evaluate automatic triage under two settings. The \textsc{offline} setting provides the model with the complete structured EHR at once, and the model is tasked with predicting a single ESI label. In contrast, the \textsc{sequential} setting provides successive prefixes of a nurse-patient conversation grounded on the same EHR record. In this case, the model must predict a label at each prefix. The two settings differ in how much of the case information the model sees at inference. 

\subsection{Task Formulation}

We formalise automatic triage as a five-class ordinal classification problem where the classes are acuity levels as prescribed in the Emergency Severity Index (ESI). Given an input $\mathcal{X}$, a model $f(\cdot; \theta)$ predicts an acuity label $\mathcal{Y} \in \{1 \prec 2 \prec 3 \prec 4 \prec 5\}$, where $\mathcal{Y}=1$ denotes the highest clinical urgency and $\mathcal{Y}=5$ the least. The information in $\mathcal{X}$ distinguishes the two information settings we study.
\begin{table}[t!]
\begin{adjustbox}{width=\linewidth,center}
    \begin{tabular}{lccccccccc}
    \toprule
    \multirow{2}{*}{Subset} & \multirow{2}{*}{Total} & \multicolumn{5}{c}{ESI} & \multicolumn{2}{c}{Turns} \\
    \cmidrule(lr){3-7} \cmidrule(lr){8-9}
     & & 1 & 2 & 3 & 4 & 5 & Mean & Range \\
    \midrule
    \textsc{simulated}       & 425 & 85 & 87 & 89 & 85 & 79 & 19.9 & 8--24  \\\hdashline
    \quad\textit{TriageSim}  & 375 & 75 & 75 & 75 & 75 & 75 & 20.2 & 8--22  \\
    \quad\textit{EHR2Dial}   &  50 & 10 & 12 & 14 & 10 &  4 & 18.0 & 13--24 \\
    \midrule
    \textsc{clinician}       &  50 &  8 & 10 & 12 & 10 & 10 & 24.8 & 16--22 \\
    \bottomrule
    \end{tabular}
\end{adjustbox}
\caption{Dataset statistics for both corpora. The \textsc{simulated} corpus pools TriageSim and EHR2Dial. ESI columns give conversation counts per acuity class. \textit{Turns} reports the average number of utterances and their range.}
\label{tab:diag-stat}
\end{table}
\subsection{Information Settings}

\noindent\textbf{\textsc{offline}.} This represents the conventional setting for prior work on automatic triage with LLMs. $\mathcal{X}$ is the complete structured EHR record, converted to text (chief complaint, presenting symptoms, vital signs, and other fields carried in the record). This implies that the model has access to all recorded evidence at once and generates a single ESI label: $f(\mathcal{X}; \theta) \rightarrow \mathcal{Y}$.

\noindent\textbf{\textsc{sequential}.} $\mathcal{X}$ is a prefix of a nurse-patient conversation. Each conversation is a sequence of utterances $(u_1, u_2, \ldots, u_T)$ in which $u_1$ is a nurse utterance and subsequent turns alternate between nurse and patient, interleaved with vital-sign observations. We evaluate the model at five checkpoints indexed by $k$, each corresponding to a longer prefix $d_k = (u_1, \ldots, u_k)$: $f(d_k; \theta) \rightarrow \mathcal{Y}_k$. The first checkpoint is fixed at $k=4$ turns, which covers the CC utterances in our corpora. The remaining four are placed at \{25\%, 50\%, 75\%, 100\%\} of the turns beyond that. We evaluate LLMs for \textit{sequential clinical triage} at every checkpoint to derive a trajectory. Figure~\ref{fig:framework} illustrates an example input and output for both settings.

%% file: sections/04-experiments.tex
\section{Experiment Setup}
\label{sec:setup}


For the \textsc{offline} setting, we use a subset of 425 EHR records from MIMIC-IV-ED~\citep{Johnson2023}, spanning all five ESI levels.

The \textsc{sequential} setting requires nurse-patient conversations grounded on the EHR records in the offline setting. Large-scale, real triage conversations are unavailable since they are protected by patient privacy regulations. Therefore, for the sequential setting, we experiment with two corpora (statistics in Table~\ref{tab:diag-stat}):

\begin{enumerate}
    \item We simulate nurse-patient conversations from a seed EHR case with a known ESI label. The simulations are created using two frameworks: (a) TriageSim~\citep{TriageSim} and (b) EHR2Dial~\citep{zhu2026elicitedehrgroundedlongitudinalinteractive}. We pool the two into an evaluation set of 425 synthetic conversations, the \textsc{simulated} corpus. 
  \item  We also curate 50 conversations authored by two expert clinicians with more than 20 years of combined experience working in an ED, forming the \textsc{clinician} corpus. This corpus is a human-authored counterpart to \textsc{simulated} and is used to verify that findings are not isolated to LLM-generated conversations. 
\end{enumerate}

\subsection{Models}
\label{sec:models}

We evaluate \textit{six} LLMs. \textit{Four} are open-weight: Gemma-4-E4B (\textsc{gemma-s}; \citealp{gemmateam2026gemma4technicalreport}), Gemma-4-31B (\textsc{gemma-l}), Qwen-3.5-9B (\textsc{qwen-s}; \citealp{qwenteam2026qwen35omnitechnicalreport}), and Qwen-3.5-27B (\textsc{qwen-l}). The remaining \textit{two} are closed-weight and accessed via public APIs: Claude Opus 4.6 (\textsc{claude}) and GPT-5.5 (\textsc{gpt}). We prompt each model to produce an ESI label under greedy decoding, with extended-reasoning modes disabled to isolate prompt-level interventions from native model reasoning (\textsc{vanilla}).

\noindent\textbf{Other prompting strategies.} For each model, we additionally compare three strategies: reasoning-native decoding with the model's native chain-of-thought mode enabled (\textsc{r-on}), Plan-and-Solve prompting (\textsc{ps+}; \citealp{wang2023plansolve}), and self-consistency with majority voting from five samples at $t=0.7$ and top-$p = 0.95$ (\textsc{sc}; \citealp{wang2023selfconsistency}). \textsc{r-on} and \textsc{ps+} also use greedy decoding, which is similar to \textsc{vanilla}. All prompts used during our evaluation are provided in Appendix~\ref{app:prompts}.

\begin{table*}[ht!]
\begin{adjustbox}{width=0.7\linewidth, center}
    \begin{tabular}{lcccccc}
    \toprule
    \textbf{Model} & \textsc{claude} & \textsc{gpt} & \textsc{gemma-s} & \textsc{gemma-l} & \textsc{qwen-s} & \textsc{qwen-l}   \\\midrule
    \textbf{QWK} & 0.701 & 0.592 & 0.612 & 0.629 & 0.604 & 0.638\\
    \bottomrule
    \end{tabular}
\end{adjustbox}
\caption{Performance comparison of model under \textsc{offline} setting and \textsc{vanilla} prompting on the \textsc{simulated} corpus.}
\label{tab:offline-qwk}
\end{table*}

\begin{figure*}[ht!]
    \centering
    \includegraphics[width=\linewidth]{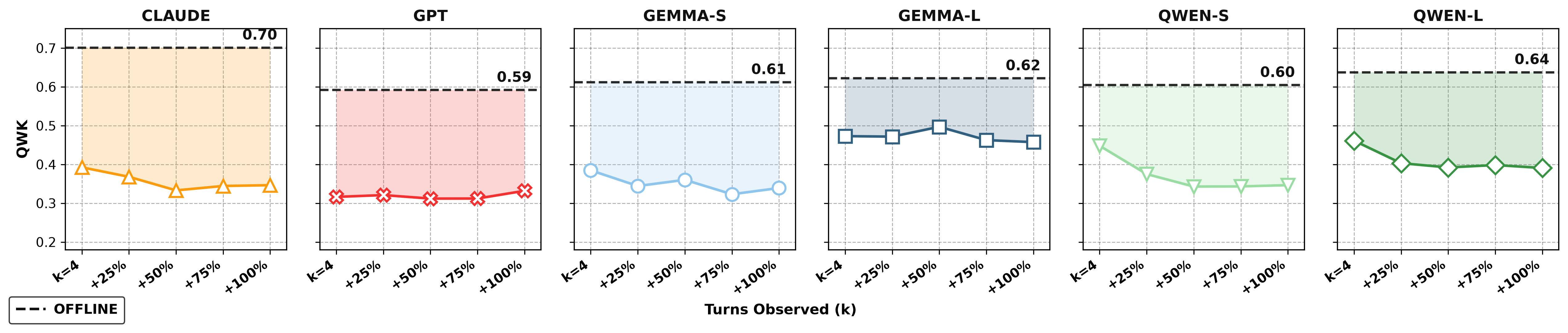}
    \caption{Performance comparison of all models across the five sequential checkpoints on the \textsc{simulated} corpus. Dashed line per panel represents the \textsc{offline} skyline.}
    \label{fig:rq1-sim}
\end{figure*}

\subsection{Metrics}
\label{sec:metrics}

The primary metric is quadratic weighted Cohen's Kappa (QWK; \citealp{cohen1968weighted}), which captures the ordinal structure of ESI: confusing ESI 1 with ESI 5 is penalised more heavily than confusing ESI 1 with ESI 2. By construction, the evaluation set is approximately balanced across ESI, so the metric's class-imbalance sensitivity does not distort the reported scores. We compute confidence intervals via paired bootstrap with 10,000 resamples (See Appendix~\ref{app:ci}).

%% file: sections/05-results.tex
\section{Results}
\label{sec:results}

We organise the results around \textit{four} questions: performance on sequential clinical triage (Section~\ref{sec:rq1}), localising the predictive signal (Section~\ref{sec:rq2}), model performance compared with expert clinicians (Section~\ref{sec:rq3}), and model indifference to additional evidence (Section~\ref{sec:rq4}).

\subsection{How well do models perform sequential clinical triage?}
\label{sec:rq1}

\begin{figure*}[t!]
\centering
\includegraphics[width=\linewidth]{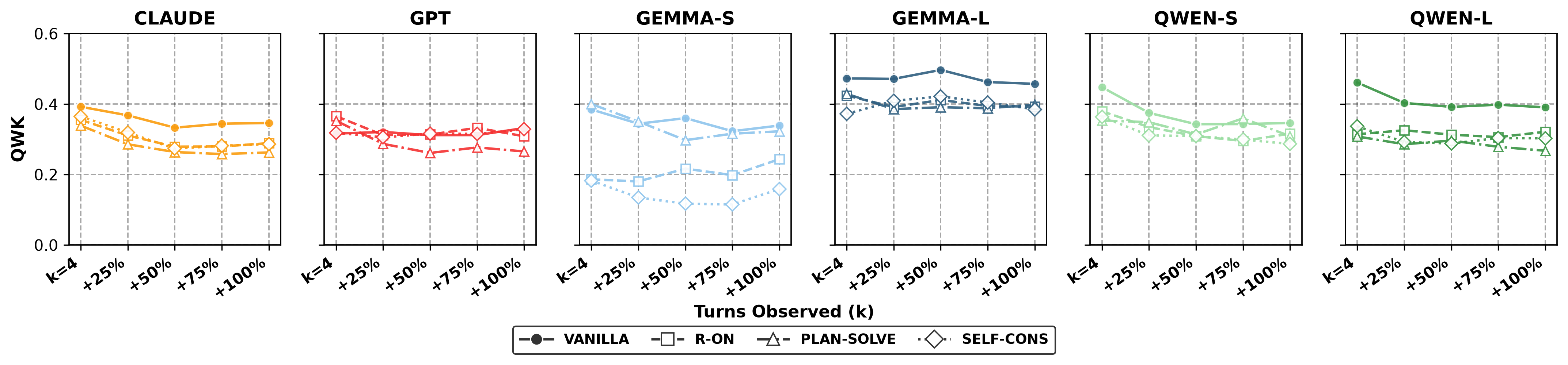}
\caption{Performance comparison of all models across the five sequential checkpoints on the \textsc{simulated} corpus under four prompting strategies.}
\label{fig:rq1-reason}
\end{figure*}

\noindent\textbf{\textsc{offline}.} Table~\ref{tab:offline-qwk} shows the performance of the six LLMs under the \textsc{offline} setting, where the full structured EHR is provided during inference. \textsc{claude} reports the best performance among all models (QWK=0.701) with \textsc{qwen-l} reporting the best performance as an open-weight model (QWK=0.638). The six models show moderate to substantial agreement with the ground truth. These offline scores set the ceiling that the sequential trajectory would approach if the model integrated the conversational evidence.

\noindent\textbf{\textsc{sequential}.} Figure~\ref{fig:rq1-sim} describes the model performances under \textsc{sequential} setting on the \textsc{simulated} corpus using QWK. Among all models, \textsc{gemma-l} reports the best performance at all checkpoints. Compared with the \textsc{offline} setting, all models degrade across all checkpoints. The agreement with the ground truth drops to fair to moderate. Model performances \textbf{plateau} at the initial checkpoint (\textit{k}=4), with evidence that subsequent checkpoints sometimes resulting in degraded performance. For example, \textsc{claude} and \textsc{qwen-l}, the best-performing models under \textsc{offline} degrade at later checkpoints when compared to \textit{k}=4.

\noindent\textbf{Other prompting strategies.} Figure~\ref{fig:rq1-reason} compares \textsc{vanilla} against three alternative prompting strategies on the \textsc{simulated} corpus: \textsc{r-on}, \textsc{ps+}, and \textsc{sc}. Across the six models and five checkpoints, \textsc{vanilla} matches or outperforms every alternative. \textsc{qwen-s} and \textsc{gemma-s}, the two smaller open-weight models, show minor gains under \textsc{ps+} and \textsc{r-on} respectively. The other four models degrade below \textsc{vanilla} under every strategy. The plateau at \textit{k=4} holds under every strategy on every model. For detailed results on the \textsc{clinician} corpus, see Appendix~\ref{app:rq1-reason-clin}.

\begin{figure*}[t!]
\centering
\begin{subfigure}[b]{0.42\linewidth}
    \centering
    \includegraphics[width=\linewidth]{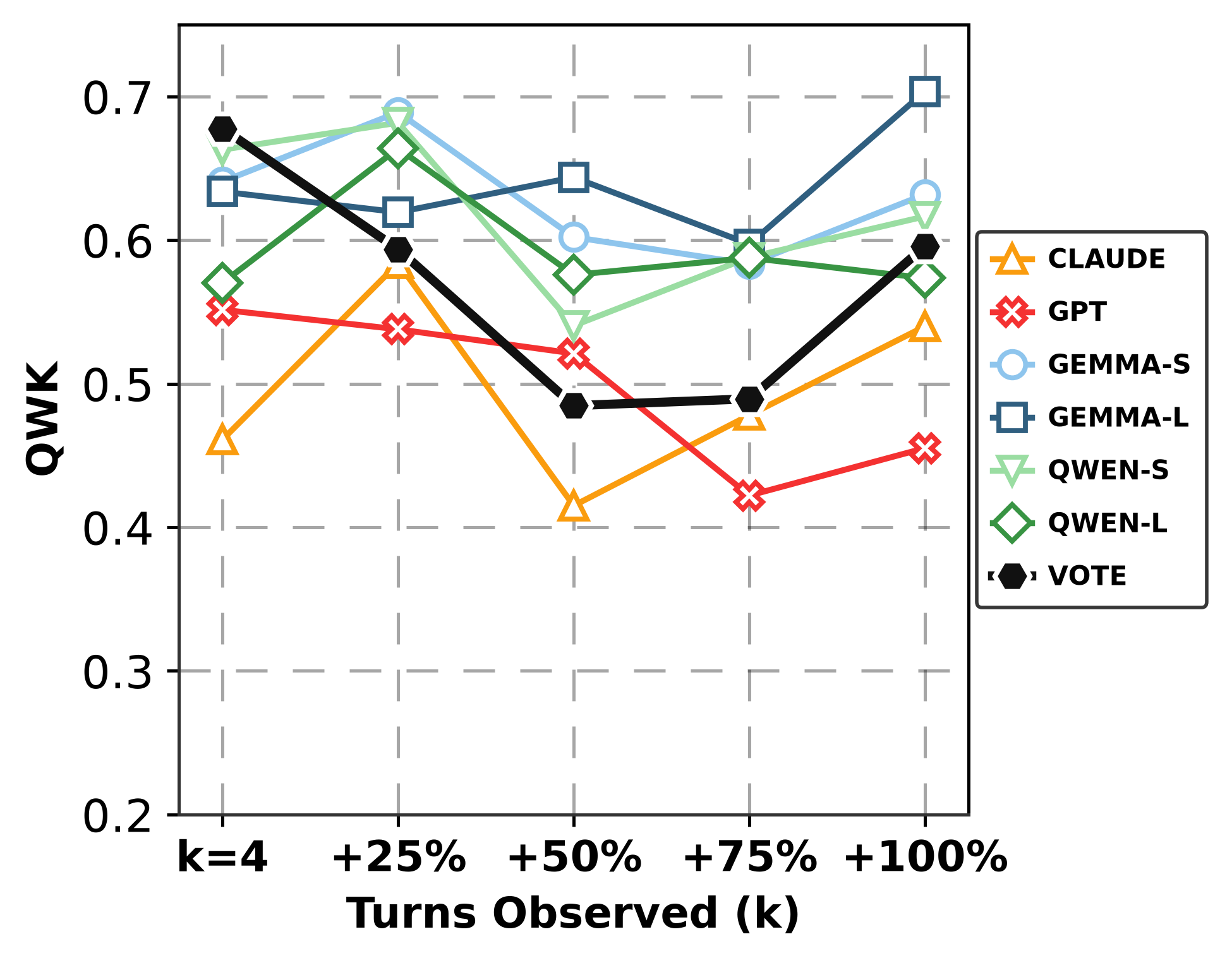}
    \caption{Performance comparison of all models across the five sequential checkpoints on the \textsc{clinician} corpus. \textsc{vote} is the majority ensemble across the models.}
    \label{fig:rq1-clin-seq}
\end{subfigure}
\hspace{0.03\linewidth}
\begin{subfigure}[b]{0.42\linewidth}
    \centering
    \includegraphics[width=0.77\linewidth]{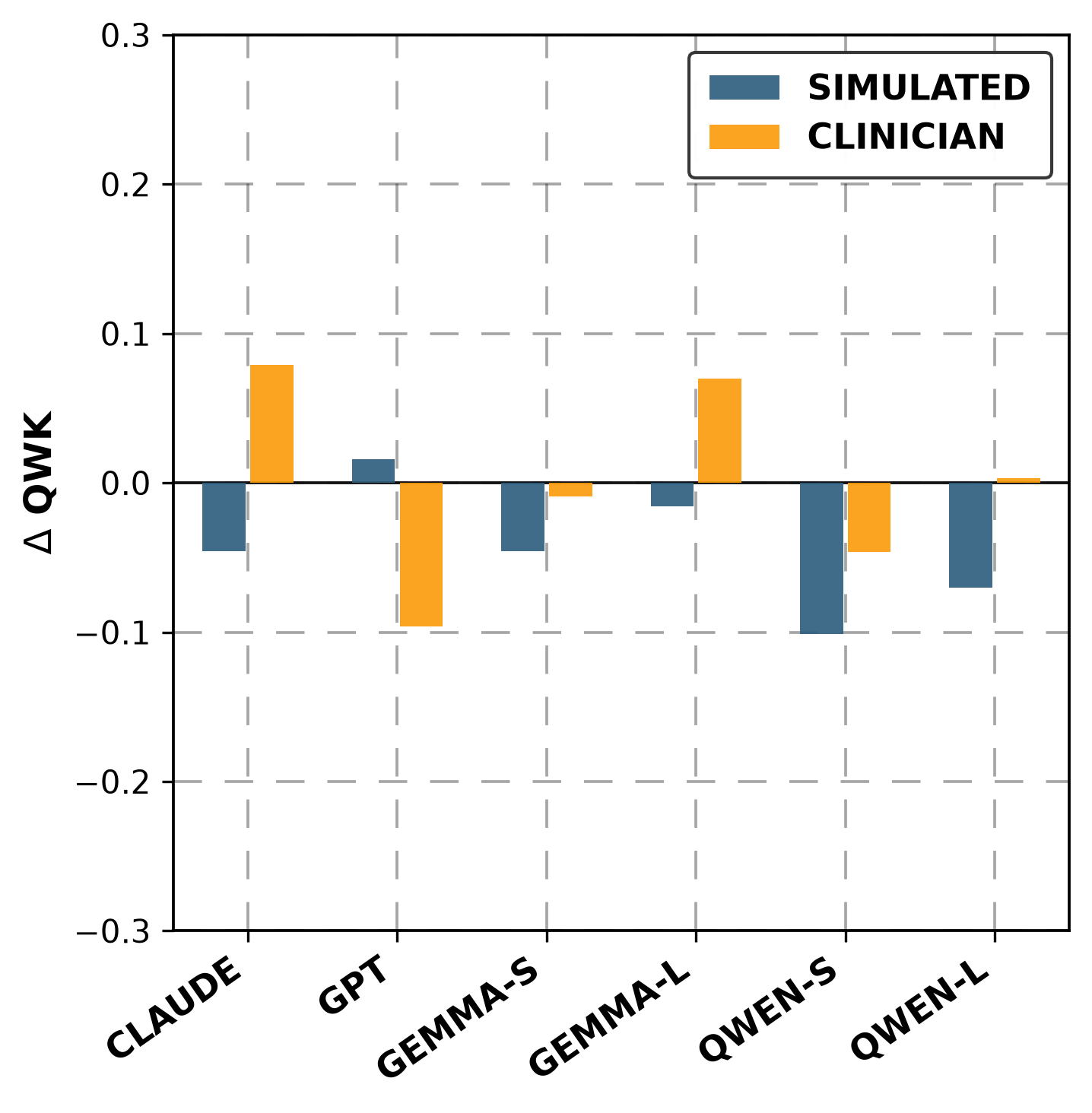}
    \caption{Change in performance of all models at \textit{k}=+100\% when compared with \textit{k}=4, reported on both \textsc{simulated} and \textsc{clinician} corpora.}
    \label{fig:rq1-clin-cross}
\end{subfigure}
\caption{Performance of all models on the \textsc{clinician} corpus and per-model $\Delta$~QWK from \textit{k}=4 to +100\% on both corpora.}
\label{fig:rq1-clin}
\end{figure*}

\noindent\textbf{Cross-corpus.} Figure~\ref{fig:rq1-clin-seq} shows the performance of all models on the \textsc{clinician} corpus under \textsc{vanilla} prompting. Three models report a minor gain over \textit{k}=4 at \textit{k}=+25\%: \textsc{claude}, \textsc{gemma-s}, and \textsc{qwen-l}. The \textsc{vote} ensemble degrades below \textit{k}=4 across the trajectory. Figure~\ref{fig:rq1-clin-cross} shows the difference in performance between predictions at \textit{k}=4 and \textit{k}=+100\%, across \textsc{simulated} and \textsc{clinician}. More turns of conversation show no significant improvement across both corpora.

\begin{figure*}[t!]
\centering
\begin{subfigure}[b]{0.35\linewidth}
    \centering
    \includegraphics[width=0.8\linewidth]{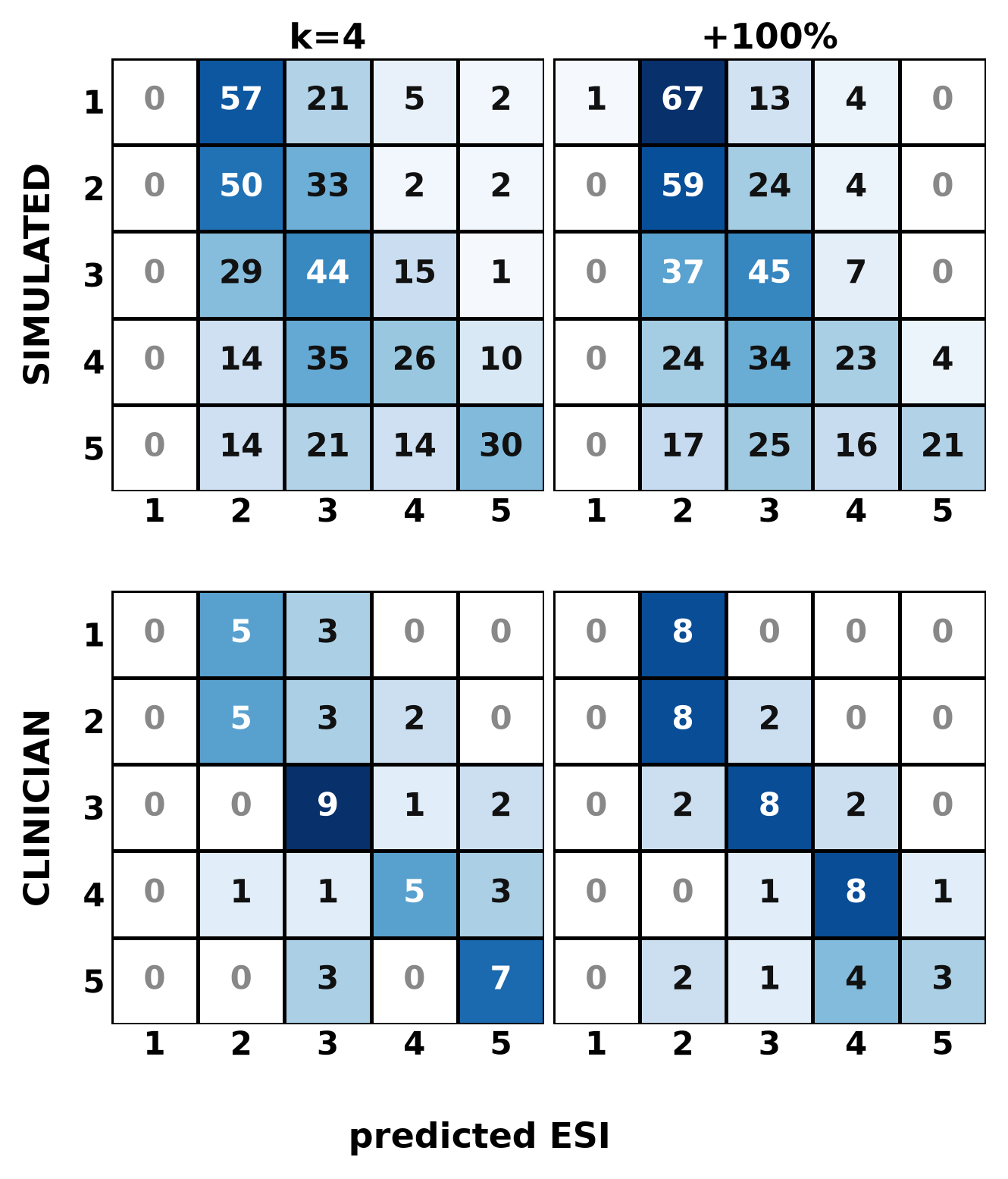}
    \caption{\textsc{gemma-l} confusion matrices at \textit{k=4} and \textit{k=+100\%}, on both corpora. Rows represent true acuity labels, and columns are predicted acuity labels.}
    \label{fig:rq1-cm}
\end{subfigure}
\hspace{0.03\linewidth}
\begin{subfigure}[b]{0.4\linewidth}
    \centering
    \includegraphics[width=0.9\linewidth]{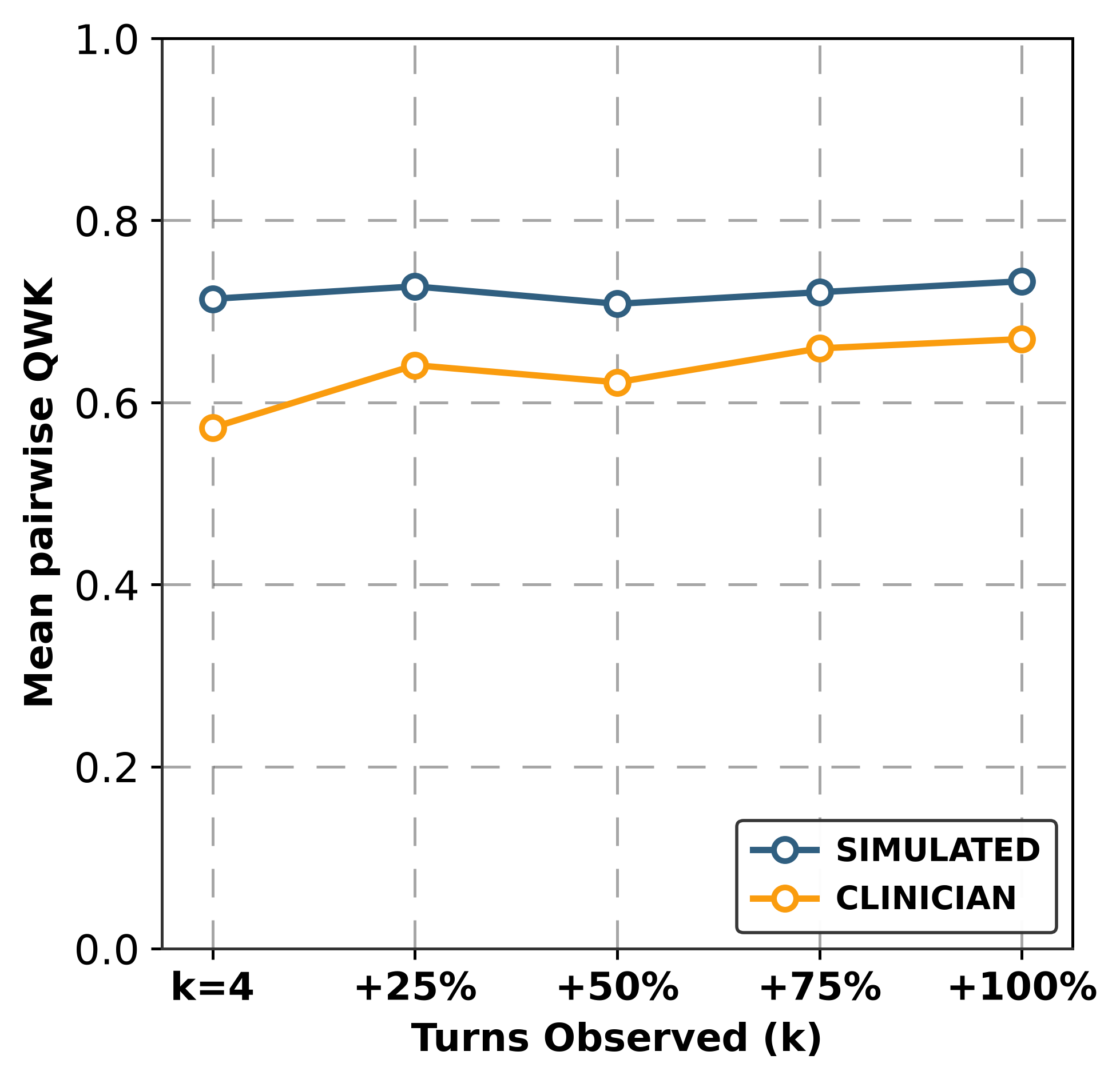}
    \caption{Pairwise agreement of all model predictions at each of the five sequential checkpoints, on both corpora.}
    \label{fig:rq1-iaa}
\end{subfigure}
\caption{\textsc{gemma-l} prediction shape and inter-model agreement across all checkpoints under \textsc{sequential}.}
\label{fig:rq1-other}
\end{figure*}

\noindent\textbf{Shape of predictions.} Figure~\ref{fig:rq1-cm} reports the \textsc{gemma-l} confusion matrix at \textit{k}=4 and \textit{k}=+100\% on both corpora. Predictions concentrate at ESI-2 and ESI-3, on both corpora and at both checkpoints. The other five models produce the same shape (Appendix~\ref{app:rq1-cm-all}).

\noindent\textbf{Model agreement.} Figure~\ref{fig:rq1-iaa} reports QWK computed across all model predictions at each sequential checkpoint, on both corpora. The six models show substantial agreement at every checkpoint, higher than each model's agreement with the ground truth.

\subsection{Where in the conversation does the model commit its triage decision?}
\label{sec:rq2}

\begin{figure*}[t!]
\centering
\includegraphics[width=\linewidth]{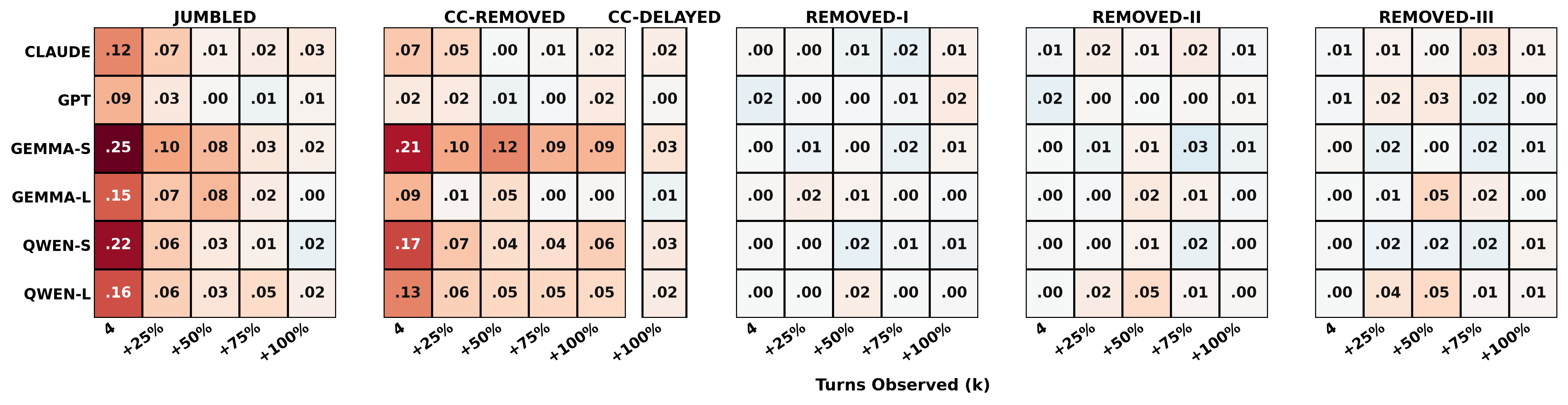}
\caption{Change in QWK per perturbation, model, and checkpoint on the \textsc{simulated} corpus. Blue = perturbation degrades QWK relative to \textsc{vanilla}; red = perturbation improves QWK.}
\label{fig:rq2-abl}
\end{figure*}

\noindent\textbf{Setup.} To localise the predictive signal, we evaluate \textit{six} controlled perturbations on the \textsc{simulated} corpus: 
\begin{itemize}[noitemsep]
\item \textbf{\textsc{jumbled}} permutes turn pairs under a deterministic seed.
\item \textbf{\textsc{cc-removed}} deletes the \textit{k}=4 turns.
\item \textbf{\textsc{cc-delayed}} relocates those \textit{k}=4 turns to \textit{k}=+100\%, and is identical to \textsc{cc-removed} through \textit{k}=+75\%.
\item \textbf{\textsc{removed-i}} deletes the utterances at \textit{k} $\in$ (4, +25\%].
\item \textbf{\textsc{removed-ii}} deletes the utterances at \textit{k} $\in$ (+25\%, +50\%].
\item \textbf{\textsc{removed-iii}} deletes the utterances at \textit{k} $\in$ (4, +50\%].
\end{itemize}
We run each perturbation under \textsc{vanilla} prompting across all six models and five checkpoints. Figure~\ref{fig:rq2-abl} reports the $\Delta$ QWK grid relative to the unperturbed \textsc{vanilla} baseline. Permuting turn order (\textsc{jumbled}) shifts QWK on every model at every checkpoint, corroborating the finding that the model's decision is fixed by whatever occupies the opening \textit{k}=4 checkpoint. Deleting the \textit{k}=4 turns (\textsc{cc-removed}) shifts QWK by 0.02 to 0.21 at \textit{k}=4, and the shift persists at every subsequent checkpoint. Relocating those turns to \textit{k}=+100\% (\textsc{cc-delayed}) matches \textsc{cc-removed} through \textit{k}=+75\% and recovers to baseline at \textit{k}=+100\% on every model. Deleting utterances at later checkpoints (\textsc{removed-i/ii/iii}) produces near-zero deltas across all models and checkpoints. These findings indicate that the predictive signal for sequential clinical triage lies in the utterances at \textit{k}=4. The first four utterances of each conversation carry the patient's chief complaint (CC), the presenting reason for the ED visit. The same pattern holds on the \textsc{clinician} corpus (Appendix~\ref{app:rq2-clin}).

\begin{figure*}[t!]
    \centering
    \includegraphics[width=0.85\linewidth]{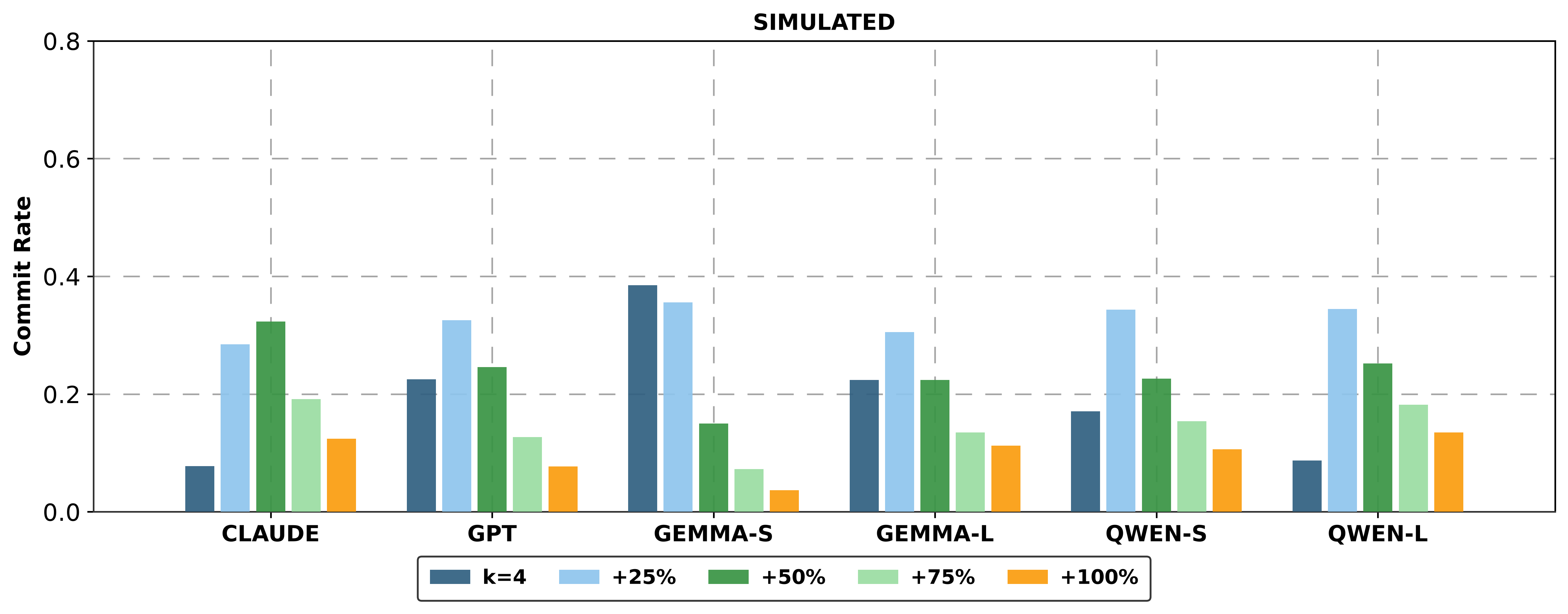}
    \caption{Proportion of conversations committed at each checkpoint. Commit rates on the \textsc{clinician} corpus are in Appendix~\ref{app:rq3-clinician}.}
    \label{fig:rq3-commit-sim}
\end{figure*}

\subsection{How does a model's chosen commitment to an acuity level compare to a clinician's under the same evidence?}
\label{sec:rq3}

\noindent\textbf{Setup.} The results presented so far require the model to predict a triage label at every checkpoint. For this experiment, we make the model choose at each checkpoint whether to predict a label or defer to the next checkpoint. A conversation that reaches the final checkpoint without committing must predict a label at \textit{k}=+100\%. We report performance using the triage label predicted at the point of commit (QWK-at-commit).

\begin{figure}[t!]
    \centering
    \includegraphics[width=0.9\linewidth]{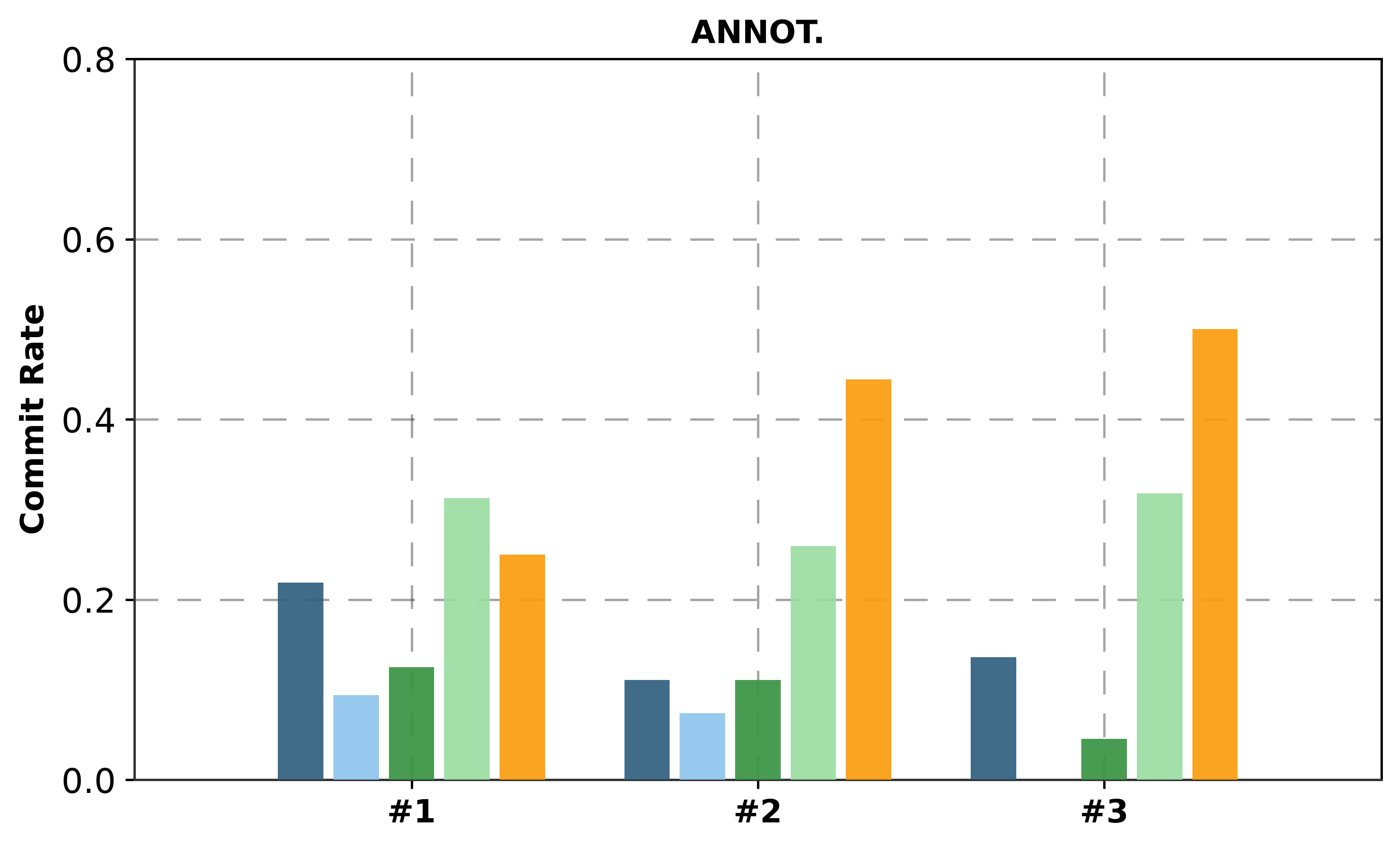}
    \caption{Commit rates per expert annotator on 50 conversations from \textsc{simulated}.}
    \label{fig:rq3-commit-expert}
\end{figure}

Figure~\ref{fig:rq3-commit-sim} reports the mean count of conversations at which each model commits, per checkpoint, on \textsc{simulated}. Most models commit at \textit{k}=+25\% or \textit{k}=+50\%, skewing the distribution toward earlier checkpoints. We run the same experiment with \textit{three} expert clinician annotators on a random sample of 50 conversations from the \textsc{simulated} corpus.\footnote{See Appendix~\ref{app:annot-exercise} for details regarding the annotation guidelines and other details.} Figure~\ref{fig:rq3-commit-expert} shows that the expert clinicians tend to wait longer before committing to triage, making the distribution skewed towards the later checkpoints. Table~\ref{tab:rq3-qwk-at-commit} reports the QWK-at-commit on the same 50 conversations from \textsc{simulated}. Annotators show near-perfect agreement with the ground-truth ESI label, while the best-performing model reports fair agreement (\textsc{gemma-l}: 0.295).

\begin{table*}[t!]
\begin{adjustbox}{width=\linewidth, center}
\begin{tabular}{lccccccccc}
\toprule
& \textsc{gemma-s} & \textsc{gemma-l} & \textsc{qwen-s} & \textsc{qwen-l} & \textsc{claude} & \textsc{gpt} & \textbf{\textsc{annot \#1}} & \textbf{\textsc{annot \#2}} & \textbf{\textsc{annot \#3}} \\
\midrule
\textbf{QWK} & 0.009 & 0.295 & 0.016 & 0.195 & 0.215 & 0.280 & 0.929 & 0.943 & 0.887 \\
\bottomrule
\end{tabular}
\end{adjustbox}
\caption{Performance of LLMs and expert clinicians at point of commit on 50 conversations from \textsc{simulated}.}
\label{tab:rq3-qwk-at-commit}
\end{table*}

\subsection{Why does more evidence not improve sequential clinical triage?}
\label{sec:rq4}

\begin{table}[t!]
\begin{adjustbox}{width=0.6\linewidth, center}
\begin{tabular}{ccc}
\toprule
ESI & \textsc{qwen-s} & \textsc{gemma-l} \\
\midrule
1 & 0.660 & 0.696 \\
2 & 0.692 & 0.753 \\
3 & 0.696 & 0.733 \\
4 & 0.695 & 0.772 \\
5 & 0.565 & 0.642 \\
\bottomrule
\end{tabular}
\end{adjustbox}
\caption{Cosine similarity between drafted and ground truth chief complaints from $\mathcal{X}^-$, stratified by ESI.}
\label{tab:cc-recovery}
\end{table}

The annotation exercise shows that clinicians wait longer to acquire more evidence from the conversations. For a model to perform sequential clinical triage without anchoring on the chief complaint, it must perform two actions on the conversation: (a) \textit{extract} information from the rest of the conversation, and (b) \textit{integrate} the extracted information into a triage decision.

\noindent\textbf{Chief complaint extraction.} We prompt each model to draft a chief complaint from the non-CC turns ($\mathcal{X}^-:= \mathcal{X} \setminus d_{k=4}$) and compute the cosine similarity to the ground truth chief complaint with \texttt{embeddinggemma-300m-medical} from SentenceTransformers~\citep{reimers-gurevych-2019-sentence}. Table~\ref{tab:cc-recovery} reports the cosine for the weakest (\textsc{qwen-s}) and strongest (\textsc{gemma-l}) models under \textsc{sequential} setting. The values range from 0.57 to 0.77 across the five ESI classes on \textsc{simulated}. Even the weakest model recovers the chief complaint from $\mathcal{X}^-$ on every ESI class. Per-model scores are in Appendix~\ref{app:cc-recovery}.

\begin{figure*}[t!]
\centering
\begin{subfigure}[b]{0.42\linewidth}
    \centering
    \includegraphics[width=0.85\linewidth]{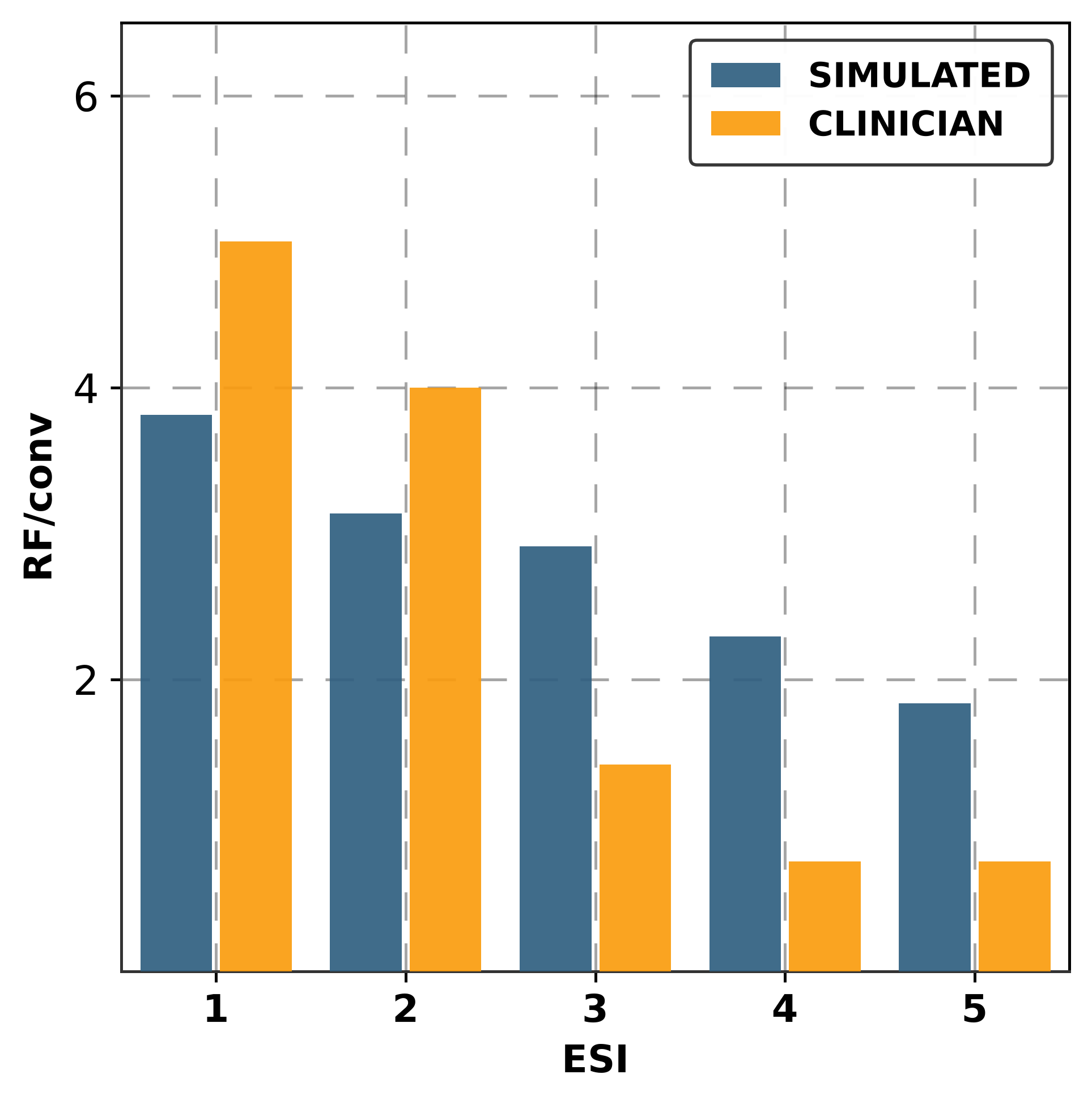}
    \caption{Mean count of identified red flag phrases per conversation, on \textsc{simulated} and \textsc{clinician}.}
    \label{fig:rq4-rf}
\end{subfigure}
\hspace{0.01\linewidth}
\begin{subfigure}[b]{0.42\linewidth}
    \centering
    \includegraphics[width=0.8\linewidth]{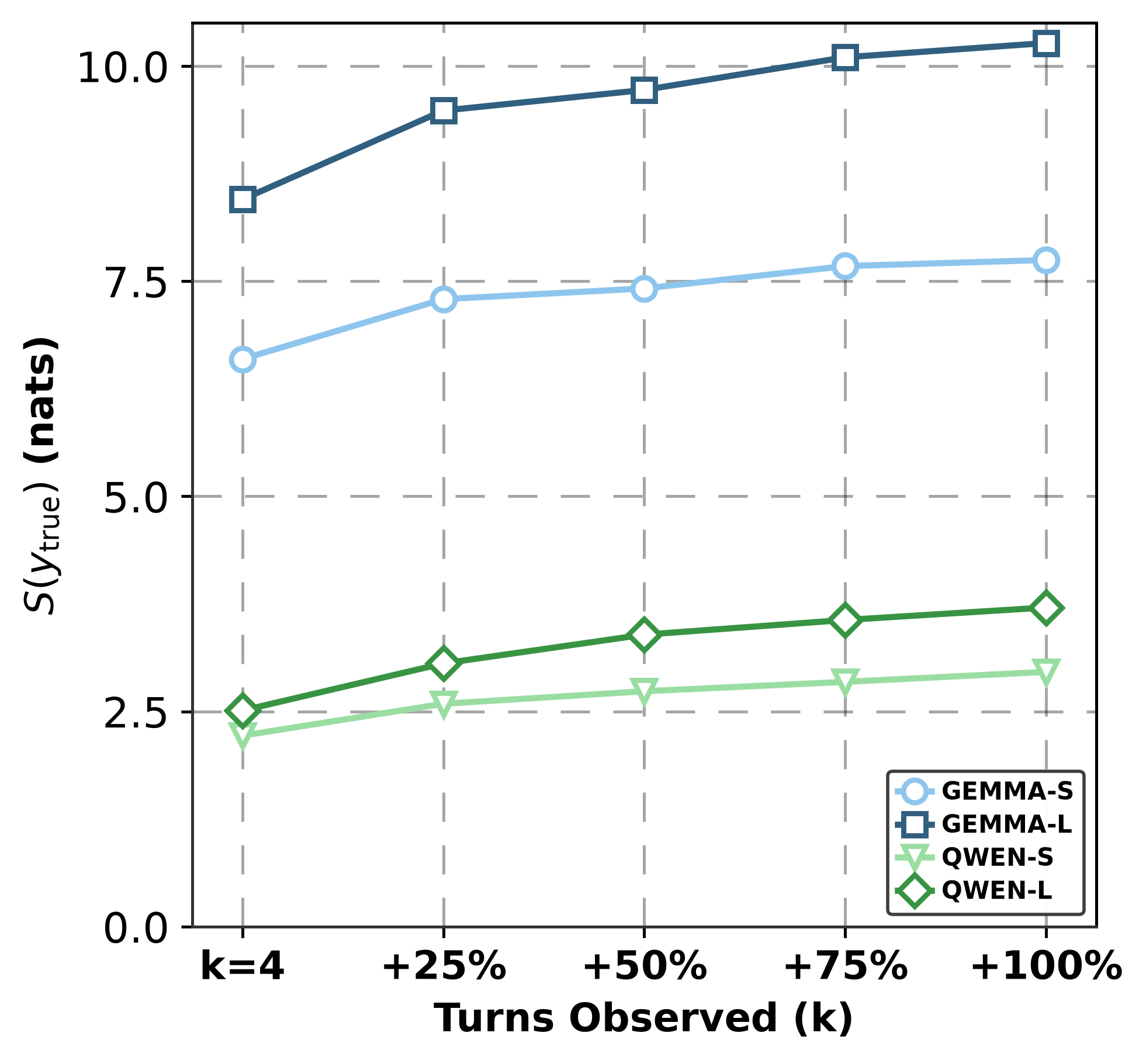}
    \caption{Surprisal of the true ESI label across the five checkpoints under \textsc{sequential} setting, for the four open-weight models on \textsc{simulated}. \textsc{clinician} is in Appendix~\ref{app:ci-surp}.}
    \label{fig:rq4-surprisal}
\end{subfigure}
\caption{Extraction of red flags in both corpora and surprisal of the true ESI label per checkpoint in \textsc{simulated}.}
\label{fig:rq4}
\end{figure*}

\noindent\textbf{Red flag extraction.} We prompt \textsc{claude} and \textsc{gpt} to extract red flags from $\mathcal{X}^-$ conversations using the criteria provided by \citet{ESPEJO202557}. Figure~\ref{fig:rq4-rf} reports the number of red flags jointly identified by the two models, computed as $\lvert\text{RF}_{\textsc{claude}} \cap \text{RF}_{\textsc{gpt}}\rvert$, per conversation and ESI class. Red-flag density decreases with acuity: ESI-1 cases contain 3.8 (\textsc{simulated}) and 5.0 (\textsc{clinician}) red flags per case, compared with 1.8 and 0.8 for ESI-5 cases, respectively. Hence, both models consistently identify clinically relevant red flags.

\noindent\textbf{Evidence integration.} We measure integration using a token-level negative log-likelihood of the true ESI label across the five sequential checkpoints. We define the measure symbolically as: $\mathcal{S}(y_{\text{true}}) = -\ln p(y_{\text{true}} \mid d_k; \theta)$

Figure~\ref{fig:rq4-surprisal} reports the mean surprisal of the true ESI label across the five sequential checkpoints, for the four open-weight models on \textsc{simulated}. Surprisal rises monotonically for every model between \textit{k}=4 and \textit{k}=+100\% (net gain +0.74 to +1.82 nats). This shows that the model updates the posterior over the ESI labels as more checkpoints are revealed. Although the model extracts clinically relevant information from the conversation, the mapping to the true label moves in the wrong direction.

%% file: sections/02-related-works.tex
\section{Related Work}
\label{sec:related}

\noindent\textbf{LLMs for Emergency Triage.} Automatic emergency triage sits within a broader family of NLP tasks studied across the ED pipeline~\citep{srirag2026triagedischargesurveynlp}. Within triage, existing evaluations of LLMs operate on retrospective records, comparing model predictions to acuity labels assigned by ED nurses or physicians. Reported behaviours range from physician-comparable discrimination~\citep{williams2024llm} to systematic over-triage matching untrained physicians~\citep{Masanneck2024}, with multi-agent, knowledge-prompted, and reasoning-model variants reporting gains over single-model baselines~\citep{lu-etal-2024-triageagent, Liu2025, Brodeur2026}. LLMs also show sociodemographic disparities, over-triaging marginalised groups when clinical content is held constant~\citep{omar2025sociodemographic}, mirroring nurse-driven disparities against Black and Hispanic patients~\citep{Joseph2023}. Nurse triage itself shows measured mistriage rates across US EDs~\citep{sax2023mistriage, Sax2025}. Across this literature, evaluations use all information available at inference. We instead evaluate LLM triage at five checkpoints across the nurse-patient encounter and find that additional evidence beyond the CC utterances does not improve performance on the task.

\noindent\textbf{Position sensitivity and conversational clinical evaluation.} 
In clinical settings, diagnostic accuracy is sensitive to the quantity and order of presented information~\citep{hager2024evaluation}, and recent work converts retrospective vignettes into multi-turn conversational benchmarks~\citep{li2024mediq,Johri2025-craft-md, chiu2025vivabench, maiwert2026q4dx}. These benchmarks report performance degradation under the conversational format, with the proposed mechanism centring on early commitment to an initial hypothesis~\citep{chiu2025vivabench} and on the broader unreliability of LLMs in multi-turn interaction~\citep{laban2025lost}. Prompting mitigations include chain-of-thought~\citep{wei2022chain}, plan-and-solve~\citep{wang2023plansolve}, and self-consistency~\citep{wang2023selfconsistency}. Diagnostic dialogue systems frame the task as iterative symptom acquisition~\citep{wei2018task, chen2022diaformer} or sequential probing over open-ended diagnosis~\citep{nori2025sequentialdiagnosislanguagemodels}, but do not track how predictions change as evidence is revealed. None of this work targets ordinal triage classification, where the structurally privileged opening is the utterance that serves as the hypothesis for the triage algorithm. 

\noindent\textbf{Anchoring in clinical and computational reasoning.} Anchoring bias, the tendency to rely on initial information under uncertainty \citep{tversky1974judgment}, is documented in clinical decision-making, including emergency medicine \citep{croskerry2002achieving, ly2023anchoring}. LLMs exhibit analogous effects, shifting toward initial or injected cues, with standard mitigations failing to eliminate them~\citep{jones2022capturing, echterhoff2024cognitive, schmidgall2024cognitive}. These studies introduce anchors artificially through prompt design or document ordering. We study anchoring on the chief complaint, a structurally privileged utterance whose early position is inherent to the clinical encounter. Our perturbation suite (Section~\ref{sec:rq3}) localises the anchoring signal to that utterance and separates the effect of its content from its position.


%% file: sections/06-discussion.tex
\section{Conclusion}
\label{sec:conclusion}

We implement a methodology for evaluating LLMs on sequential clinical triage where incremental evidence is made available to the LLM. Our evaluation of six LLMs across two settings and two corpora shows a plateau in task performance as the conversation prefix grows. Interventions like reasoning-native decoding (\textsc{r-on}) and Plan-and-Solve (\textsc{ps+}) fail to eliminate the plateau. Controlled perturbations of the input prefixes characterise the effect as \textbf{chief-complaint anchoring}. LLMs extract clinically relevant information beyond the chief complaint from the conversations. When asked to commit to an ESI label or defer at each checkpoint, models commit earlier and perform worse than expert clinicians who wait for later turns before committing. The gap shows that later turns hold information the models could use to update their predictions. Yet, the surprisal of the ground-truth label increases with additional evidence, showing that the models \textbf{fail to integrate the evidence} for sequential clinical triage.

Our analysis shows that \textbf{every model at every checkpoint under-triages ESI-1 (highest acuity) cases}. Predictions on lower-acuity cases concentrate at ESI-2 or ESI-3 on both corpora. Models also agree with each other more than with the ground truth, so ensemble methods would exacerbate anchoring and evidence insensitivity. Current methods for triage classification report high performance on completed retrospective records \citep{lu-etal-2024-triageagent}, yet sequential clinical triage is not addressed. Deployment of such systems in real-world EDs would face a mismatch between offline evaluations and sequential encounters~\citep{srirag2026triagedischargesurveynlp, yao-yu-2026-llm}. We recommend that future research evaluate LLMs for triage in the \textsc{sequential} setting and focus on methods that address failures to integrate evidence.

%% file: sections/07-limitations.tex
\section*{Limitations}
\label{sec:limitations}
We present our analysis of sequential clinical triage using \textsc{simulated} and \textsc{clinician} due to restricted or lack of access to real triage interactions. We perform the task with acuity labels derived from ESI, with five ordinal levels. Generalising the findings to Australasian Triage Scale or Manchester Triage System requires generation of similar synthetic conversations derived from structured EHR datasets label under these triage systems. We do not attempt to provide a mechanistic explanation of \textit{why} integration fails at the label head. This would require linear probes over per-checkpoint hidden states~\citep{alain2018understandingintermediatelayersusing} and causal patching between the \textit{k}=4 and \textit{k}=+100\% activations~\citep{meng2022locating}, and each needs a larger training corpus of triage interactions.

\section*{AI Usage Statement}
In this work, we used a privacy-preserving generative AI tool to help draft and revise portions of the text. Additionally, we used generative AI tools to create the web-based interface for annotation exercise, and polish and create documentation for the codebase. We did not use generative AI tools for research ideation or methodology. We have reviewed all AI-assisted work. We take responsibility for the final content of this work, including text, claims, and artifacts produced with the aid of generative AI.

\section*{Ethical Considerations}
The study uses de-identified data from MIMIC-IV-ED under the PhysioNet Credentialed Health Data License. The \textsc{simulated} corpus is generated by an LLM seeded from MIMIC-IV-ED cases and carries no identifiers beyond those already de-identified in the source. The \textsc{clinician} corpus is clinician-authored and contains no real patient information. The exercise involving clinicians in Section~\ref{sec:rq3} was approved by the host organisation's ethics committee. The paper reports safety findings about pretrained LLMs on sequential clinical triage, and we disclose them to aid deployment decisions in real-world settings.

\section*{Reproducibility Statement}
We release all code used for evaluation, including prompt templates, model-call scripts and analysis pipelines at: \url{https://anonymous.4open.science/r/seq-triage-eval/}. We do not release the \textsc{simulated} and \textsc{clinician} conversations directly. The conversations are derived from MIMIC-IV-ED which is governed by a PhysioNet credentialed-access data use agreement~\citep{Johnson2023} that prohibits redistribution of derivative data containing the underlying clinical content. Instead, we release the case identifiers (\textit{stay\_id}) and the generation parameters used. Researchers with PhysioNet credentials can regenerate the conversations from the released identifiers and parameters.

\section*{Acknowledgments}
Dipankar Srirag is supported by NHMRC Ideas Grant (RG241647), awarded to Padmanesan Narasimhan and Aditya Joshi in 2025.


%% file: appendix/A-esi.tex
\section{The Emergency Severity Index}
\label{sec:a-esi-algorithm}

\begin{figure}[t!]
    \centering
    \includegraphics[width=1\linewidth]{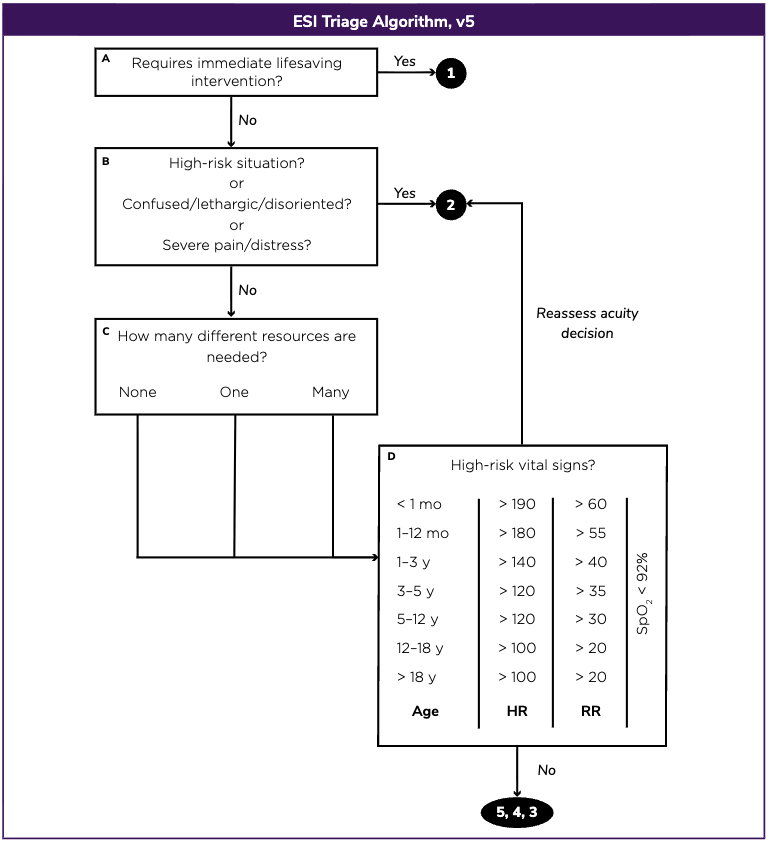}
    \caption{The ESI triage algorithm; accessed from the \href{https://media.emscimprovement.center/documents/Emergency_Severity_Index_Handbook.pdf}{ESI Handbook}.}
    \label{fig:esi-algo}
\end{figure}

The Emergency Severity Index is a five-level acuity triage schema used in the majority of US EDs \citep{esi-handbook}. Figure~\ref{fig:esi-algo} describes the ESI triage algorithm as four-stage decision making process. The chief complaint has a specific position in this process. Questions A and B resolve on the chief complaint together with the initial vital signs, before the interview elicits history or additional symptoms. Post-CC utterances narrow down the projected resource count for question C and can surface a delayed vital-sign concern under question D. The \textit{k=4} checkpoint of a conversation provides the model with the chief complaint exchange. The four proportional checkpoints (\textit{k} $\in$ \{+25\%,+50\%,+75\%,+100\%\}) reveal the post-CC turns that carry resource and vital-sign information.

%% file: appendix/B-prompts.tex
\newcounter{promptbox}
\renewcommand{\thepromptbox}{\arabic{promptbox}}

\section{Prompt templates}
\label{app:prompts}

We serialise the ESI algorithm into a system prompt for all experiments in Section~\ref{sec:results} (Box ~\ref{box:algorithm}).

\refstepcounter{promptbox}\label{box:algorithm}%
\begin{tcolorbox}[enhanced, breakable, colback=pastelred!10, colframe=pastelred!80!black, fonttitle=\bfseries\small, title={\texttt{ALGORITHM}}]
\small
ESI TRIAGE RULES (FOLLOW EXACTLY)\\[3pt]
\textbf{Step A: Immediate life-saving intervention.} Assign ESI level 1 if the patient requires immediate life-saving intervention, such as airway support, severe respiratory distress, shock, unresponsiveness, or active seizure. If YES, then ESI = 1. If NO, proceed to Step B.\\[3pt]
\textbf{Step B: High-risk situation, severe pain, or altered mental status.} Assign ESI level 2 if ANY of the following are present: high-risk situation with potential for rapid deterioration; confused, lethargic, or disoriented mental status; severe pain or distress. If YES, then ESI = 2. If NO, proceed to Step C.\\[3pt]
\textbf{Step C: Predicted resource needs.} Estimate how many DIFFERENT types of ED resources are required. Resources include: labs (1 resource: blood and urine); scanning (1 resource each: X-ray, ECG, CT, MRI, ultrasound, angiography); IV fluids (1 resource); IV, IM, or nebulised medications (1 resource each); specialist consultation (1 resource); simple procedure (1 resource, e.g.\ laceration repair, urinary catheter); complex procedure (2 resources, procedures needing sedation or anaesthesia). Prescription refills, simple wound care, and salines are NOT counted as resources. Based on this definition: no resources, then ESI = 5; one resource, then ESI = 4; two or more resources, then proceed to Step D.\\[3pt]
\textbf{Step D: High-risk vital signs.} Assign ESI = 3. If any vital sign is outside the age-specific danger zone below, upgrade to ESI = 2.\\[3pt]
\hspace*{1em}-- $<$ 1 mo: HR $>$ 190, RR $>$ 60, SpO$_2$ $<$ 92\%.\\
\hspace*{1em}-- 1 to 12 mo: HR $>$ 180, RR $>$ 55, SpO$_2$ $<$ 92\%.\\
\hspace*{1em}-- 1 to 3 y: HR $>$ 140, RR $>$ 40, SpO$_2$ $<$ 92\%.\\
\hspace*{1em}-- 3 to 5 y: HR $>$ 120, RR $>$ 35, SpO$_2$ $<$ 92\%.\\
\hspace*{1em}-- 5 to 12 y: HR $>$ 120, RR $>$ 30, SpO$_2$ $<$ 92\%.\\
\hspace*{1em}-- 12 to 18 y: HR $>$ 100, RR $>$ 20, SpO$_2$ $<$ 92\%.\\
\hspace*{1em}-- $>$ 18 y: HR $>$ 100, RR $>$ 20, SpO$_2$ $<$ 92\%.
\end{tcolorbox}

\subsection{\textsc{offline} prompt}
\label{app:prompts-offline}

Under the \textsc{offline} setting, we prompt the model with the full structured EHR record (Box~\ref{box:offline}).

\refstepcounter{promptbox}\label{box:offline}%
\begin{tcolorbox}[enhanced, breakable, colback=pastelblue!10, colframe=pastelblue!80!black, fonttitle=\bfseries\small, title={\textsc{offline} prompt}]
\small
\texttt{\textbf{SYSTEM}}:\\
\textbf{\texttt{<<ALGORITHM>>}}.\\ 
You are an experienced emergency triage nurse, trained in the Emergency Severity Index (ESI) triage algorithm. Using the provided ESI triage rules, perform triage classification on the presented patient record.\\
\texttt{\textbf{USER}}: \\
Patient record:\\
\textbf{\texttt{<<record>>}}\\[3pt]
Assign an ESI triage level (1 = most critical, 5 = least urgent).\\[3pt]
Respond in this exact format:\\
\texttt{
Triage: [1/2/3/4/5]
}
\end{tcolorbox}

\subsection{\textsc{sequential} prompt}
\label{app:prompts-sequential}

The \textsc{sequential} setting reveals a conversation prefix and prompts the model once per checkpoint. The base prompt is presented in Box~\ref{box:seq-vanilla}.

\refstepcounter{promptbox}\label{box:seq-vanilla}%
\begin{tcolorbox}[enhanced, breakable, colback=pastelblue!10, colframe=pastelblue!80!black, fonttitle=\bfseries\small, title={\textsc{vanilla}}]
\small
\texttt{\textbf{SYSTEM}}:\\
\textbf{\texttt{<<ALGORITHM>>}}.\\ 
You are an experienced emergency triage nurse assessing a patient in real time, trained in the Emergency Severity Index (ESI) triage algorithm. You will be shown a triage conversation. Perform triage classification based on the available evidence and the provided ESI triage rules.\\[6pt]
\textbf{\texttt{USER}}: Triage conversation:\\
\texttt{\textbf{<<prefix>>}}\\[3pt]
Assign an ESI triage level (1 = most critical, 5 = least urgent).\\[3pt]
Respond in this exact format:\\
\texttt{
Triage: [1/2/3/4/5]
}
\end{tcolorbox}

\paragraph{\textsc{r-on}.} This setting uses the same prompt as \textsc{vanilla}. The model's native chain-of-thought mode is enabled under this setting. We parse the reasoning trace from within the \texttt{<think>\ldots</think>} block.

\paragraph{\textsc{ps+}.} The prompt for this setting is described in Box~\ref{box:seq-ps}. The Plan-and-Solve preamble \citep{wang2023plansolve} is appended to the \textsc{sequential-vanilla} prompt.

\refstepcounter{promptbox}\label{box:seq-ps}%
\begin{tcolorbox}[enhanced, breakable, colback=pastelblue!10, colframe=pastelblue!80!black, fonttitle=\bfseries\small, title={\textsc{ps+}}]
\small
\texttt{\textbf{SYSTEM}}:\\
\textbf{\texttt{<<ALGORITHM>>}}.\\ 
You are an experienced emergency triage nurse assessing a patient in real time, trained in the Emergency Severity Index (ESI) triage algorithm. You will be shown a triage conversation. Perform triage classification based on the available evidence and the provided ESI triage rules.\\[3pt]
Before answering, follow these steps in order:\\
\hspace*{1em}1. Understand the patient's presenting complaint from the conversation.\\
\hspace*{1em}2. Extract the relevant clinical variables: vital signs, red-flag symptoms, mechanism of injury/onset, comorbidities, medications, and any high-risk features.\\
\hspace*{1em}3. Devise a plan by identifying which branch of the ESI decision algorithm applies.\\
\hspace*{2em}-- Life-threatening $\rightarrow$ Level 1.\\
\hspace*{2em}-- High-risk, severe distress, or immediate stabilisation $\rightarrow$ Level 2.\\
\hspace*{2em}-- Requires multiple resources $\rightarrow$ Level 3.\\
\hspace*{2em}-- Requires one resource $\rightarrow$ Level 4.\\
\hspace*{2em}-- Requires no resources $\rightarrow$ Level 5.\\
\hspace*{1em}4. Execute the plan by applying the ESI algorithm to the extracted variables and select the highest-acuity level that fits.\\
Then output the final label in the exact required format.\\[3pt]
\textbf{\texttt{USER}}: Triage conversation:\\
\texttt{\textbf{<<prefix>>}}\\[3pt]
Assign an ESI triage level (1 = most critical, 5 = least urgent).\\[3pt]
Respond in this exact format:\\
\texttt{
Triage: [1/2/3/4/5]
}
\end{tcolorbox}

\paragraph{\textsc{sc}.} The prompt used for this setting is identical to \textsc{vanilla}. Self-Consistency \citep{wang2023selfconsistency} draws five samples at temperature $0.7$ with top-$p=0.95$, and returns the majority-vote label across the five samples.

\subsection{Chief-complaint recovery}
\label{app:prompts-cc}

We prompt models to generate chief complaint from post-CC utterance (Box~\ref{box:cc-system}), as described in Section \ref{sec:rq4}. The reply is parsed as everything after the literal \texttt{Chief complaint:}.

\refstepcounter{promptbox}\label{box:cc-system}%
\begin{tcolorbox}[enhanced, breakable, colback=pastelred!10, colframe=pastelred!80!black, fonttitle=\bfseries\small, title={\texttt{SYSTEM} for chief-complaint recovery}]
\small
You are analysing a conversation between a triage nurse and a patient in an emergency department. Based on the conversation below, infer what the patient's chief complaint.\\[3pt]
Reply in a single line, the exact form a triage nurse would document the chief complaint (examples: ``chest pain'', ``shortness of breath'', ``abdominal pain and vomiting'', ``altered mental status''). Do not include explanation, quotes, or angle brackets. Begin your reply with the literal text `Chief complaint:' followed by the phrase.
\end{tcolorbox}

\subsection{Evidence-annotation (red flags)}
\label{app:prompts-rf}

We independently prompt \textsc{claude} and \textsc{gpt} to identify any red flags present in the post-CC utterances of each conversation. The prompt is in Box~\ref{box:rf-system}.

\refstepcounter{promptbox}\label{box:rf-system}%
\begin{tcolorbox}[enhanced, breakable, colback=pastelred!10, colframe=pastelred!80!black, fonttitle=\bfseries\small, title={\texttt{SYSTEM} for evidence-annotation}]
\small
\textbf{\texttt{SYSTEM:}} \\
\textbf{\texttt{<<ALGORITHM>>}}\\
You are a triage clinical educator reviewing a conversation between a triage nurse and a patient in an emergency department. Based on the conversation, identify red-flag phrases for each of the six categories below.\\[3pt]
Return a single JSON object with EXACTLY these six keys. Empty list if no phrase belongs to a category. Extract phrases only from what is stated in the conversation; do not infer beyond it.\\[3pt]
\hspace*{1em}-- \texttt{critical}: Lifesaving Intervention like cardiac arrest, unstable airway, active hemorrhage, imminent need for intubation.\\[2pt]
\hspace*{1em}-- \texttt{high\_risk}: High-Risk Presentation warranting immediate second-level assessment, like suspected stroke, sepsis, meningitis, severe respiratory distress, altered mental status, suicidal or homicidal ideation, severe pain.\\[2pt]
\hspace*{1em}-- \texttt{danger\_vitals}: Vital-sign findings outside the age-appropriate danger zone.\\[2pt]
\hspace*{1em}-- \texttt{nonspecific}: Nonspecific warning symptoms that raise concern without meeting above criterion, like ``intermittent dizziness'' or ``feeling unwell but cannot localise''.\\[2pt]
\hspace*{1em}-- \texttt{mechanism}: Mechanism-of-injury modifiers that elevate risk like ``fall from greater than twenty feet'', ``high-speed motor vehicle collision'', etc.\\[2pt]
\hspace*{1em}-- \texttt{chronic}: Chronic-history findings that modify the risk assessment like ``insulin-dependent diabetes'', ``known congestive heart failure'', etc.\\[2pt]
A single phrase belongs to at most one category, so pick the most severe applicable bucket. Return ONLY the JSON object.
\end{tcolorbox}

%% file: appendix/D-numerical-results.tex
\section{Numerical Results}
\label{app:ci}

This appendix reports the point estimates and 95\% bootstrap confidence intervals (B=10,000) underlying the plots in Section~\ref{sec:results}.

\subsection{Model performance under \textsc{sequential-vanilla}}
\label{app:ci-forced}

Table~\ref{tab:app-ci-forced} reports QWK per (model, checkpoint) on both paper corpora under the \textsc{vanilla} reasoning-off strategy. Values plotted in the \textsc{simulated} panel of Section~\ref{sec:rq1} and in Figure~\ref{fig:rq1-clin-seq}.

\begin{table*}[t!]
\centering
\begin{adjustbox}{width=0.8\linewidth}
\begin{tabular}{llccccc}
\toprule
Corpus & Model & \textit{k}=4 & \textit{k}=+25\% & \textit{k}=+50\% & \textit{k}=+75\% & \textit{k}=+100\% \\
\midrule
\multirow{6}{*}{\textsc{simulated}} & \textsc{claude} & 0.392\,{\scriptsize [0.319, 0.463]} & 0.368\,{\scriptsize [0.297, 0.440]} & 0.333\,{\scriptsize [0.259, 0.405]} & 0.344\,{\scriptsize [0.271, 0.416]} & 0.346\,{\scriptsize [0.276, 0.416]} \\
 & \textsc{gpt} & 0.316\,{\scriptsize [0.238, 0.392]} & 0.321\,{\scriptsize [0.241, 0.401]} & 0.312\,{\scriptsize [0.234, 0.388]} & 0.312\,{\scriptsize [0.236, 0.385]} & 0.332\,{\scriptsize [0.256, 0.405]} \\
 & \textsc{gemma-s} & 0.385\,{\scriptsize [0.310, 0.453]} & 0.344\,{\scriptsize [0.268, 0.416]} & 0.360\,{\scriptsize [0.288, 0.429]} & 0.323\,{\scriptsize [0.246, 0.397]} & 0.339\,{\scriptsize [0.268, 0.408]} \\
 & \textsc{gemma-l} & \cellcolor{yellow}\textbf{0.473}\,{\scriptsize [0.394, 0.546]} & \cellcolor{yellow}\textbf{0.472}\,{\scriptsize [0.399, 0.542]} & \cellcolor{yellow}\textbf{0.497}\,{\scriptsize [0.425, 0.565]} & \cellcolor{yellow}\textbf{0.463}\,{\scriptsize [0.387, 0.532]} & \cellcolor{yellow}\textbf{0.457}\,{\scriptsize [0.381, 0.530]} \\
 & \textsc{qwen-s} & 0.448\,{\scriptsize [0.369, 0.519]} & 0.375\,{\scriptsize [0.298, 0.446]} & 0.343\,{\scriptsize [0.261, 0.421]} & 0.343\,{\scriptsize [0.266, 0.416]} & 0.347\,{\scriptsize [0.271, 0.419]} \\
 & \textsc{qwen-l} & 0.461\,{\scriptsize [0.389, 0.530]} & 0.403\,{\scriptsize [0.326, 0.473]} & 0.392\,{\scriptsize [0.316, 0.464]} & 0.398\,{\scriptsize [0.326, 0.467]} & 0.391\,{\scriptsize [0.321, 0.459]} \\
\midrule
\multirow{6}{*}{\textsc{clinician}} & \textsc{claude} & 0.416\,{\scriptsize [0.219, 0.588]} & 0.500\,{\scriptsize [0.274, 0.692]} & 0.409\,{\scriptsize [0.185, 0.634]} & 0.427\,{\scriptsize [0.202, 0.650]} & 0.507\,{\scriptsize [0.328, 0.662]} \\
 & \textsc{gpt} & 0.522\,{\scriptsize [0.272, 0.735]} & 0.483\,{\scriptsize [0.238, 0.705]} & 0.465\,{\scriptsize [0.232, 0.687]} & 0.452\,{\scriptsize [0.228, 0.660]} & 0.481\,{\scriptsize [0.228, 0.707]} \\
 & \textsc{gemma-s} & 0.561\,{\scriptsize [0.360, 0.711]} & \cellcolor{yellow}\textbf{0.690}\,{\scriptsize [0.514, 0.809]} & \cellcolor{yellow}\textbf{0.567}\,{\scriptsize [0.388, 0.698]} & 0.566\,{\scriptsize [0.391, 0.699]} & 0.580\,{\scriptsize [0.409, 0.717]} \\
 & \textsc{gemma-l} & \cellcolor{yellow}\textbf{0.639}\,{\scriptsize [0.427, 0.797]} & 0.584\,{\scriptsize [0.338, 0.777]} & 0.549\,{\scriptsize [0.296, 0.757]} & \cellcolor{yellow}\textbf{0.573}\,{\scriptsize [0.310, 0.777]} & \cellcolor{yellow}\textbf{0.629}\,{\scriptsize [0.371, 0.826]} \\
 & \textsc{qwen-s} & 0.624\,{\scriptsize [0.428, 0.762]} & 0.635\,{\scriptsize [0.433, 0.779]} & 0.562\,{\scriptsize [0.362, 0.732]} & 0.558\,{\scriptsize [0.356, 0.717]} & 0.614\,{\scriptsize [0.471, 0.720]} \\
 & \textsc{qwen-l} & 0.562\,{\scriptsize [0.331, 0.736]} & 0.607\,{\scriptsize [0.436, 0.742]} & 0.558\,{\scriptsize [0.356, 0.726]} & 0.517\,{\scriptsize [0.297, 0.712]} & 0.544\,{\scriptsize [0.352, 0.701]} \\
\bottomrule
\end{tabular}
\end{adjustbox}
\caption{Performance of models on \textsc{simulated} and \textsc{clinician}, under \textsc{sequential-vanilla}. Best performing model for a corpus at a checkpoint is represented as \textbf{bold} and \colorbox{yellow}{highlighted}.}
\label{tab:app-ci-forced}
\end{table*}

\subsection{Model performance under other prompting strategies}
\label{app:ci-reason}

Table~\ref{tab:app-ci-reason} reports the QWK for all models across the four prompting strategies and both corpora, as compared in Figure~\ref{fig:rq1-reason}.

\begin{table*}[t!]
\centering
\begin{adjustbox}{max width=\linewidth}
\begin{tabular}{lllccccc}
\toprule
Corpus & Model & Strategy & \textit{k}=4 & \textit{k}=+25\% & \textit{k}=+50\% & \textit{k}=+75\% & \textit{k}=+100\% \\
\midrule
\multirow{24}{*}{\textsc{simulated}} & \multirow{4}{*}{\textsc{claude}} & \textsc{vanilla} & \cellcolor{yellow}\textbf{0.392}\,{\scriptsize [0.319, 0.463]} & \cellcolor{yellow}\textbf{0.368}\,{\scriptsize [0.297, 0.436]} & \cellcolor{yellow}\textbf{0.333}\,{\scriptsize [0.258, 0.405]} & \cellcolor{yellow}\textbf{0.344}\,{\scriptsize [0.270, 0.414]} & \cellcolor{yellow}\textbf{0.346}\,{\scriptsize [0.276, 0.415]} \\
 &  & \textsc{r-on} & 0.355\,{\scriptsize [0.284, 0.424]} & 0.311\,{\scriptsize [0.243, 0.377]} & 0.279\,{\scriptsize [0.214, 0.343]} & 0.280\,{\scriptsize [0.216, 0.346]} & 0.289\,{\scriptsize [0.228, 0.351]} \\
 &  & \textsc{ps+} & 0.339\,{\scriptsize [0.276, 0.402]} & 0.286\,{\scriptsize [0.227, 0.345]} & 0.264\,{\scriptsize [0.202, 0.326]} & 0.258\,{\scriptsize [0.199, 0.318]} & 0.263\,{\scriptsize [0.201, 0.325]} \\
 &  & \textsc{sc} & 0.365\,{\scriptsize [0.297, 0.430]} & 0.319\,{\scriptsize [0.251, 0.385]} & 0.273\,{\scriptsize [0.208, 0.338]} & 0.282\,{\scriptsize [0.220, 0.345]} & 0.287\,{\scriptsize [0.222, 0.351]} \\
\cmidrule(lr){2-8}
 & \multirow{4}{*}{\textsc{gpt}} & \textsc{vanilla} & 0.316\,{\scriptsize [0.237, 0.393]} & \cellcolor{yellow}\textbf{0.321}\,{\scriptsize [0.241, 0.397]} & 0.312\,{\scriptsize [0.234, 0.389]} & 0.312\,{\scriptsize [0.235, 0.387]} & \cellcolor{yellow}\textbf{0.332}\,{\scriptsize [0.259, 0.404]} \\
 &  & \textsc{r-on} & \cellcolor{yellow}\textbf{0.366}\,{\scriptsize [0.286, 0.444]} & 0.312\,{\scriptsize [0.234, 0.385]} & 0.314\,{\scriptsize [0.239, 0.390]} & \cellcolor{yellow}\textbf{0.333}\,{\scriptsize [0.259, 0.402]} & 0.309\,{\scriptsize [0.235, 0.381]} \\
 &  & \textsc{ps+} & 0.353\,{\scriptsize [0.274, 0.428]} & 0.287\,{\scriptsize [0.205, 0.367]} & 0.261\,{\scriptsize [0.183, 0.337]} & 0.277\,{\scriptsize [0.203, 0.349]} & 0.266\,{\scriptsize [0.192, 0.339]} \\
 &  & \textsc{sc} & 0.319\,{\scriptsize [0.238, 0.396]} & 0.306\,{\scriptsize [0.225, 0.384]} & \cellcolor{yellow}\textbf{0.316}\,{\scriptsize [0.239, 0.391]} & 0.316\,{\scriptsize [0.240, 0.388]} & 0.329\,{\scriptsize [0.254, 0.398]} \\
\cmidrule(lr){2-8}
 & \multirow{4}{*}{\textsc{gemma-s}} & \textsc{vanilla} & 0.385\,{\scriptsize [0.311, 0.455]} & 0.344\,{\scriptsize [0.271, 0.417]} & \cellcolor{yellow}\textbf{0.360}\,{\scriptsize [0.286, 0.428]} & \cellcolor{yellow}\textbf{0.323}\,{\scriptsize [0.245, 0.398]} & \cellcolor{yellow}\textbf{0.339}\,{\scriptsize [0.266, 0.407]} \\
 &  & \textsc{r-on} & 0.186\,{\scriptsize [0.117, 0.258]} & 0.181\,{\scriptsize [0.099, 0.262]} & 0.217\,{\scriptsize [0.131, 0.302]} & 0.198\,{\scriptsize [0.116, 0.277]} & 0.244\,{\scriptsize [0.165, 0.321]} \\
 &  & \textsc{ps+} & \cellcolor{yellow}\textbf{0.400}\,{\scriptsize [0.315, 0.476]} & \cellcolor{yellow}\textbf{0.351}\,{\scriptsize [0.269, 0.427]} & 0.298\,{\scriptsize [0.219, 0.373]} & 0.316\,{\scriptsize [0.241, 0.390]} & 0.323\,{\scriptsize [0.250, 0.395]} \\
 &  & \textsc{sc} & 0.183\,{\scriptsize [0.107, 0.260]} & 0.135\,{\scriptsize [0.059, 0.209]} & 0.117\,{\scriptsize [0.041, 0.193]} & 0.115\,{\scriptsize [0.039, 0.192]} & 0.159\,{\scriptsize [0.077, 0.242]} \\
\cmidrule(lr){2-8}
 & \multirow{4}{*}{\textsc{gemma-l}} & \textsc{vanilla} & \cellcolor{yellow}\textbf{0.473}\,{\scriptsize [0.392, 0.544]} & \cellcolor{yellow}\textbf{0.472}\,{\scriptsize [0.397, 0.542]} & \cellcolor{yellow}\textbf{0.497}\,{\scriptsize [0.424, 0.564]} & \cellcolor{yellow}\textbf{0.463}\,{\scriptsize [0.388, 0.534]} & \cellcolor{yellow}\textbf{0.457}\,{\scriptsize [0.383, 0.529]} \\
 &  & \textsc{r-on} & 0.424\,{\scriptsize [0.336, 0.505]} & 0.392\,{\scriptsize [0.313, 0.466]} & 0.412\,{\scriptsize [0.337, 0.483]} & 0.395\,{\scriptsize [0.320, 0.466]} & 0.393\,{\scriptsize [0.318, 0.465]} \\
 &  & \textsc{ps+} & 0.429\,{\scriptsize [0.347, 0.506]} & 0.386\,{\scriptsize [0.306, 0.461]} & 0.391\,{\scriptsize [0.316, 0.463]} & 0.388\,{\scriptsize [0.314, 0.457]} & 0.399\,{\scriptsize [0.327, 0.467]} \\
 &  & \textsc{sc} & 0.372\,{\scriptsize [0.286, 0.453]} & 0.409\,{\scriptsize [0.331, 0.483]} & 0.422\,{\scriptsize [0.346, 0.491]} & 0.404\,{\scriptsize [0.328, 0.474]} & 0.385\,{\scriptsize [0.311, 0.456]} \\
\cmidrule(lr){2-8}
 & \multirow{4}{*}{\textsc{qwen-s}} & \textsc{vanilla} & \cellcolor{yellow}\textbf{0.448}\,{\scriptsize [0.373, 0.519]} & \cellcolor{yellow}\textbf{0.375}\,{\scriptsize [0.298, 0.449]} & \cellcolor{yellow}\textbf{0.343}\,{\scriptsize [0.261, 0.419]} & 0.343\,{\scriptsize [0.268, 0.416]} & \cellcolor{yellow}\textbf{0.347}\,{\scriptsize [0.270, 0.419]} \\
 &  & \textsc{r-on} & 0.380\,{\scriptsize [0.300, 0.456]} & 0.335\,{\scriptsize [0.248, 0.418]} & 0.309\,{\scriptsize [0.228, 0.389]} & 0.295\,{\scriptsize [0.216, 0.375]} & 0.316\,{\scriptsize [0.236, 0.397]} \\
 &  & \textsc{ps+} & 0.353\,{\scriptsize [0.272, 0.429]} & 0.349\,{\scriptsize [0.263, 0.431]} & 0.314\,{\scriptsize [0.229, 0.396]} & \cellcolor{yellow}\textbf{0.360}\,{\scriptsize [0.273, 0.442]} & 0.309\,{\scriptsize [0.222, 0.396]} \\
 &  & \textsc{sc} & 0.364\,{\scriptsize [0.291, 0.432]} & 0.311\,{\scriptsize [0.238, 0.383]} & 0.309\,{\scriptsize [0.242, 0.376]} & 0.299\,{\scriptsize [0.233, 0.365]} & 0.287\,{\scriptsize [0.218, 0.356]} \\
\cmidrule(lr){2-8}
 & \multirow{4}{*}{\textsc{qwen-l}} & \textsc{vanilla} & \cellcolor{yellow}\textbf{0.461}\,{\scriptsize [0.388, 0.530]} & \cellcolor{yellow}\textbf{0.403}\,{\scriptsize [0.327, 0.475]} & \cellcolor{yellow}\textbf{0.392}\,{\scriptsize [0.316, 0.462]} & \cellcolor{yellow}\textbf{0.398}\,{\scriptsize [0.326, 0.470]} & \cellcolor{yellow}\textbf{0.391}\,{\scriptsize [0.320, 0.458]} \\
 &  & \textsc{r-on} & 0.314\,{\scriptsize [0.239, 0.387]} & 0.326\,{\scriptsize [0.248, 0.403]} & 0.313\,{\scriptsize [0.237, 0.388]} & 0.306\,{\scriptsize [0.231, 0.377]} & 0.321\,{\scriptsize [0.246, 0.397]} \\
 &  & \textsc{ps+} & 0.308\,{\scriptsize [0.228, 0.387]} & 0.286\,{\scriptsize [0.197, 0.373]} & 0.297\,{\scriptsize [0.211, 0.381]} & 0.279\,{\scriptsize [0.199, 0.357]} & 0.268\,{\scriptsize [0.186, 0.348]} \\
 &  & \textsc{sc} & 0.337\,{\scriptsize [0.264, 0.407]} & 0.292\,{\scriptsize [0.223, 0.359]} & 0.288\,{\scriptsize [0.220, 0.355]} & 0.304\,{\scriptsize [0.239, 0.368]} & 0.303\,{\scriptsize [0.236, 0.369]} \\
\midrule
\multirow{24}{*}{\textsc{clinician}} & \multirow{4}{*}{\textsc{claude}} & \textsc{vanilla} & 0.416\,{\scriptsize [0.217, 0.592]} & \cellcolor{yellow}\textbf{0.500}\,{\scriptsize [0.280, 0.689]} & \cellcolor{yellow}\textbf{0.409}\,{\scriptsize [0.179, 0.641]} & \cellcolor{yellow}\textbf{0.427}\,{\scriptsize [0.201, 0.648]} & \cellcolor{yellow}\textbf{0.507}\,{\scriptsize [0.326, 0.664]} \\
 &  & \textsc{r-on} & 0.423\,{\scriptsize [0.233, 0.588]} & 0.336\,{\scriptsize [0.146, 0.520]} & 0.317\,{\scriptsize [0.130, 0.517]} & 0.410\,{\scriptsize [0.215, 0.600]} & 0.346\,{\scriptsize [0.156, 0.539]} \\
 &  & \textsc{ps+} & 0.352\,{\scriptsize [0.150, 0.549]} & 0.317\,{\scriptsize [0.137, 0.508]} & 0.304\,{\scriptsize [0.125, 0.509]} & 0.304\,{\scriptsize [0.125, 0.503]} & 0.354\,{\scriptsize [0.170, 0.539]} \\
 &  & \textsc{sc} & \cellcolor{yellow}\textbf{0.459}\,{\scriptsize [0.257, 0.622]} & 0.340\,{\scriptsize [0.154, 0.530]} & 0.346\,{\scriptsize [0.156, 0.539]} & 0.373\,{\scriptsize [0.183, 0.564]} & 0.376\,{\scriptsize [0.191, 0.559]} \\
\cmidrule(lr){2-8}
 & \multirow{4}{*}{\textsc{gpt}} & \textsc{vanilla} & 0.522\,{\scriptsize [0.270, 0.736]} & 0.483\,{\scriptsize [0.237, 0.705]} & 0.465\,{\scriptsize [0.228, 0.683]} & 0.452\,{\scriptsize [0.232, 0.662]} & \cellcolor{yellow}\textbf{0.481}\,{\scriptsize [0.233, 0.707]} \\
 &  & \textsc{r-on} & \cellcolor{yellow}\textbf{0.692}\,{\scriptsize [0.521, 0.811]} & \cellcolor{yellow}\textbf{0.530}\,{\scriptsize [0.296, 0.736]} & 0.438\,{\scriptsize [0.217, 0.651]} & \cellcolor{yellow}\textbf{0.468}\,{\scriptsize [0.234, 0.690]} & 0.405\,{\scriptsize [0.165, 0.631]} \\
 &  & \textsc{ps+} & 0.488\,{\scriptsize [0.229, 0.710]} & 0.492\,{\scriptsize [0.247, 0.719]} & \cellcolor{yellow}\textbf{0.496}\,{\scriptsize [0.263, 0.709]} & 0.415\,{\scriptsize [0.185, 0.640]} & 0.393\,{\scriptsize [0.168, 0.618]} \\
 &  & \textsc{sc} & 0.533\,{\scriptsize [0.276, 0.739]} & 0.453\,{\scriptsize [0.221, 0.679]} & 0.468\,{\scriptsize [0.232, 0.685]} & 0.396\,{\scriptsize [0.164, 0.628]} & 0.453\,{\scriptsize [0.208, 0.678]} \\
\cmidrule(lr){2-8}
 & \multirow{4}{*}{\textsc{gemma-s}} & \textsc{vanilla} & \cellcolor{yellow}\textbf{0.561}\,{\scriptsize [0.361, 0.711]} & \cellcolor{yellow}\textbf{0.690}\,{\scriptsize [0.517, 0.810]} & \cellcolor{yellow}\textbf{0.567}\,{\scriptsize [0.390, 0.697]} & \cellcolor{yellow}\textbf{0.566}\,{\scriptsize [0.397, 0.697]} & \cellcolor{yellow}\textbf{0.580}\,{\scriptsize [0.409, 0.714]} \\
 &  & \textsc{r-on} & 0.202\,{\scriptsize [0.016, 0.405]} & 0.102\,{\scriptsize [-0.085, 0.338]} & 0.310\,{\scriptsize [0.031, 0.560]} & 0.474\,{\scriptsize [0.307, 0.626]} & 0.510\,{\scriptsize [0.351, 0.642]} \\
 &  & \textsc{ps+} & 0.433\,{\scriptsize [0.145, 0.674]} & 0.624\,{\scriptsize [0.406, 0.785]} & 0.438\,{\scriptsize [0.204, 0.646]} & 0.479\,{\scriptsize [0.253, 0.680]} & 0.435\,{\scriptsize [0.233, 0.607]} \\
 &  & \textsc{sc} & 0.245\,{\scriptsize [0.035, 0.455]} & 0.213\,{\scriptsize [0.003, 0.450]} & 0.131\,{\scriptsize [-0.083, 0.357]} & 0.104\,{\scriptsize [-0.108, 0.365]} & 0.410\,{\scriptsize [0.162, 0.636]} \\
\cmidrule(lr){2-8}
 & \multirow{4}{*}{\textsc{gemma-l}} & \textsc{vanilla} & \cellcolor{yellow}\textbf{0.639}\,{\scriptsize [0.427, 0.802]} & \cellcolor{yellow}\textbf{0.584}\,{\scriptsize [0.343, 0.779]} & \cellcolor{yellow}\textbf{0.549}\,{\scriptsize [0.281, 0.756]} & \cellcolor{yellow}\textbf{0.573}\,{\scriptsize [0.313, 0.775]} & \cellcolor{yellow}\textbf{0.629}\,{\scriptsize [0.372, 0.828]} \\
 &  & \textsc{r-on} & 0.598\,{\scriptsize [0.270, 0.838]} & 0.449\,{\scriptsize [0.191, 0.687]} & 0.401\,{\scriptsize [0.144, 0.645]} & 0.398\,{\scriptsize [0.135, 0.640]} & 0.494\,{\scriptsize [0.226, 0.717]} \\
 &  & \textsc{ps+} & 0.475\,{\scriptsize [0.202, 0.691]} & 0.491\,{\scriptsize [0.240, 0.704]} & 0.392\,{\scriptsize [0.128, 0.630]} & 0.400\,{\scriptsize [0.145, 0.638]} & 0.430\,{\scriptsize [0.181, 0.643]} \\
 &  & \textsc{sc} & 0.428\,{\scriptsize [0.142, 0.673]} & 0.392\,{\scriptsize [0.113, 0.665]} & 0.407\,{\scriptsize [0.144, 0.649]} & 0.433\,{\scriptsize [0.164, 0.679]} & 0.507\,{\scriptsize [0.246, 0.726]} \\
\cmidrule(lr){2-8}
 & \multirow{4}{*}{\textsc{qwen-s}} & \textsc{vanilla} & 0.624\,{\scriptsize [0.433, 0.761]} & \cellcolor{yellow}\textbf{0.635}\,{\scriptsize [0.434, 0.779]} & \cellcolor{yellow}\textbf{0.562}\,{\scriptsize [0.364, 0.730]} & 0.558\,{\scriptsize [0.358, 0.717]} & \cellcolor{yellow}\textbf{0.614}\,{\scriptsize [0.465, 0.721]} \\
 &  & \textsc{r-on} & 0.538\,{\scriptsize [0.332, 0.703]} & 0.393\,{\scriptsize [0.168, 0.616]} & 0.359\,{\scriptsize [0.169, 0.565]} & \cellcolor{yellow}\textbf{0.564}\,{\scriptsize [0.324, 0.749]} & 0.482\,{\scriptsize [0.237, 0.697]} \\
 &  & \textsc{ps+} & \cellcolor{yellow}\textbf{0.731}\,{\scriptsize [0.578, 0.819]} & 0.417\,{\scriptsize [0.182, 0.639]} & 0.266\,{\scriptsize [0.077, 0.496]} & 0.175\,{\scriptsize [0.004, 0.451]} & 0.257\,{\scriptsize [0.055, 0.498]} \\
 &  & \textsc{sc} & 0.534\,{\scriptsize [0.288, 0.726]} & 0.463\,{\scriptsize [0.248, 0.658]} & 0.348\,{\scriptsize [0.157, 0.556]} & 0.327\,{\scriptsize [0.133, 0.552]} & 0.392\,{\scriptsize [0.167, 0.620]} \\
\cmidrule(lr){2-8}
 & \multirow{4}{*}{\textsc{qwen-l}} & \textsc{vanilla} & \cellcolor{yellow}\textbf{0.562}\,{\scriptsize [0.332, 0.735]} & \cellcolor{yellow}\textbf{0.607}\,{\scriptsize [0.439, 0.741]} & \cellcolor{yellow}\textbf{0.558}\,{\scriptsize [0.359, 0.724]} & \cellcolor{yellow}\textbf{0.517}\,{\scriptsize [0.293, 0.708]} & \cellcolor{yellow}\textbf{0.544}\,{\scriptsize [0.350, 0.701]} \\
 &  & \textsc{r-on} & 0.389\,{\scriptsize [0.120, 0.631]} & 0.447\,{\scriptsize [0.228, 0.646]} & 0.400\,{\scriptsize [0.171, 0.612]} & 0.373\,{\scriptsize [0.146, 0.608]} & 0.431\,{\scriptsize [0.178, 0.649]} \\
 &  & \textsc{ps+} & 0.439\,{\scriptsize [0.202, 0.640]} & 0.424\,{\scriptsize [0.142, 0.699]} & 0.378\,{\scriptsize [0.100, 0.685]} & 0.355\,{\scriptsize [0.111, 0.617]} & 0.265\,{\scriptsize [0.059, 0.531]} \\
 &  & \textsc{sc} & 0.502\,{\scriptsize [0.265, 0.689]} & 0.397\,{\scriptsize [0.178, 0.601]} & 0.349\,{\scriptsize [0.125, 0.570]} & 0.350\,{\scriptsize [0.124, 0.576]} & 0.369\,{\scriptsize [0.158, 0.587]} \\
\bottomrule
\end{tabular}
\end{adjustbox}
\caption{Performance of models on \textsc{simulated} and \textsc{clinician}, across both corpora and all prompting strategies. Best performing prompt strategy for a model at a checkpoint is represented as \textbf{bold} and \colorbox{yellow}{highlighted}.}
\label{tab:app-ci-reason}
\end{table*}

\subsection{Model performance under other corpus}
\label{app:ci-plateau}

Table~\ref{tab:app-ci-plateau} reports the paired-bootstrap plateau contrast $\Delta = $ QWK@$k$=+100\% $-$ QWK@$k$=4 per model on both paper corpora. A $\Delta$ CI containing $0$ is direct evidence that more evidence beyond $k$=4 does not change QWK.

\begin{table*}[t!]
\centering
\begin{adjustbox}{width=0.7\linewidth}
\begin{tabular}{llccc}
\toprule
Corpus & Model & QWK @ $k$=4 & QWK @ $k$=+100\% & $\Delta$ (95\% CI) \\
\midrule
\multirow{6}{*}{\textsc{simulated}} & \textsc{claude} & \cellcolor{yellow}\textbf{0.392}\,{\scriptsize [0.318, 0.462]} & 0.346\,{\scriptsize [0.276, 0.416]} & -0.046\,{\scriptsize [-0.110, +0.015]} \\
 & \textsc{gpt} & 0.316\,{\scriptsize [0.237, 0.393]} & \cellcolor{yellow}\textbf{0.329}\,{\scriptsize [0.252, 0.401]} & +0.013\,{\scriptsize [-0.057, +0.080]} \\
 & \textsc{gemma-s} & \cellcolor{yellow}\textbf{0.385}\,{\scriptsize [0.309, 0.454]} & 0.339\,{\scriptsize [0.267, 0.410]} & -0.046\,{\scriptsize [-0.107, +0.016]} \\
 & \textsc{gemma-l} & \cellcolor{yellow}\textbf{0.473}\,{\scriptsize [0.394, 0.549]} & 0.457\,{\scriptsize [0.384, 0.527]} & -0.015\,{\scriptsize [-0.075, +0.042]} \\
 & \textsc{qwen-s} & \cellcolor{yellow}\textbf{0.448}\,{\scriptsize [0.370, 0.519]} & 0.347\,{\scriptsize [0.272, 0.417]} & \textbf{-0.101}\,{\scriptsize [-0.163, -0.039]} \\
 & \textsc{qwen-l} & \cellcolor{yellow}\textbf{0.461}\,{\scriptsize [0.387, 0.530]} & 0.391\,{\scriptsize [0.321, 0.457]} & \textbf{-0.070}\,{\scriptsize [-0.132, -0.006]} \\
\midrule
\multirow{6}{*}{\textsc{clinician}} & \textsc{claude} & 0.416\,{\scriptsize [0.216, 0.592]} & \cellcolor{yellow}\textbf{0.507}\,{\scriptsize [0.326, 0.669]} & +0.090\,{\scriptsize [-0.061, +0.243]} \\
 & \textsc{gpt} & \cellcolor{yellow}\textbf{0.522}\,{\scriptsize [0.272, 0.734]} & 0.481\,{\scriptsize [0.227, 0.708]} & -0.042\,{\scriptsize [-0.290, +0.186]} \\
 & \textsc{gemma-s} & 0.561\,{\scriptsize [0.356, 0.713]} & \cellcolor{yellow}\textbf{0.580}\,{\scriptsize [0.408, 0.716]} & +0.019\,{\scriptsize [-0.140, +0.185]} \\
 & \textsc{gemma-l} & \cellcolor{yellow}\textbf{0.639}\,{\scriptsize [0.430, 0.803]} & 0.629\,{\scriptsize [0.366, 0.826]} & -0.010\,{\scriptsize [-0.260, +0.231]} \\
 & \textsc{qwen-s} & \cellcolor{yellow}\textbf{0.624}\,{\scriptsize [0.428, 0.761]} & 0.614\,{\scriptsize [0.474, 0.724]} & -0.010\,{\scriptsize [-0.172, +0.180]} \\
 & \textsc{qwen-l} & \cellcolor{yellow}\textbf{0.562}\,{\scriptsize [0.333, 0.739]} & 0.544\,{\scriptsize [0.349, 0.700]} & -0.018\,{\scriptsize [-0.243, +0.244]} \\
\bottomrule
\end{tabular}
\end{adjustbox}
\caption{Change in performance between the first and last checkpoint on \textsc{simulated} and \textsc{clinician}, under \textsc{sequential-vanilla}. Better performing checkpoint for a model is represented as \textbf{bold} and \colorbox{yellow}{highlighted}; $\Delta$ is \textbf{bold} when its 95\% CI excludes zero.}
\label{tab:app-ci-plateau}
\end{table*}

\subsection{Impact of input perturbation}
\label{app:ci-pert}

Table~\ref{tab:app-ci-pert} reports $\Delta$~QWK relative to the unperturbed \textsc{vanilla} baseline for the six perturbations described in Section~\ref{sec:rq2}. Values plotted in Figure~\ref{fig:rq2-abl} on \textsc{simulated} and Figure~\ref{fig:app-abl-clin} on \textsc{clinician}. Sign is (perturbed $-$ vanilla): a negative cell means the perturbation hurt.

\begin{table*}[t!]
\centering
\small
\begin{adjustbox}{max width=\linewidth}
\begin{tabular}{lllccccc}
\toprule
Corpus & Model & Perturbation & \textit{k}=4 & \textit{k}=+25\% & \textit{k}=+50\% & \textit{k}=+75\% & \textit{k}=+100\% \\
\midrule
\multirow{36}{*}{\textsc{simulated}} & \multirow{6}{*}{\textsc{claude}} & \textsc{jumbled} & \cellcolor{yellow}\textbf{-0.124}\,{\scriptsize [-0.190, -0.059]} & -0.066\,{\scriptsize [-0.116, -0.015]} & -0.013\,{\scriptsize [-0.058, 0.034]} & -0.021\,{\scriptsize [-0.062, 0.021]} & -0.025\,{\scriptsize [-0.059, 0.007]} \\
 &  & \textsc{cc-removed} & \cellcolor{yellow}\textbf{-0.069}\,{\scriptsize [-0.137, -0.002]} & -0.062\,{\scriptsize [-0.113, -0.013]} & -0.012\,{\scriptsize [-0.050, 0.026]} & -0.020\,{\scriptsize [-0.060, 0.019]} & -0.019\,{\scriptsize [-0.050, 0.012]} \\
 &  & \textsc{cc-delayed} & $-$ & $-$ & $-$ & $-$ & -0.018\,{\scriptsize [-0.049, 0.011]} \\
 &  & \textsc{removed-i} & -0.002\,{\scriptsize [-0.019, 0.014]} & -0.004\,{\scriptsize [-0.036, 0.029]} & 0.014\,{\scriptsize [-0.014, 0.042]} & 0.020\,{\scriptsize [-0.007, 0.048]} & \cellcolor{yellow}\textbf{-0.013}\,{\scriptsize [-0.039, 0.012]} \\
 &  & \textsc{removed-ii} & 0.009\,{\scriptsize [-0.005, 0.024]} & -0.017\,{\scriptsize [-0.068, 0.035]} & -0.006\,{\scriptsize [-0.045, 0.033]} & \cellcolor{yellow}\textbf{-0.020}\,{\scriptsize [-0.053, 0.013]} & 0.006\,{\scriptsize [-0.020, 0.033]} \\
 &  & \textsc{removed-iii} & 0.005\,{\scriptsize [-0.013, 0.024]} & -0.011\,{\scriptsize [-0.065, 0.042]} & -0.004\,{\scriptsize [-0.045, 0.037]} & \cellcolor{yellow}\textbf{-0.032}\,{\scriptsize [-0.066, 0.001]} & -0.011\,{\scriptsize [-0.040, 0.017]} \\
\cmidrule(lr){2-8}
 & \multirow{6}{*}{\textsc{gpt}} & \textsc{jumbled} & \cellcolor{yellow}\textbf{-0.092}\,{\scriptsize [-0.177, -0.005]} & -0.027\,{\scriptsize [-0.092, 0.039]} & -0.003\,{\scriptsize [-0.059, 0.053]} & 0.012\,{\scriptsize [-0.039, 0.066]} & -0.009\,{\scriptsize [-0.050, 0.033]} \\
 &  & \textsc{cc-removed} & -0.026\,{\scriptsize [-0.107, 0.055]} & \cellcolor{yellow}\textbf{-0.031}\,{\scriptsize [-0.091, 0.031]} & 0.017\,{\scriptsize [-0.033, 0.068]} & -0.003\,{\scriptsize [-0.060, 0.055]} & -0.022\,{\scriptsize [-0.065, 0.022]} \\
 &  & \textsc{cc-delayed} & $-$ & $-$ & $-$ & $-$ & -0.003\,{\scriptsize [-0.043, 0.036]} \\
 &  & \textsc{removed-i} & 0.019\,{\scriptsize [-0.023, 0.063]} & 0.004\,{\scriptsize [-0.048, 0.055]} & 0.004\,{\scriptsize [-0.035, 0.043]} & 0.007\,{\scriptsize [-0.037, 0.050]} & \cellcolor{yellow}\textbf{-0.023}\,{\scriptsize [-0.057, 0.012]} \\
 &  & \textsc{removed-ii} & 0.020\,{\scriptsize [-0.022, 0.064]} & \cellcolor{yellow}\textbf{-0.005}\,{\scriptsize [-0.060, 0.050]} & 0.001\,{\scriptsize [-0.049, 0.050]} & \cellcolor{yellow}\textbf{-0.005}\,{\scriptsize [-0.046, 0.036]} & \cellcolor{yellow}\textbf{-0.005}\,{\scriptsize [-0.040, 0.030]} \\
 &  & \textsc{removed-iii} & -0.000\,{\scriptsize [-0.045, 0.045]} & -0.019\,{\scriptsize [-0.085, 0.048]} & \cellcolor{yellow}\textbf{-0.025}\,{\scriptsize [-0.079, 0.027]} & 0.017\,{\scriptsize [-0.032, 0.067]} & 0.004\,{\scriptsize [-0.035, 0.042]} \\
\cmidrule(lr){2-8}
 & \multirow{6}{*}{\textsc{gemma-s}} & \textsc{jumbled} & \cellcolor{yellow}\textbf{-0.253}\,{\scriptsize [-0.338, -0.167]} & -0.102\,{\scriptsize [-0.169, -0.036]} & -0.083\,{\scriptsize [-0.137, -0.030]} & -0.029\,{\scriptsize [-0.081, 0.025]} & -0.016\,{\scriptsize [-0.057, 0.025]} \\
 &  & \textsc{cc-removed} & \cellcolor{yellow}\textbf{-0.206}\,{\scriptsize [-0.295, -0.116]} & -0.106\,{\scriptsize [-0.168, -0.045]} & -0.120\,{\scriptsize [-0.174, -0.066]} & -0.096\,{\scriptsize [-0.146, -0.048]} & -0.089\,{\scriptsize [-0.133, -0.046]} \\
 &  & \textsc{cc-delayed} & $-$ & $-$ & $-$ & $-$ & -0.034\,{\scriptsize [-0.074, 0.006]} \\
 &  & \textsc{removed-i} & 0.000\,{\scriptsize [0.000, 0.000]} & 0.014\,{\scriptsize [-0.020, 0.049]} & -0.002\,{\scriptsize [-0.033, 0.029]} & 0.016\,{\scriptsize [-0.010, 0.042]} & \cellcolor{yellow}\textbf{-0.009}\,{\scriptsize [-0.028, 0.009]} \\
 &  & \textsc{removed-ii} & 0.000\,{\scriptsize [0.000, 0.000]} & 0.013\,{\scriptsize [-0.028, 0.053]} & \cellcolor{yellow}\textbf{-0.013}\,{\scriptsize [-0.058, 0.034]} & 0.034\,{\scriptsize [-0.003, 0.073]} & 0.012\,{\scriptsize [-0.028, 0.052]} \\
 &  & \textsc{removed-iii} & \cellcolor{yellow}\textbf{-0.004}\,{\scriptsize [-0.013, 0.001]} & 0.018\,{\scriptsize [-0.033, 0.072]} & 0.002\,{\scriptsize [-0.048, 0.052]} & 0.021\,{\scriptsize [-0.019, 0.064]} & 0.007\,{\scriptsize [-0.035, 0.049]} \\
\cmidrule(lr){2-8}
 & \multirow{6}{*}{\textsc{gemma-l}} & \textsc{jumbled} & \cellcolor{yellow}\textbf{-0.153}\,{\scriptsize [-0.243, -0.064]} & -0.070\,{\scriptsize [-0.129, -0.012]} & -0.084\,{\scriptsize [-0.129, -0.040]} & -0.021\,{\scriptsize [-0.057, 0.016]} & -0.001\,{\scriptsize [-0.029, 0.028]} \\
 &  & \textsc{cc-removed} & \cellcolor{yellow}\textbf{-0.085}\,{\scriptsize [-0.158, -0.013]} & -0.013\,{\scriptsize [-0.069, 0.041]} & -0.060\,{\scriptsize [-0.101, -0.020]} & -0.018\,{\scriptsize [-0.053, 0.018]} & -0.010\,{\scriptsize [-0.044, 0.024]} \\
 &  & \textsc{cc-delayed} & $-$ & $-$ & $-$ & $-$ & 0.013\,{\scriptsize [-0.012, 0.041]} \\
 &  & \textsc{removed-i} & -0.003\,{\scriptsize [-0.010, 0.002]} & \cellcolor{yellow}\textbf{-0.016}\,{\scriptsize [-0.047, 0.016]} & -0.010\,{\scriptsize [-0.031, 0.010]} & -0.005\,{\scriptsize [-0.023, 0.013]} & 0.004\,{\scriptsize [-0.010, 0.019]} \\
 &  & \textsc{removed-ii} & 0.000\,{\scriptsize [0.000, 0.000]} & 0.004\,{\scriptsize [-0.041, 0.049]} & \cellcolor{yellow}\textbf{-0.025}\,{\scriptsize [-0.061, 0.010]} & -0.013\,{\scriptsize [-0.044, 0.019]} & 0.005\,{\scriptsize [-0.020, 0.030]} \\
 &  & \textsc{removed-iii} & 0.000\,{\scriptsize [0.000, 0.000]} & 0.006\,{\scriptsize [-0.043, 0.058]} & \cellcolor{yellow}\textbf{-0.054}\,{\scriptsize [-0.098, -0.011]} & -0.017\,{\scriptsize [-0.050, 0.016]} & 0.002\,{\scriptsize [-0.024, 0.027]} \\
\cmidrule(lr){2-8}
 & \multirow{6}{*}{\textsc{qwen-s}} & \textsc{jumbled} & \cellcolor{yellow}\textbf{-0.219}\,{\scriptsize [-0.310, -0.132]} & -0.065\,{\scriptsize [-0.133, 0.003]} & -0.026\,{\scriptsize [-0.086, 0.035]} & -0.015\,{\scriptsize [-0.065, 0.037]} & 0.018\,{\scriptsize [-0.021, 0.057]} \\
 &  & \textsc{cc-removed} & \cellcolor{yellow}\textbf{-0.169}\,{\scriptsize [-0.252, -0.085]} & -0.076\,{\scriptsize [-0.146, -0.007]} & -0.055\,{\scriptsize [-0.110, 0.001]} & -0.050\,{\scriptsize [-0.096, -0.003]} & -0.062\,{\scriptsize [-0.111, -0.015]} \\
 &  & \textsc{cc-delayed} & $-$ & $-$ & $-$ & $-$ & -0.026\,{\scriptsize [-0.069, 0.015]} \\
 &  & \textsc{removed-i} & \cellcolor{yellow}\textbf{-0.002}\,{\scriptsize [-0.007, 0.001]} & \cellcolor{yellow}\textbf{-0.002}\,{\scriptsize [-0.038, 0.032]} & 0.020\,{\scriptsize [-0.012, 0.052]} & 0.006\,{\scriptsize [-0.021, 0.034]} & 0.010\,{\scriptsize [-0.007, 0.029]} \\
 &  & \textsc{removed-ii} & -0.002\,{\scriptsize [-0.007, 0.001]} & -0.001\,{\scriptsize [-0.056, 0.054]} & \cellcolor{yellow}\textbf{-0.010}\,{\scriptsize [-0.061, 0.039]} & 0.019\,{\scriptsize [-0.025, 0.064]} & -0.000\,{\scriptsize [-0.033, 0.032]} \\
 &  & \textsc{removed-iii} & -0.002\,{\scriptsize [-0.007, 0.001]} & 0.015\,{\scriptsize [-0.045, 0.075]} & 0.015\,{\scriptsize [-0.040, 0.070]} & 0.020\,{\scriptsize [-0.025, 0.065]} & \cellcolor{yellow}\textbf{-0.008}\,{\scriptsize [-0.042, 0.026]} \\
\cmidrule(lr){2-8}
 & \multirow{6}{*}{\textsc{qwen-l}} & \textsc{jumbled} & \cellcolor{yellow}\textbf{-0.163}\,{\scriptsize [-0.236, -0.090]} & -0.060\,{\scriptsize [-0.115, -0.004]} & -0.032\,{\scriptsize [-0.076, 0.012]} & -0.048\,{\scriptsize [-0.087, -0.009]} & -0.016\,{\scriptsize [-0.049, 0.017]} \\
 &  & \textsc{cc-removed} & \cellcolor{yellow}\textbf{-0.126}\,{\scriptsize [-0.199, -0.054]} & -0.068\,{\scriptsize [-0.126, -0.010]} & -0.069\,{\scriptsize [-0.112, -0.027]} & -0.059\,{\scriptsize [-0.105, -0.014]} & -0.040\,{\scriptsize [-0.070, -0.012]} \\
 &  & \textsc{cc-delayed} & $-$ & $-$ & $-$ & $-$ & -0.020\,{\scriptsize [-0.051, 0.011]} \\
 &  & \textsc{removed-i} & 0.000\,{\scriptsize [0.000, 0.000]} & 0.002\,{\scriptsize [-0.029, 0.032]} & \cellcolor{yellow}\textbf{-0.020}\,{\scriptsize [-0.043, 0.003]} & 0.000\,{\scriptsize [-0.019, 0.021]} & -0.002\,{\scriptsize [-0.019, 0.016]} \\
 &  & \textsc{removed-ii} & 0.000\,{\scriptsize [0.000, 0.000]} & -0.020\,{\scriptsize [-0.066, 0.026]} & \cellcolor{yellow}\textbf{-0.048}\,{\scriptsize [-0.088, -0.009]} & -0.006\,{\scriptsize [-0.034, 0.023]} & -0.003\,{\scriptsize [-0.029, 0.023]} \\
 &  & \textsc{removed-iii} & 0.000\,{\scriptsize [0.000, 0.000]} & -0.035\,{\scriptsize [-0.087, 0.020]} & \cellcolor{yellow}\textbf{-0.049}\,{\scriptsize [-0.096, -0.002]} & -0.007\,{\scriptsize [-0.040, 0.027]} & -0.006\,{\scriptsize [-0.035, 0.023]} \\
\midrule
\multirow{36}{*}{\textsc{clinician}} & \multirow{6}{*}{\textsc{claude}} & \textsc{jumbled} & 0.052\,{\scriptsize [-0.121, 0.222]} & -0.011\,{\scriptsize [-0.158, 0.154]} & 0.069\,{\scriptsize [-0.015, 0.156]} & -0.040\,{\scriptsize [-0.090, 0.006]} & \cellcolor{yellow}\textbf{-0.074}\,{\scriptsize [-0.226, 0.077]} \\
 &  & \textsc{cc-removed} & 0.024\,{\scriptsize [-0.200, 0.263]} & \cellcolor{yellow}\textbf{-0.103}\,{\scriptsize [-0.232, 0.050]} & 0.063\,{\scriptsize [-0.044, 0.182]} & 0.037\,{\scriptsize [-0.042, 0.125]} & -0.074\,{\scriptsize [-0.218, 0.072]} \\
 &  & \textsc{cc-delayed} & $-$ & $-$ & $-$ & $-$ & -0.030\,{\scriptsize [-0.154, 0.096]} \\
 &  & \textsc{removed-i} & -0.004\,{\scriptsize [-0.115, 0.116]} & \cellcolor{yellow}\textbf{-0.062}\,{\scriptsize [-0.148, 0.025]} & 0.051\,{\scriptsize [-0.026, 0.147]} & 0.046\,{\scriptsize [-0.025, 0.138]} & 0.017\,{\scriptsize [-0.098, 0.131]} \\
 &  & \textsc{removed-ii} & \cellcolor{yellow}\textbf{-0.045}\,{\scriptsize [-0.134, 0.002]} & -0.040\,{\scriptsize [-0.169, 0.089]} & 0.124\,{\scriptsize [-0.034, 0.286]} & 0.116\,{\scriptsize [-0.008, 0.231]} & 0.063\,{\scriptsize [-0.011, 0.134]} \\
 &  & \textsc{removed-iii} & -0.007\,{\scriptsize [-0.119, 0.104]} & \cellcolor{yellow}\textbf{-0.068}\,{\scriptsize [-0.214, 0.072]} & 0.046\,{\scriptsize [-0.121, 0.198]} & 0.004\,{\scriptsize [-0.175, 0.152]} & -0.028\,{\scriptsize [-0.173, 0.088]} \\
\cmidrule(lr){2-8}
 & \multirow{6}{*}{\textsc{gpt}} & \textsc{jumbled} & -0.140\,{\scriptsize [-0.491, 0.186]} & \cellcolor{yellow}\textbf{-0.151}\,{\scriptsize [-0.307, 0.003]} & -0.107\,{\scriptsize [-0.254, 0.023]} & -0.102\,{\scriptsize [-0.238, 0.035]} & -0.089\,{\scriptsize [-0.189, -0.004]} \\
 &  & \textsc{cc-removed} & 0.047\,{\scriptsize [-0.178, 0.271]} & 0.012\,{\scriptsize [-0.141, 0.137]} & \cellcolor{yellow}\textbf{-0.076}\,{\scriptsize [-0.242, 0.072]} & 0.006\,{\scriptsize [-0.116, 0.130]} & -0.026\,{\scriptsize [-0.140, 0.098]} \\
 &  & \textsc{cc-delayed} & $-$ & $-$ & $-$ & $-$ & -0.092\,{\scriptsize [-0.186, 0.000]} \\
 &  & \textsc{removed-i} & 0.084\,{\scriptsize [-0.052, 0.218]} & 0.031\,{\scriptsize [-0.107, 0.172]} & 0.021\,{\scriptsize [-0.127, 0.162]} & 0.030\,{\scriptsize [-0.051, 0.124]} & \cellcolor{yellow}\textbf{0.012}\,{\scriptsize [-0.065, 0.095]} \\
 &  & \textsc{removed-ii} & -0.051\,{\scriptsize [-0.223, 0.113]} & 0.081\,{\scriptsize [-0.121, 0.287]} & 0.022\,{\scriptsize [-0.166, 0.231]} & \cellcolor{yellow}\textbf{-0.076}\,{\scriptsize [-0.247, 0.136]} & -0.032\,{\scriptsize [-0.171, 0.158]} \\
 &  & \textsc{removed-iii} & 0.063\,{\scriptsize [-0.103, 0.231]} & \cellcolor{yellow}\textbf{-0.043}\,{\scriptsize [-0.288, 0.227]} & 0.012\,{\scriptsize [-0.232, 0.262]} & -0.017\,{\scriptsize [-0.205, 0.195]} & -0.024\,{\scriptsize [-0.190, 0.158]} \\
\cmidrule(lr){2-8}
 & \multirow{6}{*}{\textsc{gemma-s}} & \textsc{jumbled} & \cellcolor{yellow}\textbf{-0.558}\,{\scriptsize [-0.784, -0.290]} & -0.360\,{\scriptsize [-0.568, -0.175]} & -0.096\,{\scriptsize [-0.249, 0.037]} & -0.086\,{\scriptsize [-0.203, -0.001]} & -0.031\,{\scriptsize [-0.089, 0.012]} \\
 &  & \textsc{cc-removed} & -0.012\,{\scriptsize [-0.200, 0.175]} & \cellcolor{yellow}\textbf{-0.122}\,{\scriptsize [-0.279, 0.021]} & -0.019\,{\scriptsize [-0.175, 0.127]} & -0.063\,{\scriptsize [-0.171, 0.016]} & -0.041\,{\scriptsize [-0.165, 0.065]} \\
 &  & \textsc{cc-delayed} & $-$ & $-$ & $-$ & $-$ & -0.053\,{\scriptsize [-0.128, -0.007]} \\
 &  & \textsc{removed-i} & 0.000\,{\scriptsize [0.000, 0.000]} & \cellcolor{yellow}\textbf{-0.113}\,{\scriptsize [-0.212, -0.031]} & -0.015\,{\scriptsize [-0.112, 0.075]} & -0.042\,{\scriptsize [-0.149, 0.044]} & -0.047\,{\scriptsize [-0.139, 0.002]} \\
 &  & \textsc{removed-ii} & 0.000\,{\scriptsize [0.000, 0.000]} & -0.060\,{\scriptsize [-0.171, 0.034]} & -0.055\,{\scriptsize [-0.211, 0.079]} & \cellcolor{yellow}\textbf{-0.111}\,{\scriptsize [-0.230, -0.014]} & -0.085\,{\scriptsize [-0.221, 0.039]} \\
 &  & \textsc{removed-iii} & 0.000\,{\scriptsize [0.000, 0.000]} & \cellcolor{yellow}\textbf{-0.300}\,{\scriptsize [-0.481, -0.128]} & -0.061\,{\scriptsize [-0.217, 0.106]} & -0.188\,{\scriptsize [-0.313, -0.058]} & -0.107\,{\scriptsize [-0.244, 0.032]} \\
\cmidrule(lr){2-8}
 & \multirow{6}{*}{\textsc{gemma-l}} & \textsc{jumbled} & \cellcolor{yellow}\textbf{-0.228}\,{\scriptsize [-0.532, 0.070]} & -0.063\,{\scriptsize [-0.233, 0.115]} & -0.075\,{\scriptsize [-0.269, 0.098]} & -0.059\,{\scriptsize [-0.233, 0.064]} & -0.021\,{\scriptsize [-0.146, 0.102]} \\
 &  & \textsc{cc-removed} & \cellcolor{yellow}\textbf{-0.205}\,{\scriptsize [-0.473, 0.042]} & -0.069\,{\scriptsize [-0.285, 0.133]} & -0.126\,{\scriptsize [-0.385, 0.118]} & -0.065\,{\scriptsize [-0.280, 0.125]} & 0.000\,{\scriptsize [-0.117, 0.125]} \\
 &  & \textsc{cc-delayed} & $-$ & $-$ & $-$ & $-$ & -0.015\,{\scriptsize [-0.132, 0.101]} \\
 &  & \textsc{removed-i} & 0.001\,{\scriptsize [-0.008, 0.010]} & 0.037\,{\scriptsize [-0.118, 0.199]} & 0.047\,{\scriptsize [-0.073, 0.182]} & 0.012\,{\scriptsize [-0.140, 0.161]} & \cellcolor{yellow}\textbf{0.000}\,{\scriptsize [-0.036, 0.037]} \\
 &  & \textsc{removed-ii} & 0.001\,{\scriptsize [-0.008, 0.009]} & -0.026\,{\scriptsize [-0.222, 0.163]} & -0.014\,{\scriptsize [-0.119, 0.082]} & -0.064\,{\scriptsize [-0.199, 0.040]} & \cellcolor{yellow}\textbf{-0.098}\,{\scriptsize [-0.222, -0.001]} \\
 &  & \textsc{removed-iii} & 0.001\,{\scriptsize [-0.008, 0.009]} & -0.048\,{\scriptsize [-0.224, 0.125]} & 0.021\,{\scriptsize [-0.132, 0.168]} & \cellcolor{yellow}\textbf{-0.071}\,{\scriptsize [-0.207, 0.032]} & -0.006\,{\scriptsize [-0.101, 0.086]} \\
\cmidrule(lr){2-8}
 & \multirow{6}{*}{\textsc{qwen-s}} & \textsc{jumbled} & \cellcolor{yellow}\textbf{-0.570}\,{\scriptsize [-0.871, -0.245]} & -0.235\,{\scriptsize [-0.455, -0.028]} & -0.069\,{\scriptsize [-0.193, 0.031]} & -0.047\,{\scriptsize [-0.221, 0.130]} & -0.023\,{\scriptsize [-0.142, 0.074]} \\
 &  & \textsc{cc-removed} & \cellcolor{yellow}\textbf{-0.120}\,{\scriptsize [-0.379, 0.136]} & -0.033\,{\scriptsize [-0.147, 0.069]} & -0.035\,{\scriptsize [-0.190, 0.099]} & -0.057\,{\scriptsize [-0.215, 0.082]} & -0.116\,{\scriptsize [-0.278, 0.025]} \\
 &  & \textsc{cc-delayed} & $-$ & $-$ & $-$ & $-$ & -0.001\,{\scriptsize [-0.107, 0.086]} \\
 &  & \textsc{removed-i} & 0.000\,{\scriptsize [0.000, 0.000]} & \cellcolor{yellow}\textbf{-0.038}\,{\scriptsize [-0.168, 0.095]} & 0.009\,{\scriptsize [-0.138, 0.151]} & 0.020\,{\scriptsize [-0.030, 0.093]} & -0.025\,{\scriptsize [-0.129, 0.052]} \\
 &  & \textsc{removed-ii} & 0.000\,{\scriptsize [0.000, 0.000]} & \cellcolor{yellow}\textbf{-0.041}\,{\scriptsize [-0.145, 0.045]} & -0.006\,{\scriptsize [-0.152, 0.129]} & 0.008\,{\scriptsize [-0.081, 0.101]} & 0.038\,{\scriptsize [0.000, 0.109]} \\
 &  & \textsc{removed-iii} & 0.000\,{\scriptsize [0.000, 0.000]} & \cellcolor{yellow}\textbf{-0.218}\,{\scriptsize [-0.432, -0.030]} & -0.044\,{\scriptsize [-0.205, 0.099]} & -0.020\,{\scriptsize [-0.081, 0.101]} & -0.041\,{\scriptsize [-0.116, 0.008]} \\
\cmidrule(lr){2-8}
 & \multirow{6}{*}{\textsc{qwen-l}} & \textsc{jumbled} & \cellcolor{yellow}\textbf{-0.246}\,{\scriptsize [-0.564, 0.058]} & -0.181\,{\scriptsize [-0.324, -0.033]} & -0.085\,{\scriptsize [-0.235, 0.065]} & -0.056\,{\scriptsize [-0.177, 0.057]} & -0.035\,{\scriptsize [-0.167, 0.093]} \\
 &  & \textsc{cc-removed} & -0.036\,{\scriptsize [-0.291, 0.210]} & -0.027\,{\scriptsize [-0.177, 0.110]} & \cellcolor{yellow}\textbf{-0.048}\,{\scriptsize [-0.225, 0.104]} & 0.017\,{\scriptsize [-0.121, 0.131]} & -0.000\,{\scriptsize [-0.135, 0.128]} \\
 &  & \textsc{cc-delayed} & $-$ & $-$ & $-$ & $-$ & 0.075\,{\scriptsize [-0.006, 0.176]} \\
 &  & \textsc{removed-i} & 0.000\,{\scriptsize [0.000, 0.000]} & -0.089\,{\scriptsize [-0.287, 0.075]} & \cellcolor{yellow}\textbf{-0.107}\,{\scriptsize [-0.218, -0.012]} & 0.055\,{\scriptsize [-0.006, 0.147]} & 0.024\,{\scriptsize [-0.020, 0.095]} \\
 &  & \textsc{removed-ii} & 0.000\,{\scriptsize [0.000, 0.000]} & \cellcolor{yellow}\textbf{-0.044}\,{\scriptsize [-0.218, 0.123]} & 0.051\,{\scriptsize [-0.135, 0.222]} & 0.054\,{\scriptsize [-0.044, 0.168]} & 0.007\,{\scriptsize [-0.068, 0.092]} \\
 &  & \textsc{removed-iii} & 0.000\,{\scriptsize [0.000, 0.000]} & \cellcolor{yellow}\textbf{-0.148}\,{\scriptsize [-0.334, 0.027]} & 0.006\,{\scriptsize [-0.194, 0.193]} & 0.022\,{\scriptsize [-0.111, 0.159]} & 0.065\,{\scriptsize [-0.056, 0.182]} \\
\bottomrule
\end{tabular}
\end{adjustbox}
\caption{Change in performance under each perturbation on \textsc{simulated} and \textsc{clinician}, under \textsc{sequential-vanilla}. Checkpoint at which a perturbation is most harmful for a model is represented as \textbf{bold} and \colorbox{yellow}{highlighted}.}
\label{tab:app-ci-pert}
\end{table*}

\subsection{Six-model IAA per checkpoint}
\label{app:ci-iaa}

Table~\ref{tab:app-ci-iaa} reports Krippendorff $\alpha$ and the mean of QWK across the fifteen pairs of the six-model roster, computed on the six-model intersection at each checkpoint. Values plotted in Figure~\ref{fig:rq1-iaa}.

\begin{table}[t!]
\centering
\begin{adjustbox}{max width=\linewidth}
\begin{tabular}{llcc}
\toprule
Corpus & Checkpoint & Krippendorff $\alpha$ & Mean pairwise QWK \\
\midrule
\multirow{5}{*}{\textsc{simulated}} & \textit{k}=4 & \cellcolor{yellow}\textbf{0.690}\,{\scriptsize [0.648, 0.724]} & 0.714\,{\scriptsize [0.674, 0.746]} \\
 & \textit{k}=+25\% & 0.688\,{\scriptsize [0.647, 0.727]} & 0.728\,{\scriptsize [0.689, 0.758]} \\
 & \textit{k}=+50\% & 0.674\,{\scriptsize [0.630, 0.713]} & 0.708\,{\scriptsize [0.664, 0.746]} \\
 & \textit{k}=+75\% & 0.677\,{\scriptsize [0.631, 0.716]} & 0.721\,{\scriptsize [0.678, 0.758]} \\
 & \textit{k}=+100\% & \cellcolor{yellow}\textbf{0.690}\,{\scriptsize [0.649, 0.729]} & \cellcolor{yellow}\textbf{0.733}\,{\scriptsize [0.692, 0.770]} \\
\midrule
\multirow{5}{*}{\textsc{clinician}} & \textit{k}=4 & 0.604\,{\scriptsize [0.444, 0.725]} & 0.584\,{\scriptsize [0.422, 0.703]} \\
 & \textit{k}=+25\% & 0.639\,{\scriptsize [0.458, 0.757]} & 0.624\,{\scriptsize [0.456, 0.739]} \\
 & \textit{k}=+50\% & 0.618\,{\scriptsize [0.452, 0.739]} & 0.619\,{\scriptsize [0.469, 0.728]} \\
 & \textit{k}=+75\% & 0.684\,{\scriptsize [0.535, 0.791]} & 0.679\,{\scriptsize [0.545, 0.778]} \\
 & \textit{k}=+100\% & \cellcolor{yellow}\textbf{0.702}\,{\scriptsize [0.560, 0.801]} & \cellcolor{yellow}\textbf{0.698}\,{\scriptsize [0.577, 0.785]} \\
\bottomrule
\end{tabular}
\end{adjustbox}
\caption{Inter-annotator agreement across the six models at each checkpoint on \textsc{simulated} and \textsc{clinician}, under \textsc{sequential-vanilla}. Checkpoint with the highest agreement for a corpus and metric is represented as \textbf{bold} and \colorbox{yellow}{highlighted}.}
\label{tab:app-ci-iaa}
\end{table}

\subsection{Commit-rate distribution}
\label{app:ci-commit}

Table~\ref{tab:app-ci-commit} reports the fraction of committed conversations for each model, when evaluated under the conditions described in Section~\ref{sec:rq3}. Values plotted in Figure~\ref{fig:rq3-commit-sim} on \textsc{simulated} and Figure~\ref{fig:app-rq3-clinician} on \textsc{clinician}.

\begin{table}[t!]
\centering
\begin{adjustbox}{max width=\linewidth}
\begin{tabular}{llccccc}
\toprule
Corpus & Model & \textit{k}=4 & \textit{k}=+25\% & \textit{k}=+50\% & \textit{k}=+75\% & \textit{k}=+100\% \\
\midrule
\multirow{6}{*}{\textsc{simulated}} & \textsc{claude} & 0.078 & 0.284 & \cellcolor{yellow}\textbf{0.323} & 0.191 & 0.124 \\
 & \textsc{gpt} & 0.225 & \cellcolor{yellow}\textbf{0.325} & 0.246 & 0.127 & 0.077 \\
 & \textsc{gemma-s} & \cellcolor{yellow}\textbf{0.385} & 0.356 & 0.150 & 0.073 & 0.036 \\
 & \textsc{gemma-l} & 0.224 & \cellcolor{yellow}\textbf{0.305} & 0.224 & 0.134 & 0.112 \\
 & \textsc{qwen-s} & 0.170 & \cellcolor{yellow}\textbf{0.344} & 0.226 & 0.154 & 0.106 \\
 & \textsc{qwen-l} & 0.087 & \cellcolor{yellow}\textbf{0.345} & 0.252 & 0.182 & 0.134 \\
\midrule
\multirow{6}{*}{\textsc{clinician}} & \textsc{claude} & 0.050 & \cellcolor{yellow}\textbf{0.300} & 0.175 & \cellcolor{yellow}\textbf{0.300} & 0.175 \\
 & \textsc{gpt} & 0.128 & \cellcolor{yellow}\textbf{0.410} & 0.205 & 0.154 & 0.103 \\
 & \textsc{gemma-s} & 0.200 & \cellcolor{yellow}\textbf{0.425} & 0.175 & 0.100 & 0.100 \\
 & \textsc{gemma-l} & 0.081 & \cellcolor{yellow}\textbf{0.351} & 0.162 & 0.189 & 0.216 \\
 & \textsc{qwen-s} & 0.026 & \cellcolor{yellow}\textbf{0.487} & 0.077 & 0.179 & 0.231 \\
 & \textsc{qwen-l} & 0.025 & \cellcolor{yellow}\textbf{0.375} & 0.125 & 0.225 & 0.250 \\
\bottomrule
\end{tabular}
\end{adjustbox}
\caption{Distribution of the checkpoint at which each model first commits to a decision on \textsc{simulated} and \textsc{clinician}, under \textsc{sequential-free}. Each row is a fraction of committed conversations and sums to one (up to rounding). Most frequent commit checkpoint for a model is represented as \textbf{bold} and \colorbox{yellow}{highlighted}.}
\label{tab:app-ci-commit}
\end{table}

\subsection{Surprisal of the true label}
\label{app:ci-surp}

Table~\ref{tab:app-ci-surp} reports $S(y_{\text{true}}) = -\ln p(y_{\text{true}} \mid \text{context})$ for each open-weights model on all checkpoints, across both corpora. Values plotted in Figure~\ref{fig:rq4-surprisal} on \textsc{simulated} and \textsc{clinician}.

\begin{table}[t!]
\centering
\begin{adjustbox}{max width=\linewidth}
\begin{tabular}{llccccc}
\toprule
Corpus & Model & \textit{k}=4 & \textit{k}=+25\% & \textit{k}=+50\% & \textit{k}=+75\% & \textit{k}=+100\% \\
\midrule
\multirow{4}{*}{\textsc{simulated}} & \textsc{gemma-s} & \cellcolor{yellow}\textbf{6.60} & 7.29 & 7.42 & 7.68 & 7.75 \\
 & \textsc{gemma-l} & \cellcolor{yellow}\textbf{8.45} & 9.48 & 9.72 & 10.10 & 10.27 \\
 & \textsc{qwen-s} & \cellcolor{yellow}\textbf{2.22} & 2.59 & 2.74 & 2.84 & 2.96 \\
 & \textsc{qwen-l} & \cellcolor{yellow}\textbf{2.51} & 3.06 & 3.39 & 3.56 & 3.71 \\
\midrule
\multirow{4}{*}{\textsc{clinician}} & \textsc{gemma-s} & 4.64 & \cellcolor{yellow}\textbf{3.56} & 3.69 & 4.20 & 4.04 \\
 & \textsc{gemma-l} & \cellcolor{yellow}\textbf{4.52} & 6.09 & 6.64 & 6.78 & 6.75 \\
 & \textsc{qwen-s} & 1.49 & \cellcolor{yellow}\textbf{1.39} & 1.47 & 1.65 & 1.73 \\
 & \textsc{qwen-l} & \cellcolor{yellow}\textbf{1.76} & \cellcolor{yellow}\textbf{1.76} & 1.98 & 2.27 & 2.50 \\
\bottomrule
\end{tabular}
\end{adjustbox}
\caption{Mean surprisal of the true label, $S(y_{\text{true}})$ in nats, for the four open-weight models at each checkpoint on \textsc{simulated} and \textsc{clinician}, under \textsc{sequential-vanilla}. Checkpoint with the least surprisal for a model is represented as \textbf{bold} and \colorbox{yellow}{highlighted}.}
\label{tab:app-ci-surp}
\end{table}

\subsection{Chief-complaint recovery per model}
\label{app:cc-recovery}

Table~\ref{tab:app-cc-cos} reports the mean cosine ($\bar{\text{cos}}$) per model and ESI label, on \textsc{simulated} and \textsc{clinician}.

\begin{table}[t!]
\centering
\small
\setlength{\tabcolsep}{4pt}
\begin{adjustbox}{max width=\linewidth}
\begin{tabular}{llcccccc}
\toprule
Corpus & Model & ESI 1 & ESI 2 & ESI 3 & ESI 4 & ESI 5 & all \\
\midrule
\multirow{6}{*}{\textsc{simulated}} & \textsc{claude} & 0.667 & 0.746 & 0.694 & 0.720 & 0.585 & 0.684 \\
 & \textsc{gpt} & \cellcolor{yellow}\textbf{0.706} & \cellcolor{yellow}\textbf{0.756} & 0.706 & 0.729 & 0.576 & 0.697 \\
 & \textsc{gemma-s} & 0.644 & 0.675 & 0.681 & 0.709 & 0.593 & 0.662 \\
 & \textsc{gemma-l} & 0.696 & 0.753 & \cellcolor{yellow}\textbf{0.733} & \cellcolor{yellow}\textbf{0.772} & \cellcolor{yellow}\textbf{0.642} & \cellcolor{yellow}\textbf{0.720} \\
 & \textsc{qwen-s} & 0.660 & 0.692 & 0.696 & 0.695 & 0.565 & 0.663 \\
 & \textsc{qwen-l} & 0.673 & 0.720 & 0.696 & 0.718 & 0.581 & 0.679 \\
\midrule
\multirow{6}{*}{\textsc{clinician}} & \textsc{claude} & \cellcolor{yellow}\textbf{0.585} & 0.646 & 0.709 & 0.552 & 0.620 & 0.635 \\
 & \textsc{gpt} & 0.547 & 0.658 & 0.750 & \cellcolor{yellow}\textbf{0.648} & 0.656 & 0.672 \\
 & \textsc{gemma-s} & 0.476 & 0.566 & 0.678 & 0.498 & 0.528 & 0.569 \\
 & \textsc{gemma-l} & 0.466 & \cellcolor{yellow}\textbf{0.661} & \cellcolor{yellow}\textbf{0.804} & 0.613 & \cellcolor{yellow}\textbf{0.686} & \cellcolor{yellow}\textbf{0.680} \\
 & \textsc{qwen-s} & 0.562 & 0.551 & 0.720 & 0.516 & 0.500 & 0.586 \\
 & \textsc{qwen-l} & 0.546 & 0.588 & 0.750 & 0.600 & 0.582 & 0.633 \\
\bottomrule
\end{tabular}
\end{adjustbox}
\caption{Mean cosine similarity ($\bar{\text{cos}}$) between the model-drafted and gold chief complaint by model and ESI level on \textsc{simulated} and \textsc{clinician}. Best performing model for a corpus at an ESI level is represented as \textbf{bold} and \colorbox{yellow}{highlighted}.}
\label{tab:app-cc-cos}
\end{table}

%% file: appendix/E-clinician-corp.tex
\section{Additional Results on \textsc{clinician}}
\label{app:d-clin-corp}

\subsection{Reasoning-strategy trajectories on \textsc{clinician}}
\label{app:rq1-reason-clin}

\begin{figure*}[t!]
\centering
\includegraphics[width=\linewidth]{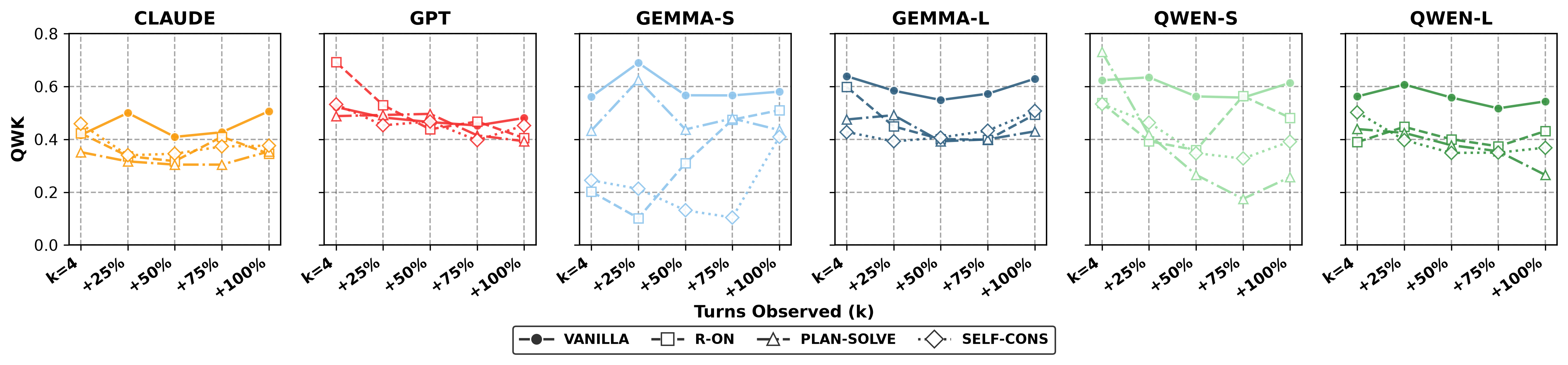}
\caption{Performance comparison of all models across the five sequential checkpoints on the \textsc{clinician} corpus under four prompting strategies. Companion of Figure~\ref{fig:rq1-reason}.}
\label{fig:app-rq1-reason-clin}
\end{figure*}

Figure~\ref{fig:app-rq1-reason-clin} reports the four decoding-time strategies (\textsc{vanilla}, \textsc{r-on}, \textsc{ps+}, \textsc{sc}) on the \textsc{clinician} corpus. Numerical values including 95\% CIs are in the \textsc{clinician} block of Table~\ref{tab:app-ci-reason}.

\subsection{Confusion matrices for the remaining five models}
\label{app:rq1-cm-all}

\begin{figure*}[t!]
\centering
\includegraphics[width=\linewidth]{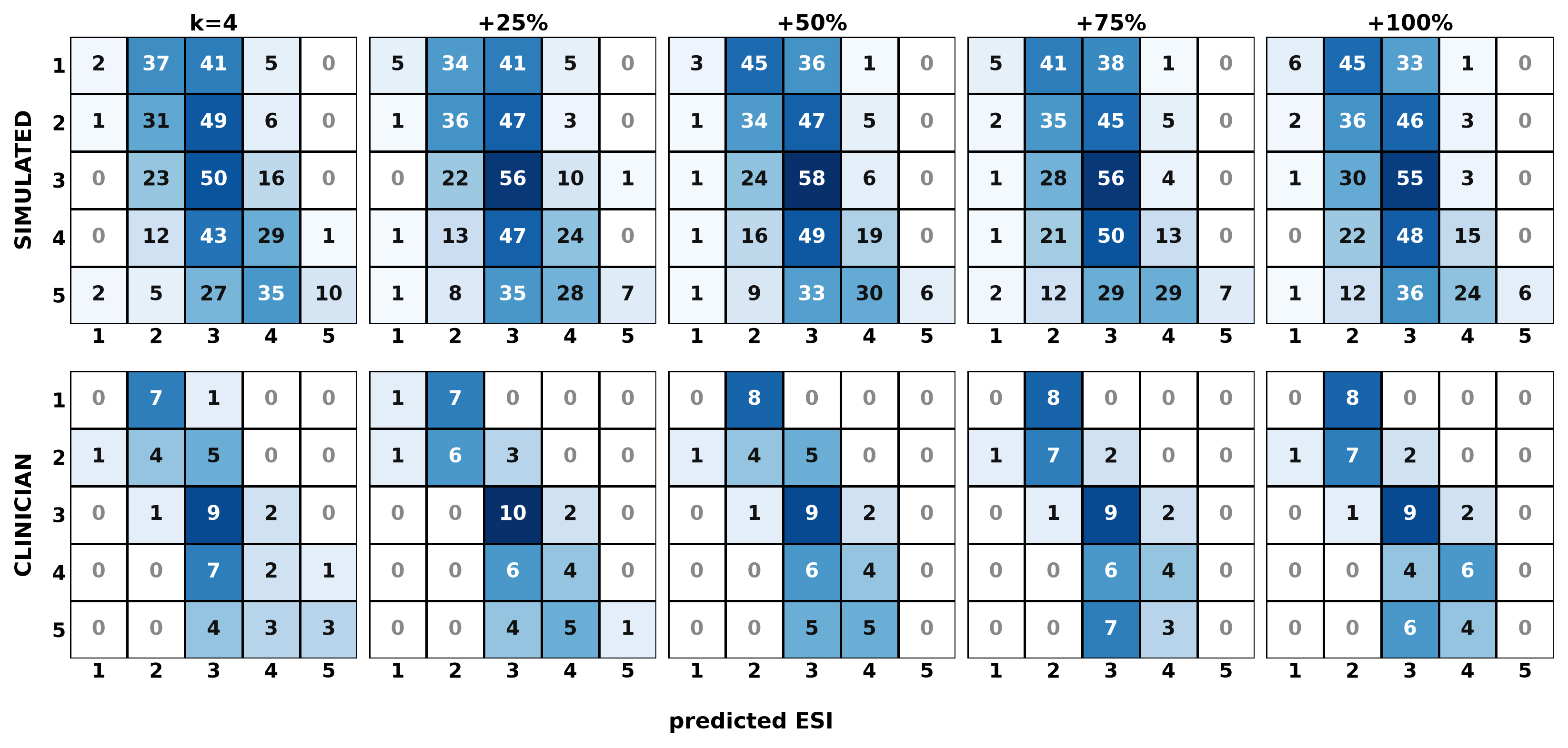}
\caption{\textsc{gemma-s} confusion matrices at each checkpoint on \textsc{simulated} (top row, n=425) and \textsc{clinician} (bottom row, n=40).}
\label{fig:app-cm-gemma-s}
\end{figure*}

\begin{figure*}[t!]
\centering
\includegraphics[width=\linewidth]{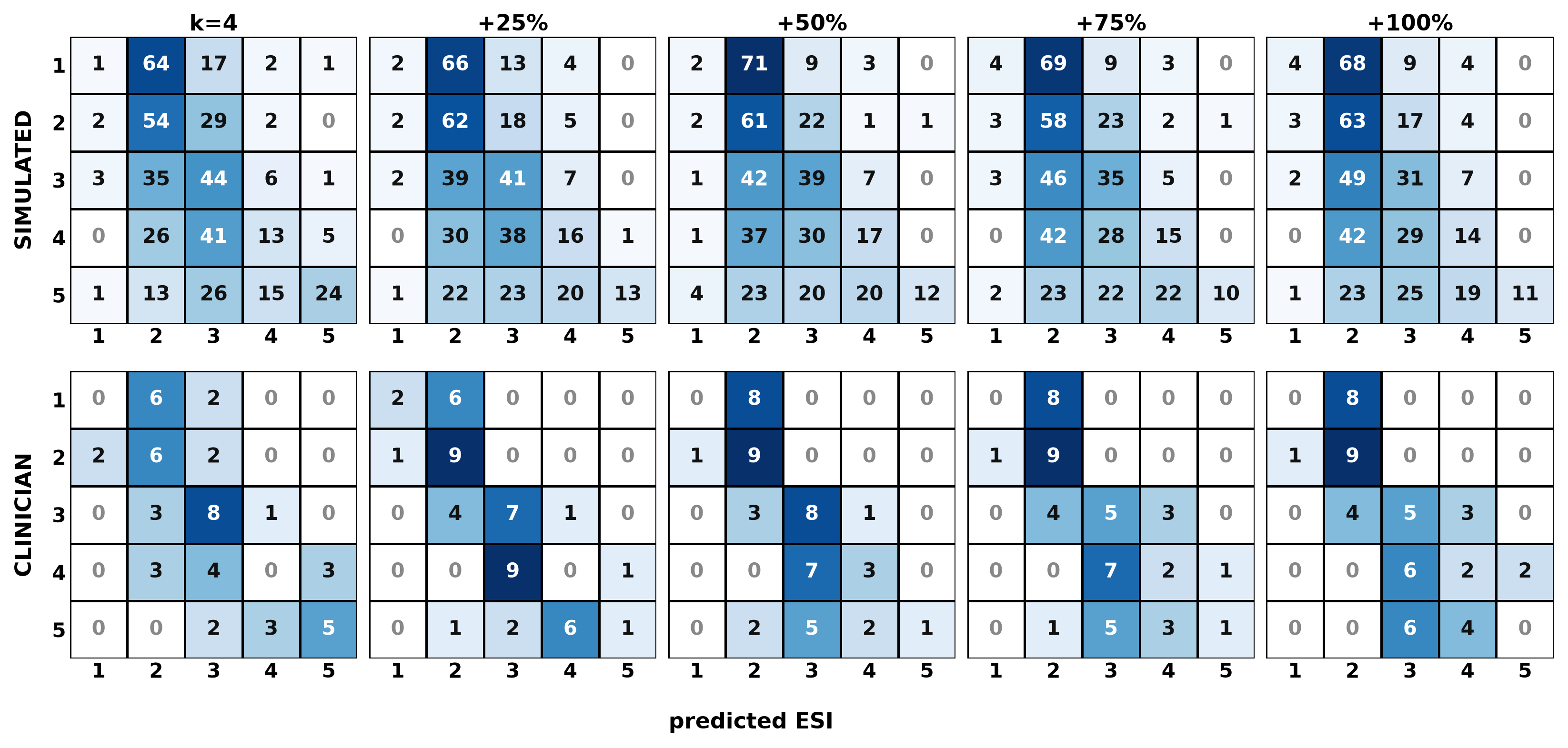}
\caption{\textsc{qwen-s} confusion matrices at each checkpoint on \textsc{simulated} (top row, n=425) and \textsc{clinician} (bottom row, n=40).}
\label{fig:app-cm-qwen-s}
\end{figure*}

\begin{figure*}[t!]
\centering
\includegraphics[width=\linewidth]{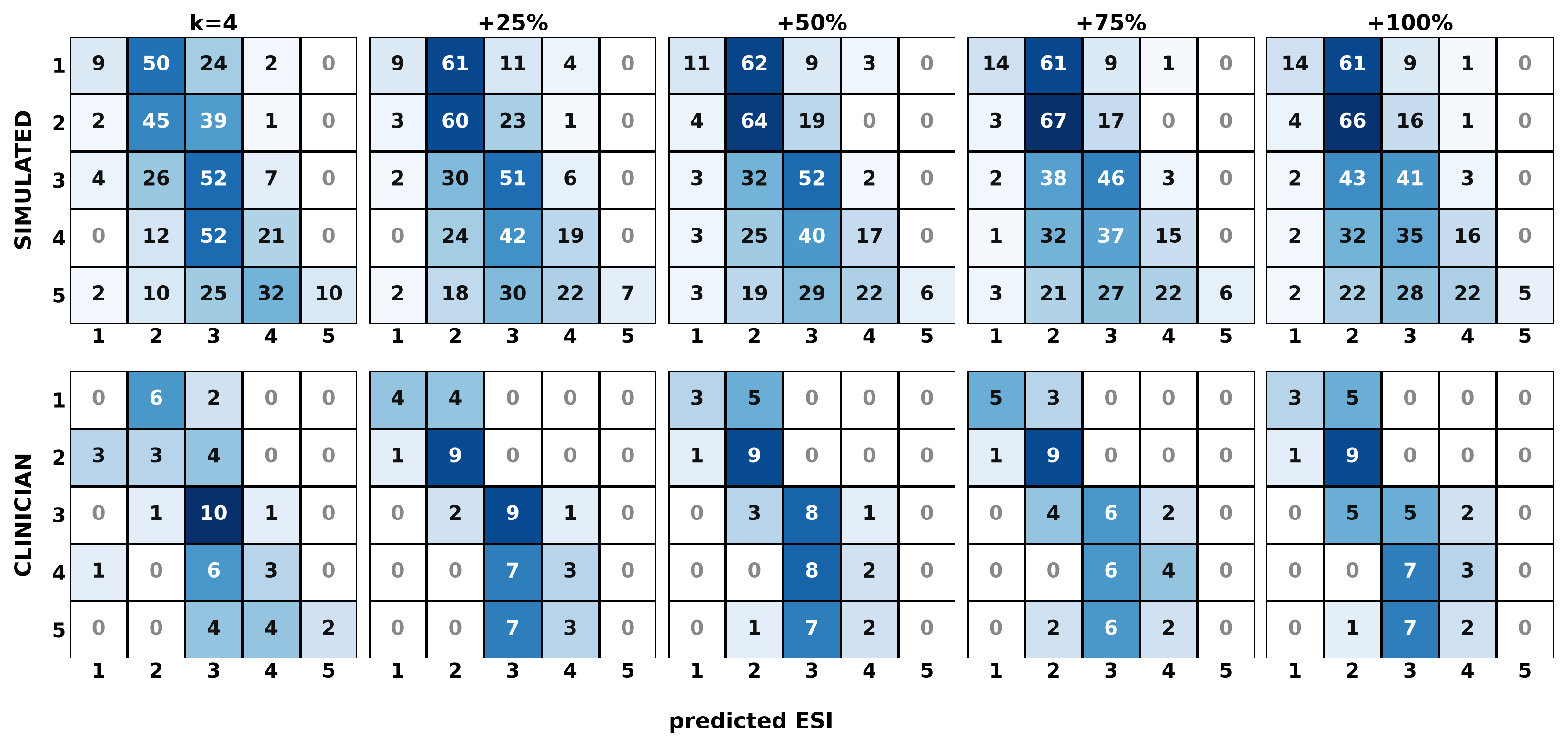}
\caption{\textsc{qwen-l} confusion matrices at each checkpoint on \textsc{simulated} (top row, n=425) and \textsc{clinician} (bottom row, n=40).}
\label{fig:app-cm-qwen-l}
\end{figure*}

\begin{figure*}[t!]
\centering
\includegraphics[width=\linewidth]{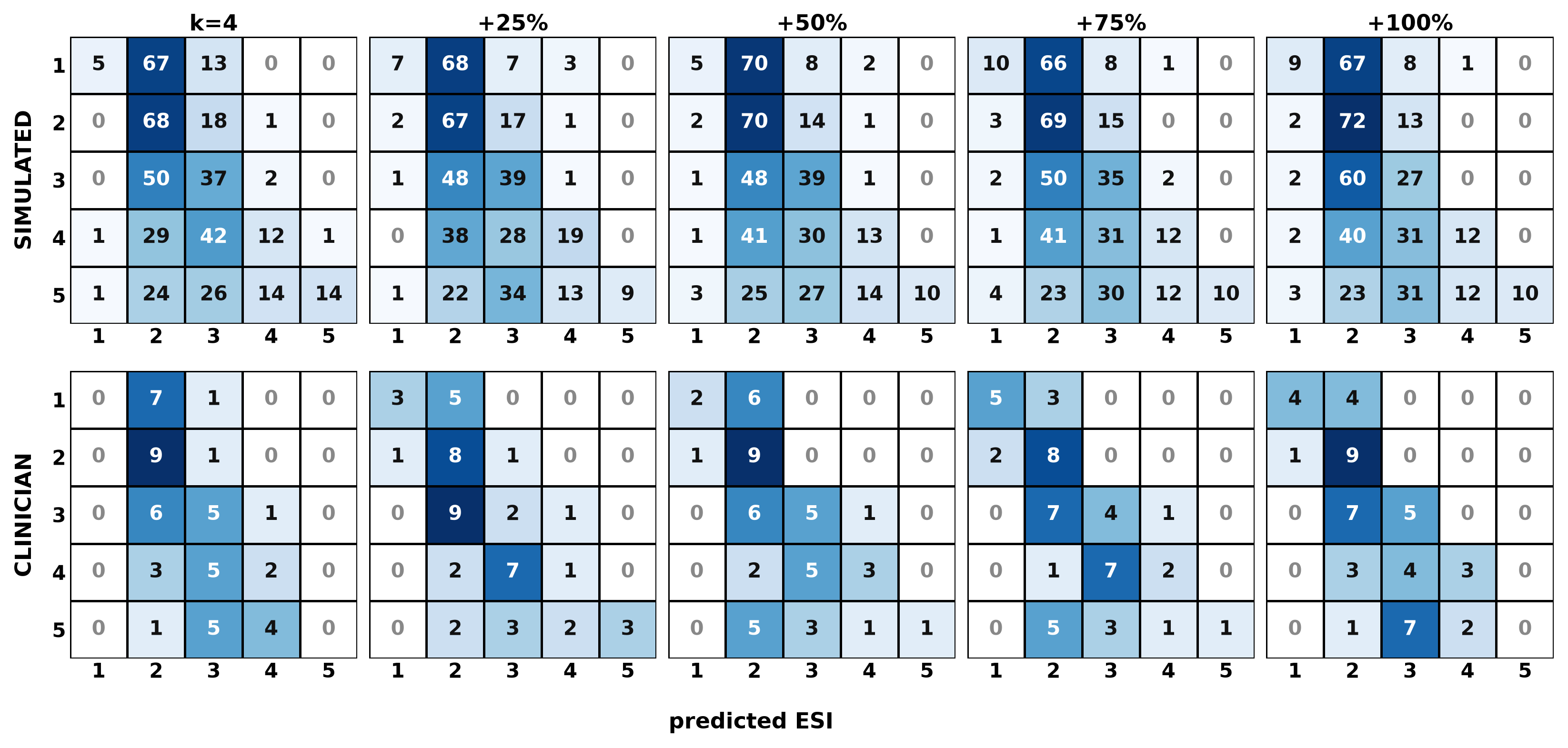}
\caption{\textsc{claude} confusion matrices at each checkpoint on \textsc{simulated} (top row, n=425) and \textsc{clinician} (bottom row, n=40).}
\label{fig:app-cm-claude}
\end{figure*}

\begin{figure*}[t!]
\centering
\includegraphics[width=\linewidth]{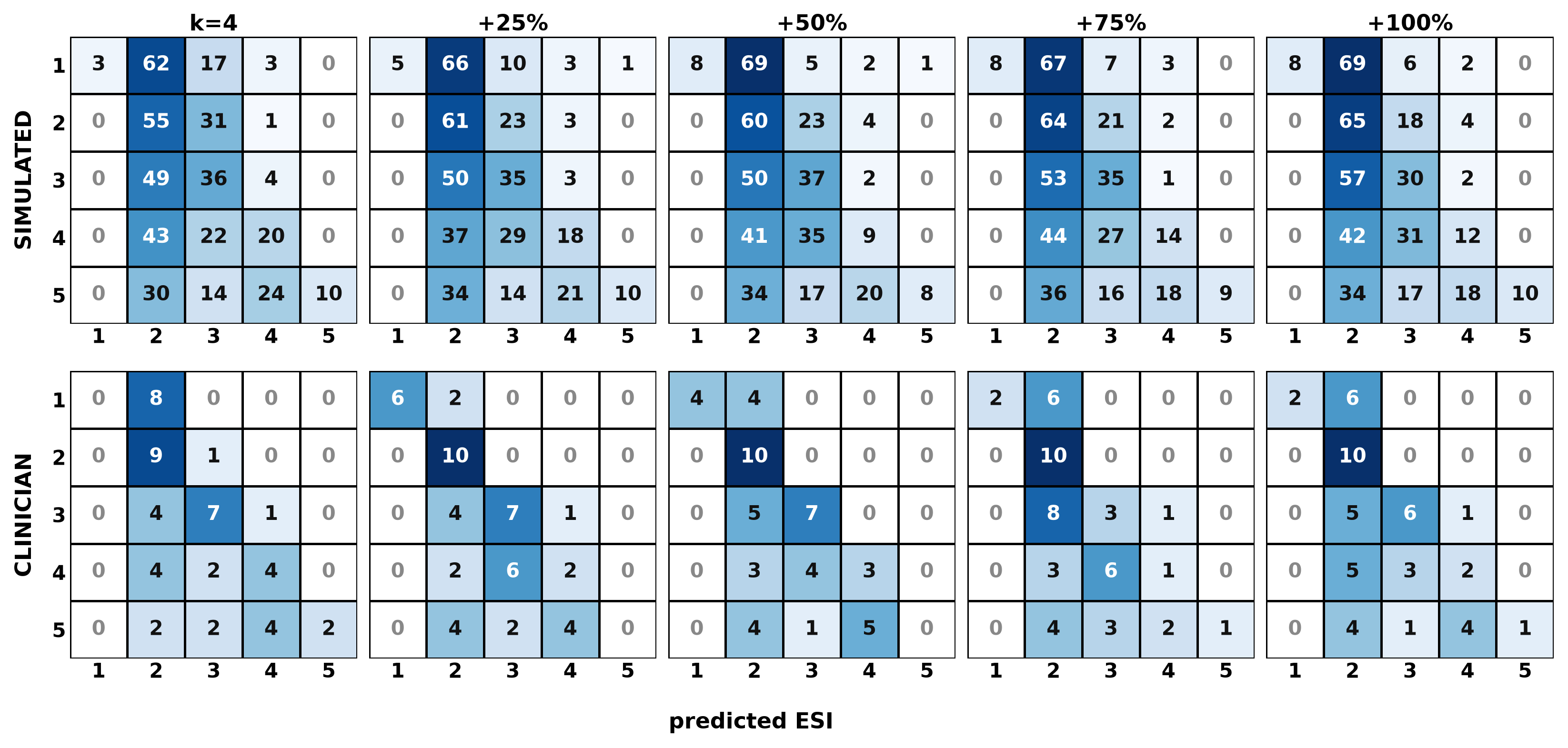}
\caption{\textsc{gpt} confusion matrices at each checkpoint on \textsc{simulated} (top row, n=425) and \textsc{clinician} (bottom row, n=40).}
\label{fig:app-cm-gpt}
\end{figure*}

Figures~\ref{fig:app-cm-gemma-s} to \ref{fig:app-cm-gpt} report the confusion matrices for the five non-headline models under the FORCED protocol on both corpora across the five checkpoints. Top row of each figure*: \textsc{simulated} (n=425); bottom row: \textsc{clinician}. Predictions concentrate at ESI-2 and ESI-3 on every model at every checkpoint. Row ESI-1 is under-triaged towards the centre on every model on both corpora.

\subsection{Perturbation heatmap on the \textsc{clinician} corpus}
\label{app:rq2-clin}

\begin{figure*}[t!]
\centering
\includegraphics[width=\linewidth]{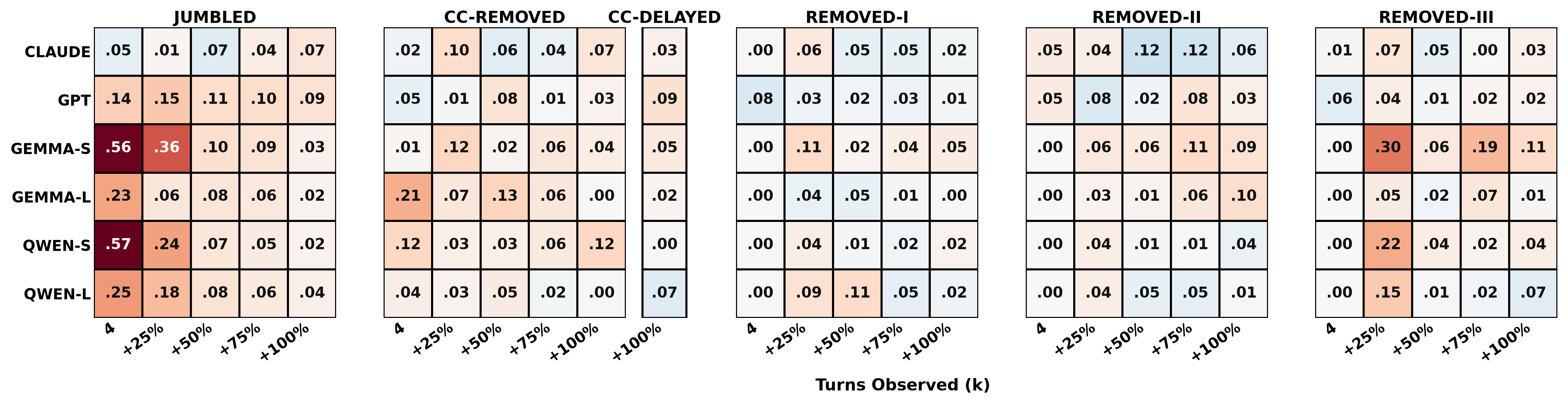}
\caption{Change in QWK per perturbation, model, and checkpoint on the \textsc{clinician} corpus. Blue = perturbation degrades QWK relative to VANILLA; red = perturbation improves QWK. Companion of Figure~\ref{fig:rq2-abl}.}
\label{fig:app-abl-clin}
\end{figure*}

Figure~\ref{fig:app-abl-clin} reports the $\Delta$ QWK grid on \textsc{clinician} under the same six perturbations as Figure~\ref{fig:rq2-abl}. The pattern from \textsc{simulated} holds. \textsc{jumbled} shifts $\Delta$ QWK on every model at every checkpoint. \textsc{cc-removed} shifts $\Delta$ QWK at \textit{k=4} and the shift persists through later checkpoints. \textsc{cc-delayed} tracks \textsc{cc-removed} through \textit{k=+75\%} and recovers to baseline at \textit{k=+100\%}. The three interval-removal perturbations (\textsc{removed-i}, \textsc{removed-ii}, \textsc{removed-iii}) produce $\Delta$ values near zero at every checkpoint on every model.

\subsection{Model commit distribution on the \textsc{clinician} corpus}
\label{app:rq3-clinician}

\begin{figure*}[t!]
\centering
\includegraphics[width=0.75\linewidth]{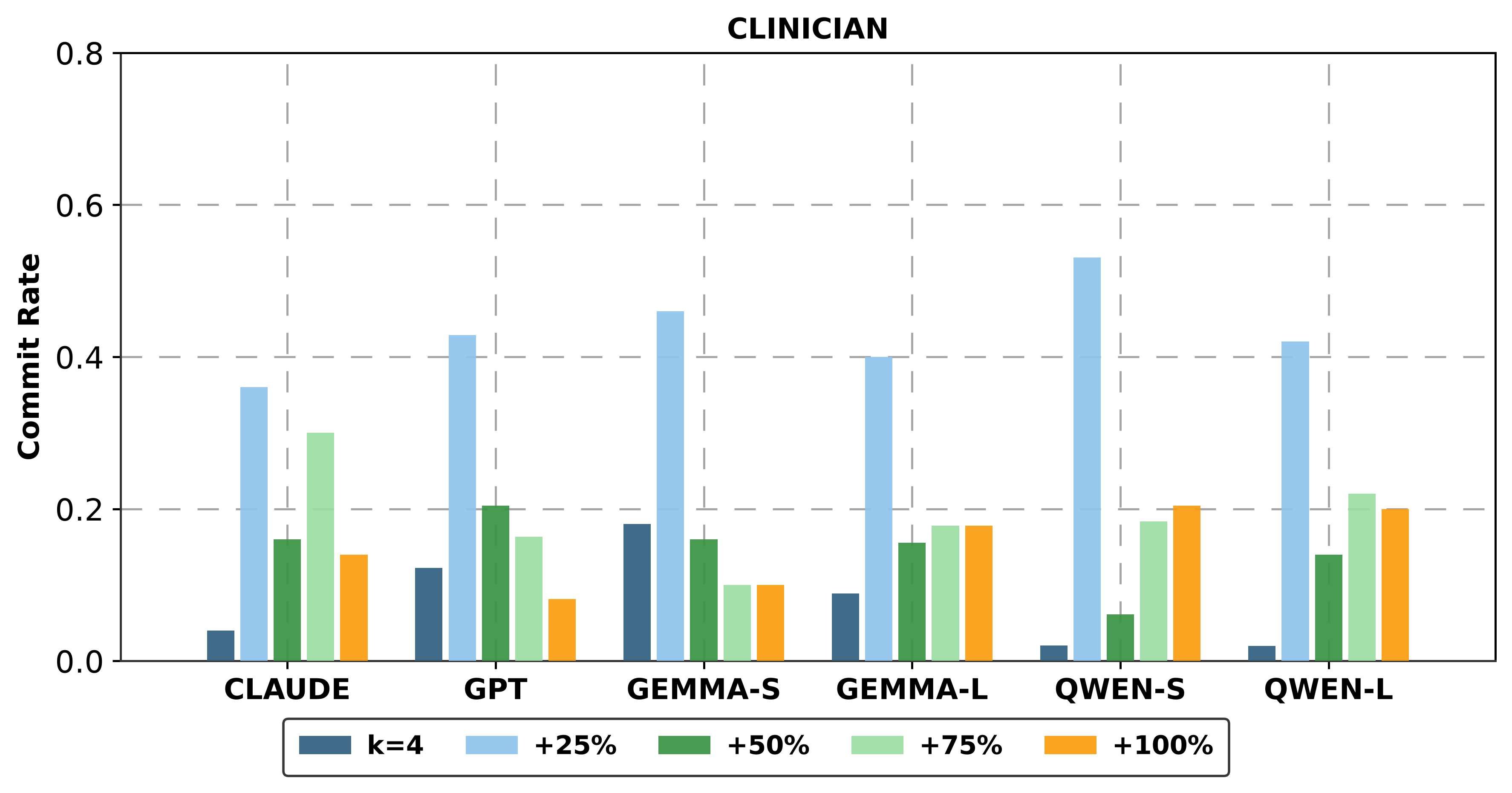}
\caption{Proportion of conversations committed at each checkpoint, per model on \textsc{clinician}.}
\label{fig:app-rq3-clinician}
\end{figure*}

The same central pattern from Section~\ref{sec:rq3} holds on the \textsc{clinician} corpus: most model commits land at \textit{k=+25\%} or \textit{k=+50\%}.

%% file: appendix/C-annot-exercise.tex
\section{Annotation Exercise}
\label{app:annot-exercise}


\paragraph{Annotators.}
The three annotators invited to participate in the exercise were made co-authors for their contribution to this paper. All annotators have superior proficiency in English, and their experience working in the ED is provided below: 
\begin{itemize}
    \item \textsc{annot \#1} has been practising medicine for 7 years with 2 years of work experience in the ED.
    \item \textsc{annot \#2} is a trained nurse with 1 year of work experience in the ED.
    \item \textsc{annot \#3} has been practising medicine for 2 years with 3 months of work experience in the ED.
\end{itemize}

The annotation exercise was approved by the internal ethics committee at the lead author's host organisation. All conversations are simulated, hence no patient data was shown to annotators.

\begin{figure*}[t!]
    \centering
    \includegraphics[width=1\linewidth]{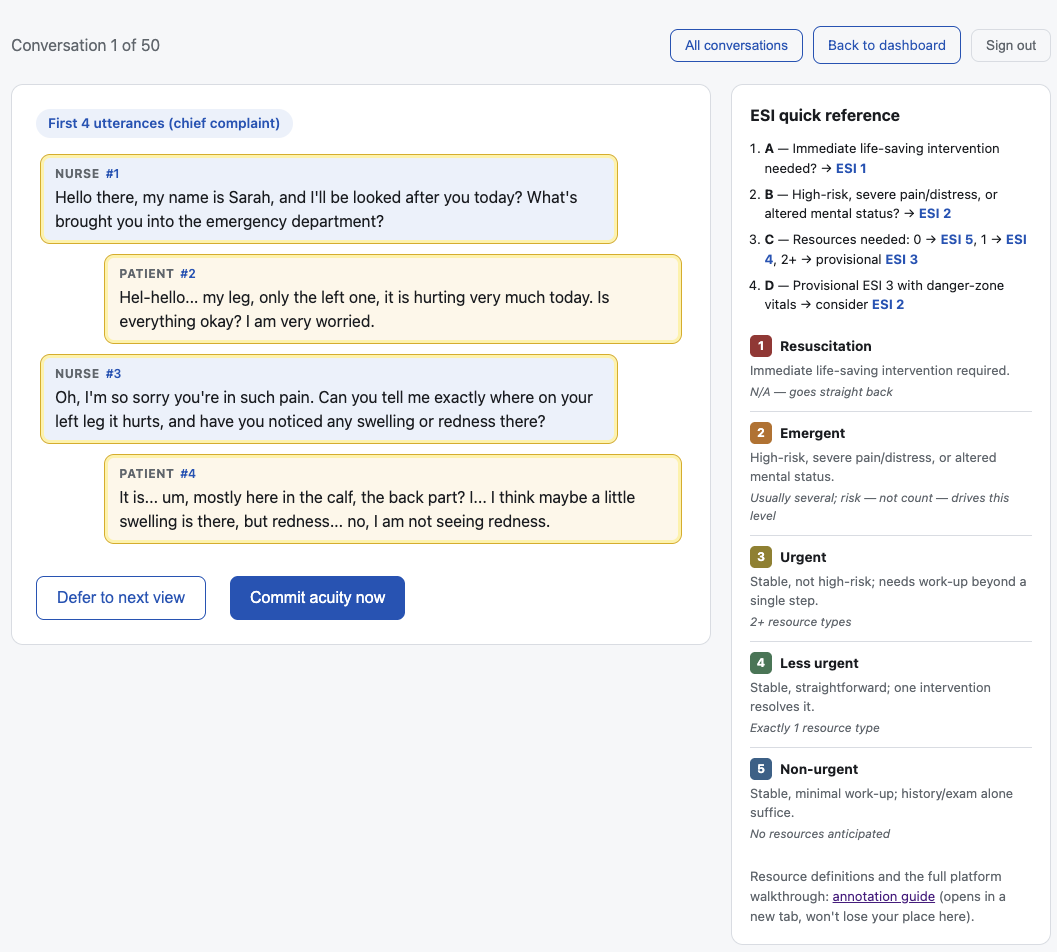}
    \caption{Screenshot of an example prefix provided to an annotator.}
    \label{fig:annot-portal}
\end{figure*}

\paragraph{Instrument.} Annotators participated in the exercise through a custom web-based interface, built with the help of a generative AI tool. Each conversation is revealed incrementally with the checkpoint \textit{k} $\in$ \{4, +25\%, +50\%, +75\%, +100\%\}. At every checkpoint the annotator is asked to perform either of the two actions: (a) commit to an ESI level, or (b) defer to see the next checkpoint. As shown in Figure~\ref{fig:annot-portal}, we also provide a quick reference guide of the ESI algorithm to the annotators during the exercise. The final checkpoint forces a commit for any conversation not yet resolved, matching the evaluation setup for the models in Section~\ref{sec:rq3}. The interface records, per conversation: the checkpoint at which the annotator committed, every checkpoint at which they deferred, and the acuity they committed. Annotators never see the ground-truth label, any model prediction, or any other annotator's responses.

\paragraph{Committing and deferring.}
Annotators are instructed to defer when they do not yet have enough information to assign acuity with confidence, and to commit as soon as they would in practice.

\paragraph{Workload.}
Every annotator sees the same conversations, and the annotators are asked not to discuss specific cases with each other while they perform the exercise. The guidelines advise that most conversations take two to four minutes and the exercise may be split across sessions. Both annotators, on average, spend 1.5 minutes on a conversation.

%% file: references.bib
@inproceedings{lu-etal-2024-triageagent,
    title = "{T}riage{A}gent: Towards Better Multi-Agents Collaborations for Large Language Model-Based Clinical Triage",
    author = "Lu, Meng  and
      Ho, Brandon  and
      Ren, Dennis  and
      Wang, Xuan",
    editor = "Al-Onaizan, Yaser  and
      Bansal, Mohit  and
      Chen, Yun-Nung",
    booktitle = "Findings of the Association for Computational Linguistics: EMNLP 2024",
    month = nov,
    year = "2024",
    address = "Miami, Florida, USA",
    publisher = "Association for Computational Linguistics",
    url = "https://aclanthology.org/2024.findings-emnlp.329/",
    doi = "10.18653/v1/2024.findings-emnlp.329",
    pages = "5747--5764"
}

@misc{Johnson2023,
  doi = {10.13026/5NTK-KM72},
  url = {https://physionet.org/content/mimic-iv-ed/2.2/},
  author = {Johnson,  Alistair and Bulgarelli,  Lucas and Pollard,  Tom and Celi,  Leo Anthony and Mark,  Roger and Horng,  Steven},
  title = {MIMIC-IV-ED},
  publisher = {PhysioNet},
  year = {2023}
}

@report{esi-handbook,
  author       = {Gilboy, Nicki and Tanabe, Paula and Travers, Debbie A. and Rosenau, Alexander M. and Eitel, David R.},
  title        = {Emergency Severity Index, Version 4: Implementation Handbook},
  institution  = {Agency for Healthcare Research and Quality},
  number       = {05-0046-2},
  year         = {2005},
  location     = {Rockville, MD},
  url          = {https://www.sgnor.ch/fileadmin/user_upload/Dokumente/Downloads/Esi_Handbook.pdf},
  urldate      = {2025-09-19}
}

@article{Liu2025,
  title = {A foundational triage system for improving accuracy in moderate acuity level emergency classifications},
  volume = {5},
  ISSN = {2730-664X},
  url = {http://dx.doi.org/10.1038/s43856-025-01052-w},
  DOI = {10.1038/s43856-025-01052-w},
  number = {1},
  journal = {Communications Medicine},
  publisher = {Springer Science and Business Media LLC},
  author = {Liu,  Tuo and Gu,  Yang and Chen,  Hongyi and Zhang,  Yan and Zheng,  Leqi and Huang,  Xuanqi and Xu,  Yanjun and Wen,  Cai and Chen,  Mansheng and Lin,  Jiaqi and Huang,  Dongguo and Chen,  Feixia and Zhong,  Yulan and Chen,  Hui and Guo,  Yanfeng and Lu,  Mei and Zhang,  Guangwei and Wu,  Hao and Wang,  Changdong and Xi,  Xiaotu and Li,  Li and Yu,  Tao},
  year = {2025},
  month = jul 
}

@article{Masanneck2024,
  title = {Triage Performance Across Large Language Models,  ChatGPT,  and Untrained Doctors in Emergency Medicine: Comparative Study},
  volume = {26},
  ISSN = {1438-8871},
  url = {http://dx.doi.org/10.2196/53297},
  DOI = {10.2196/53297},
  journal = {Journal of Medical Internet Research},
  publisher = {JMIR Publications Inc.},
  author = {Masanneck,  Lars and Schmidt,  Linea and Seifert,  Antonia and K\"{o}lsche,  Tristan and Huntemann,  Niklas and Jansen,  Robin and Mehsin,  Mohammed and Bernhard,  Michael and Meuth,  Sven G and B\"{o}hm,  Lennert and Pawlitzki,  Marc},
  year = {2024},
  month = jun,
  pages = {e53297}
}

@article{tversky1974judgment,
  title = {Judgment under Uncertainty: Heuristics and Biases: Biases in judgments reveal some heuristics of thinking under uncertainty.},
  volume = {185},
  ISSN = {1095-9203},
  url = {http://dx.doi.org/10.1126/science.185.4157.1124},
  DOI = {10.1126/science.185.4157.1124},
  number = {4157},
  journal = {Science},
  publisher = {American Association for the Advancement of Science (AAAS)},
  author = {Tversky,  Amos and Kahneman,  Daniel},
  year = {1974},
  month = sep,
  pages = {1124–1131}
}

@article{ly2023anchoring,
  title = {Evidence for Anchoring Bias During Physician Decision-Making},
  volume = {183},
  ISSN = {2168-6106},
  url = {http://dx.doi.org/10.1001/jamainternmed.2023.2366},
  DOI = {10.1001/jamainternmed.2023.2366},
  number = {8},
  journal = {JAMA Internal Medicine},
  publisher = {American Medical Association (AMA)},
  author = {Ly,  Dan P. and Shekelle,  Paul G. and Song,  Zirui},
  year = {2023},
  month = aug,
  pages = {818}
}

@article{croskerry2002achieving,
  title = {Achieving Quality in Clinical Decision Making: Cognitive Strategies and Detection of Bias},
  volume = {9},
  ISSN = {1553-2712},
  url = {http://dx.doi.org/10.1197/aemj.9.11.1184},
  DOI = {10.1197/aemj.9.11.1184},
  number = {11},
  journal = {Academic Emergency Medicine},
  publisher = {Wiley},
  author = {Croskerry,  Pat},
  year = {2002},
  month = nov,
  pages = {1184–1204}
}

@article{cohen1968weighted,
  title = {Weighted kappa: Nominal scale agreement provision for scaled disagreement or partial credit.},
  volume = {70},
  ISSN = {0033-2909},
  url = {http://dx.doi.org/10.1037/h0026256},
  DOI = {10.1037/h0026256},
  number = {4},
  journal = {Psychological Bulletin},
  publisher = {American Psychological Association (APA)},
  author = {Cohen,  Jacob},
  year = {1968},
  pages = {213–220}
}

@article{williams2024llm,
  title = {Use of a Large Language Model to Assess Clinical Acuity of Adults in the Emergency Department},
  volume = {7},
  ISSN = {2574-3805},
  url = {http://dx.doi.org/10.1001/jamanetworkopen.2024.8895},
  DOI = {10.1001/jamanetworkopen.2024.8895},
  number = {5},
  journal = {JAMA Network Open},
  publisher = {American Medical Association (AMA)},
  author = {Williams,  Christopher Y. K. and Zack,  Travis and Miao,  Brenda Y. and Sushil,  Madhumita and Wang,  Michelle and Kornblith,  Aaron E. and Butte,  Atul J.},
  year = {2024},
  month = may,
  pages = {e248895}
}

@inproceedings{wei2018task,
    title = "Task-oriented Dialogue System for Automatic Diagnosis",
    author = "Wei, Zhongyu  and
      Liu, Qianlong  and
      Peng, Baolin  and
      Tou, Huaixiao  and
      Chen, Ting  and
      Huang, Xuanjing  and
      Wong, Kam-fai  and
      Dai, Xiangying",
    editor = "Gurevych, Iryna  and
      Miyao, Yusuke",
    booktitle = "Proceedings of the 56th Annual Meeting of the Association for Computational Linguistics (Volume 2: Short Papers)",
    month = jul,
    year = "2018",
    address = "Melbourne, Australia",
    publisher = "Association for Computational Linguistics",
    url = "https://aclanthology.org/P18-2033/",
    doi = "10.18653/v1/P18-2033",
    pages = "201--207"
}

@article{chen2022diaformer,
  title = {Diaformer: Automatic Diagnosis via Symptoms Sequence Generation},
  volume = {36},
  ISSN = {2159-5399},
  url = {http://dx.doi.org/10.1609/aaai.v36i4.20365},
  DOI = {10.1609/aaai.v36i4.20365},
  number = {4},
  journal = {Proceedings of the AAAI Conference on Artificial Intelligence},
  publisher = {Association for the Advancement of Artificial Intelligence (AAAI)},
  author = {Chen,  Junying and Li,  Dongfang and Chen,  Qingcai and Zhou,  Wenxiu and Liu,  Xin},
  year = {2022},
  month = {june},
  pages = {4432–4440}
}

@article{hager2024evaluation,
  title = {Evaluation and mitigation of the limitations of large language models in clinical decision-making},
  volume = {30},
  ISSN = {1546-170X},
  url = {http://dx.doi.org/10.1038/s41591-024-03097-1},
  DOI = {10.1038/s41591-024-03097-1},
  number = {9},
  journal = {Nature Medicine},
  publisher = {Springer Science and Business Media LLC},
  author = {Hager,  Paul and Jungmann,  Friederike and Holland,  Robbie and Bhagat,  Kunal and Hubrecht,  Inga and Knauer,  Manuel and Vielhauer,  Jakob and Makowski,  Marcus and Braren,  Rickmer and Kaissis,  Georgios and Rueckert,  Daniel},
  year = {2024},
  month = {july},
  pages = {2613–2622}
}

@inproceedings{echterhoff2024cognitive,
    title = "Cognitive Bias in Decision-Making with {LLM}s",
    author = "Echterhoff, Jessica Maria  and
      Liu, Yao  and
      Alessa, Abeer  and
      McAuley, Julian  and
      He, Zexue",
    editor = "Al-Onaizan, Yaser  and
      Bansal, Mohit  and
      Chen, Yun-Nung",
    booktitle = "Findings of the Association for Computational Linguistics: EMNLP 2024",
    month = nov,
    year = "2024",
    address = "Miami, Florida, USA",
    publisher = "Association for Computational Linguistics",
    url = "https://aclanthology.org/2024.findings-emnlp.739/",
    doi = "10.18653/v1/2024.findings-emnlp.739",
    pages = "12640--12653"
}

@article{schmidgall2024cognitive,
  title = {Evaluation and mitigation of cognitive biases in medical language models},
  volume = {7},
  ISSN = {2398-6352},
  url = {http://dx.doi.org/10.1038/s41746-024-01283-6},
  DOI = {10.1038/s41746-024-01283-6},
  number = {1},
  journal = {npj Digital Medicine},
  publisher = {Springer Science and Business Media LLC},
  author = {Schmidgall,  Samuel and Harris,  Carl and Essien,  Ime and Olshvang,  Daniel and Rahman,  Tawsifur and Kim,  Ji Woong and Ziaei,  Rojin and Eshraghian,  Jason and Abadir,  Peter and Chellappa,  Rama},
  year = {2024},
  month = oct 
}

@misc{TriageSim,
      title={TriageSim: A Conversational Emergency Triage Simulation Framework from Structured Electronic Health Records}, 
      author={Dipankar Srirag and Quoc Dung Nguyen and Aditya Joshi and Padmanesan Narasimhan and Salil Kanhere},
      year={2026},
      eprint={2603.10035},
      archivePrefix={arXiv},
      primaryClass={cs.CL},
      url={https://arxiv.org/abs/2603.10035}, 
}

@inproceedings{jones2022capturing,
author = {Jones, Erik and Steinhardt, Jacob},
title = {Capturing failures of large language models via human cognitive biases},
year = {2022},
isbn = {9781713871088},
publisher = {Curran Associates Inc.},
address = {Red Hook, NY, USA},
booktitle = {Proceedings of the 36th International Conference on Neural Information Processing Systems},
articleno = {856},
numpages = {15},
location = {New Orleans, LA, USA},
series = {NIPS '22}
}

@inproceedings{wei2022chain,
author = {Wei, Jason and Wang, Xuezhi and Schuurmans, Dale and Bosma, Maarten and Ichter, Brian and Xia, Fei and Chi, Ed H. and Le, Quoc V. and Zhou, Denny},
title = {Chain-of-thought prompting elicits reasoning in large language models},
year = {2022},
isbn = {9781713871088},
publisher = {Curran Associates Inc.},
address = {Red Hook, NY, USA},
booktitle = {Proceedings of the 36th International Conference on Neural Information Processing Systems},
articleno = {1800},
numpages = {14},
location = {New Orleans, LA, USA},
series = {NIPS '22}
}

@article{Johri2025-craft-md,
  title = {An evaluation framework for clinical use of large language models in patient interaction tasks},
  volume = {31},
  ISSN = {1546-170X},
  url = {http://dx.doi.org/10.1038/s41591-024-03328-5},
  DOI = {10.1038/s41591-024-03328-5},
  number = {1},
  journal = {Nature Medicine},
  publisher = {Springer Science and Business Media LLC},
  author = {Johri,  Shreya and Jeong,  Jaehwan and Tran,  Benjamin A. and Schlessinger,  Daniel I. and Wongvibulsin,  Shannon and Barnes,  Leandra A. and Zhou,  Hong-Yu and Cai,  Zhuo Ran and Van Allen,  Eliezer M. and Kim,  David and Daneshjou,  Roxana and Rajpurkar,  Pranav},
  year = {2025},
  month = jan,
  pages = {77–86}
}

@inproceedings{chiu2025vivabench,
    title={Simulating Viva Voce Examinations to Evaluate Clinical Reasoning in Large Language Models},
    author={Christopher Chiu and Silviu Pitis and Mihaela van der Schaar},
    booktitle={The Thirty-ninth Annual Conference on Neural Information Processing Systems Datasets and Benchmarks Track},
    year={2025},
    url={https://openreview.net/forum?id=FEVfIPMy5b}
}

@article{maiwert2026q4dx,
  title = {A benchmark for evaluating diagnostic questioning efficiency of LLMs in patient conversations},
  volume = {16},
  ISSN = {2045-2322},
  url = {http://dx.doi.org/10.1038/s41598-026-37022-y},
  DOI = {10.1038/s41598-026-37022-y},
  number = {1},
  journal = {Scientific Reports},
  publisher = {Springer Science and Business Media LLC},
  author = {Werthaim,  Mai and Kimhi,  Maya and Apartsin,  Alexander and Aperstein,  Yehudit},
  year = {2026},
  month = jan 
}

@inproceedings{laban2025lost,
title={{LLM}s Get Lost In Multi-Turn Conversation},
author={Philippe Laban and Hiroaki Hayashi and Yingbo Zhou and Jennifer Neville},
booktitle={The Fourteenth International Conference on Learning Representations},
year={2026},
url={https://openreview.net/forum?id=VKGTGGcwl6}
}

@article{sax2023mistriage,
  title = {Evaluation of Version 4 of the Emergency Severity Index in US Emergency Departments for the Rate of Mistriage},
  volume = {6},
  ISSN = {2574-3805},
  url = {http://dx.doi.org/10.1001/jamanetworkopen.2023.3404},
  DOI = {10.1001/jamanetworkopen.2023.3404},
  number = {3},
  journal = {JAMA Network Open},
  publisher = {American Medical Association (AMA)},
  author = {Sax,  Dana R. and Warton,  E. Margaret and Mark,  Dustin G. and Vinson,  David R. and Kene,  Mamata V. and Ballard,  Dustin W. and Vitale,  Tina J. and McGaughey,  Katherine R. and Beardsley,  Aaron and Pines,  Jesse M. and Reed,  Mary E. and Rauchwerger,  Adina S and Zhang,  Jennifer Y},
  year = {2023},
  month = mar,
  pages = {e233404}
}

@article{Sax2025,
  title = {Emergency Department Triage Accuracy and Delays in Care for High-Risk Conditions},
  volume = {8},
  ISSN = {2574-3805},
  url = {http://dx.doi.org/10.1001/jamanetworkopen.2025.8498},
  DOI = {10.1001/jamanetworkopen.2025.8498},
  number = {5},
  journal = {JAMA Network Open},
  publisher = {American Medical Association (AMA)},
  author = {Sax,  Dana R. and Warton,  E. Margaret and Mark,  Dustin G. and Reed,  Mary E.},
  year = {2025},
  month = May,
  pages = {e258498}
}

@inproceedings{wang2023selfconsistency,
  title     = {Self-Consistency Improves Chain of Thought Reasoning in Language Models},
  author    = {Wang, Xuezhi and Wei, Jason and Schuurmans, Dale and Le, Quoc V. and 
               Chi, Ed H. and Narang, Sharan and Chowdhery, Aakanksha and Zhou, Denny},
  booktitle = {The Eleventh International Conference on Learning Representations},
  year      = {2023},
  url       = {https://openreview.net/forum?id=1PL1NIMMrw}
}

@inproceedings{wang2023plansolve,
    title = "Plan-and-Solve Prompting: Improving Zero-Shot Chain-of-Thought Reasoning by Large Language Models",
    author = "Wang, Lei  and
      Xu, Wanyu  and
      Lan, Yihuai  and
      Hu, Zhiqiang  and
      Lan, Yunshi  and
      Lee, Roy Ka-Wei  and
      Lim, Ee-Peng",
    editor = "Rogers, Anna  and
      Boyd-Graber, Jordan  and
      Okazaki, Naoaki",
    booktitle = "Proceedings of the 61st Annual Meeting of the Association for Computational Linguistics (Volume 1: Long Papers)",
    month = jul,
    year = "2023",
    address = "Toronto, Canada",
    publisher = "Association for Computational Linguistics",
    url = "https://aclanthology.org/2023.acl-long.147/",
    doi = "10.18653/v1/2023.acl-long.147",
    pages = "2609--2634"
}

@misc{nori2025sequentialdiagnosislanguagemodels,
      title={Sequential Diagnosis with Language Models}, 
      author={Harsha Nori and Mayank Daswani and Christopher Kelly and Scott Lundberg and Marco Tulio Ribeiro and Marc Wilson and Xiaoxuan Liu and Viknesh Sounderajah and Jonathan Carlson and Matthew P Lungren and Bay Gross and Peter Hames and Mustafa Suleyman and Dominic King and Eric Horvitz},
      year={2025},
      eprint={2506.22405},
      archivePrefix={arXiv},
      primaryClass={cs.CL},
      url={https://arxiv.org/abs/2506.22405}, 
}

@article{Joseph2023,
  title = {Race and Ethnicity and Primary Language in Emergency Department Triage},
  volume = {6},
  ISSN = {2574-3805},
  url = {http://dx.doi.org/10.1001/jamanetworkopen.2023.37557},
  DOI = {10.1001/jamanetworkopen.2023.37557},
  number = {10},
  journal = {JAMA Network Open},
  publisher = {American Medical Association (AMA)},
  author = {Joseph,  Joshua W. and Kennedy,  Maura and Landry,  Alden M. and Marsh,  Regan H. and Baymon,  Da’Marcus E. and Im,  Dana E. and Chen,  Paul C. and Samuels-Kalow,  Margaret E. and Nentwich,  Lauren M. and Elhadad,  Noémie and Sánchez,  León D.},
  year = {2023},
  month = Oct,
  pages = {e2337557}
}

@article{omar2025sociodemographic,
  title = {Sociodemographic biases in medical decision making by large language models},
  volume = {31},
  ISSN = {1546-170X},
  url = {http://dx.doi.org/10.1038/s41591-025-03626-6},
  DOI = {10.1038/s41591-025-03626-6},
  number = {6},
  journal = {Nature Medicine},
  publisher = {Springer Science and Business Media LLC},
  author = {Omar,  Mahmud and Soffer,  Shelly and Agbareia,  Reem and Bragazzi,  Nicola Luigi and Apakama,  Donald U. and Horowitz,  Carol R. and Charney,  Alexander W. and Freeman,  Robert and Kummer,  Benjamin and Glicksberg,  Benjamin S. and Nadkarni,  Girish N. and Klang,  Eyal},
  year = {2025},
  month = Apr,
  pages = {1873–1881}
}

@inproceedings{li2024mediq,
 author = {Li, Shuyue Stella and Balachandran, Vidhisha and Feng, Shangbin and Ilgen, Jonathan S. and Pierson, Emma and Koh, Pang Wei and Tsvetkov, Yulia},
 booktitle = {Advances in Neural Information Processing Systems},
 doi = {10.52202/079017-0908},
 editor = {A. Globerson and L. Mackey and D. Belgrave and A. Fan and U. Paquet and J. Tomczak and C. Zhang},
 pages = {28858--28888},
 publisher = {Curran Associates, Inc.},
 title = {MediQ: Question-Asking LLMs and a Benchmark for Reliable Interactive Clinical Reasoning},
 url = {https://proceedings.neurips.cc/paper_files/paper/2024/file/32b80425554e081204e5988ab1c97e9a-Paper-Conference.pdf},
 volume = {37},
 year = {2024}
}

@article{Brodeur2026,
  title = {Performance of a large language model on the reasoning tasks of a physician},
  volume = {392},
  ISSN = {1095-9203},
  url = {http://dx.doi.org/10.1126/science.adz4433},
  DOI = {10.1126/science.adz4433},
  number = {6797},
  journal = {Science},
  publisher = {American Association for the Advancement of Science (AAAS)},
  author = {Brodeur,  Peter G. and Buckley,  Thomas A. and Kanjee,  Zahir and Goh,  Ethan and Ling,  Evelyn Bin and Jain,  Priyank and Cabral,  Stephanie and Abdulnour,  Raja-Elie and Haimovich,  Adrian D. and Freed,  Jason A. and Olson,  Andrew and Morgan,  Daniel J. and Hom,  Jason and Gallo,  Robert and McCoy,  Liam G. and Mombini,  Haadi and Lucas,  Christopher and Fotoohi,  Misha and Gwiazdon,  Matthew and Restifo,  Daniele and Restrepo,  Daniel and Horvitz,  Eric and Chen,  Jonathan and Manrai,  Arjun K. and Rodman,  Adam},
  year = {2026},
  month = Apr,
  pages = {524–527}
}

@misc{zhu2026elicitedehrgroundedlongitudinalinteractive,
      title={ELICITED: EHR-grounded Longitudinal Interactive Conversations for Information-seeking Triage Evaluation and Decision-making}, 
      author={Haohao Zhu and Xiaolin Shi and Jiayu Zhou},
      year={2026},
      eprint={2608.09024},
      archivePrefix={arXiv},
      primaryClass={cs.CL},
      url={https://arxiv.org/abs/2608.09024}, 
}

@misc{srirag2026triagedischargesurveynlp,
      title={From Triage to Discharge: A Survey of NLP Tasks, Methods, and Open Challenges in the Emergency Department}, 
      author={Dipankar Srirag and Aditya Joshi and Salil Kanhere and Padmanesan Narasimhan},
      year={2026},
      eprint={2608.23627},
      archivePrefix={arXiv},
      primaryClass={cs.CL},
      url={https://arxiv.org/abs/2608.23627}, 
}

@misc{gemmateam2026gemma4technicalreport,
      title={Gemma 4 Technical Report}, 
      author={{Gemma Team} and Sherif El Abd and Vaibhav Aggarwal and Robin Algayres and Alek Andreev and Olivier Bachem and Ian Ballantyne and Cormac Brick and Victor Cărbune and Michelle Casbon and Mayank Chaturvedi and Aditya Chawla and Victor Cotruta and Alice Coucke and Phil Culliton and Robert Dadashi and Lucas Dixon and Mohamed Elhawaty and Utku Evci and Clément Farabet and Johan Ferret and Filippo Galgani and Sertan Girgin and Jean-Bastien Grill and Maarten Grootendorst and Jiaxian Guo and Cassidy Hardin and Yanzhang He and Steven M. Hernandez and Omri Homburger and Léonard Hussenot and Juyeong Ji and Armand Joulin and Aishwarya Kamath and Parnian Kassraie and Olivier Lacombe and Preethi Lahoti and Gaël Liu and Gus Martins and Luciano Martins and Tatiana Matejovicova and Ramona Merhej and Nikola Momchev and Sneha Mondal and Ryan Mullins and Sindhu Raghuram Panyam and Shreya Pathak and Sarah Perrin and André Susano Pinto and Etienne Pot and Angéline Pouget and Alexandre Ramé and Sabela Ramos and Douglas Reid and David Rim and Morgane Rivière and Karsten Roth and Louis Rouillard and Omar Sanseviero and Pier Giuseppe Sessa and Shane Settle and Danila Sinopalnikov and Sara Smoot and Piotr Stanczyk and Andreas Steiner and Lawrence Stewart and Ilya Tolstikhin and Michael Tschannen and Anton Tsitsulin and Nino Vieillard and Renjie Wu and Pingmei Xu and Haichuan Yang and Edouard Yvinec and Biao Zhang and Li Zhang and Joe Zou and Nicolas Aagnes and Abdelrahman Abdelhamed and Jakub Adamek and Shivani Agrawal and Shubham Agrawal and Ibrahim Alabdulmohsin and Jean Baptiste Alayrac and Uri Alon and Chandramouli Amarnath and Ankesh Anand and Chrysovalantis Anastasiou and Setareh Ariafar and François-Xavier Aubet and Kyriakos Axiotis and Federico Barbero and Joelle Barral and Alexei Bendebury and Urs Bergmann and Stanley Bileschi and Kat Black and Mathieu Blondel and Sebastian Borgeaud and Arthur Bražinskas and Ryan Burnell and Robert Busa-Fekete and Mu Cai and Daniele Calandriello and Glenn Cameron and Charlotte Caucheteux and Rahma Chaabouni and Garima Chadha and Jetha Chan and Blake Jianhang Chen and Jesse Chen and Lin Chen and Xu Chen and Derek Cheng and Tzu-hsiang Chien and Nikolai Chinaev and Yi Chou and Zhaohui Chu and Benjamin Coleman and Pooja Consul and Sam Conway-Rahman and Scott Crowell and Dylan Cutler and Vivek Dani and Samira Daruki and Anil Das and Daniel Deutsch and Nishanth Dikkala and Li Ding and Qiuhan Ding and Shenil Dodhia and Konstantin Donhauser and Tulsee Doshi and Anca Dragan and Alex Druinsky and Sahil Dua and Zoltan Egyed and Danielle Eisenbud and Daniel Eppens and Cindy Fan and Bahare Fatemi and Yassir Fathullah and Vlad Feinberg and Milen Ferev and Sebastian Flennerhag and Takumi Fujimoto and João Gabriel Oliveira and Isaac Galatzer-Levy and João Gante and Simon Geisler and Soham Ghosal and Antonious M. Girgis and Tamara von Glehn and Alec Go and Alhaad Gokhale and Alex Grills and Yiming Gu and Mayank Gupta and Pramod Gupta and Guru Guruganesh and Raia Hadsell and Hamza Harkous and Jitendra Harlalka and Demis Hassabis and Anja Hauth and Joe Heyward and Arian Hosseini and Chih-Yang Hsia and I-Hung Hsu and Xiaopeng Huang and Yangsibo Huang and Kevin Hui and Adrian Hutter and Te I and Fotis Iliopoulos and Advait Jain and Ganesh Jawahar and Ziwei Ji and Qilin Jin and Melvin Johnson and Kandarp Joshi and Arun Kandoor and Wang-Cheng Kang and Koray Kavukcuoglu and Mehran Kazemi and Kathleen Kenealy and Amr Khalifa and Phoebe Kirk and Ivan Korotkov and Suraj Kothawade and Vitaly Kovalev and Neel Kovelamudi and Adam Kraft and Ravin Kumar and Vivek Kumar and Harish Kuppam and Justin Lannin and Chen-Yu Lee and Seungji Lee and Dmitry Lepikhin and Alon Levkovitch and Dongdong Li and Qiujia Li and Valentin Liévin and Ethan Lin and Ziqian Lin and Casper Liu and Tianlin Liu and Tianqi Liu and Xin Liu and Ivan Lobov and Mayank Lunayach and Min Ma and Gagan Madan and Andrii Maksai and Eric Malmi and Michal Matuszak and Daniel McDuff and Gaurav Menghani and Maciej Mikuła and Daniil Mirylenka and Karolis Misiunas and Vedant Misra and Andreea Mitran and Kareem Mohamed and Maksim Mukha and Eric Noland and James O'Donnell and Brendan O'Donoghue and Kate Olszewska and Bernett Orlando and Wanqiong Pan and Rina Panigrahy and Unnati Parekh and Nicolas Perez-Nieves and Chunjong Park and Eric Paskie and Liqian Peng and Bryce Petrini and Slav Petrov and Jonas Pfeiffer and Bilal Piot and Martyna Plomecka and Siim Poder and Octavio Ponce and Arijit Pramanik and David Racz and Anish Rajan and Michelle Ramanovich and Anand Rao and Marvin Ritter and Vitor Rodrigues and Evan Rosen and Mikołaj Rybiński and Noveen Sachdeva and Michaël E. Sander and Rohit Sathyanarayana and Sagar Savla and Samuel Schmidgall and Tal Schuster and George Scrivener and Benoit Seguin and Andrew Sellergren and Aliaksei Severyn and Izhak Shafran and Dhruv Shah and Bobak Shahriari and Yuan Shangguan and Ashish Shenoy and Pradeep Shenoy and Rakesh Shivanna and Pauline Sho and Lucas Spangher and Wojciech Stokowiec and Tim Strother and Yao Su and Yinghao Sun and Mukund Sundararajan and Andrea Tacchetti and Mor Hazan Taege and Pouya Tafti and Jean Tarbouriech and Chetan Tekur and Shantanu Thakoor and Rahul Thapa and Madeleine Traverse and Lenart Treven and Tao Tu and Chien Te Tung and Çağlar Ünlü and Petar Veličković and Malini Pooni Venkat and Sagar Gubbi Venkatesh and Vidya Venkiteswaran and Francesco Visin and Alex Vitvitskyi and Kiran Vodrahalli and Weiyi Wang and Xin Wang and Tris Warkentin and Jan Wassenberg and John Wieting and Cindy Wu and Lechao Xiao and Hao Xu and Yuhui Xu and Fuzhao Xue and Arun Yadav and Jun Yan and Antoine Yang and Lin Yang and Ming-Hsuan Yang and Ziyu Ying and Jae Hyeon Yoo and Morteza Zadimoghaddam and Sajjad Zafar and Fred Zhang and Jiageng Zhang and Jianyi Zhang and Xiaofan Zhang and Chao Zhao and David Zhou and Chen Zou},
      year={2026},
      eprint={2607.02770},
      archivePrefix={arXiv},
      primaryClass={cs.CL},
      url={https://arxiv.org/abs/2607.02770}, 
}

@misc{qwenteam2026qwen35omnitechnicalreport,
      title={Qwen3.5-Omni Technical Report}, 
      author={{Qwen Team}},
      year={2026},
      eprint={2604.15804},
      archivePrefix={arXiv},
      primaryClass={cs.CL},
      url={https://arxiv.org/abs/2604.15804}, 
}

@inproceedings{reimers-gurevych-2019-sentence,
    title = "Sentence-{BERT}: Sentence Embeddings using {S}iamese {BERT}-Networks",
    author = "Reimers, Nils  and
      Gurevych, Iryna",
    editor = "Inui, Kentaro  and
      Jiang, Jing  and
      Ng, Vincent  and
      Wan, Xiaojun",
    booktitle = "Proceedings of the 2019 Conference on Empirical Methods in Natural Language Processing and the 9th International Joint Conference on Natural Language Processing (EMNLP-IJCNLP)",
    month = nov,
    year = "2019",
    address = "Hong Kong, China",
    publisher = "Association for Computational Linguistics",
    url = "https://aclanthology.org/D19-1410/",
    doi = "10.18653/v1/D19-1410",
    pages = "3982--3992"
}

@article{ESPEJO202557,
title = {The Emergency Severity Index (ESI) Version 5: Simulation of Predictive Validity and Triage Level Distribution},
journal = {The Journal of Emergency Medicine},
volume = {78},
pages = {57-70},
year = {2025},
issn = {0736-4679},
doi = {https://doi.org/10.1016/j.jemermed.2025.07.035},
url = {https://www.sciencedirect.com/science/article/pii/S0736467925002884},
author = {Tanguy Espejo and Florian F. Grossmann and Henk B. Riedel and Roland Bingisser and Christian H. Nickel}
}

@inproceedings{yao-yu-2026-llm,
    title = "{LLM}-Based Multi-Agent Systems for Clinical Workflows: A Survey of {AI} Hospitals",
    author = "Yao, Zonghai  and
      Yu, Hong",
    editor = "Liakata, Maria  and
      Moreira, Viviane P.  and
      Zhang, Jiajun  and
      Jurgens, David",
    booktitle = "Proceedings of the 64th Annual Meeting of the {A}ssociation for {C}omputational {L}inguistics (Volume 1: Long Papers)",
    month = jul,
    year = "2026",
    address = "San Diego, California, United States",
    publisher = "Association for Computational Linguistics",
    url = "https://aclanthology.org/2026.acl-long.2123/",
    doi = "10.18653/v1/2026.acl-long.2123",
    pages = "45772--45793",
    ISBN = "979-8-89176-390-6"
}

@misc{alain2018understandingintermediatelayersusing,
      title={Understanding intermediate layers using linear classifier probes}, 
      author={Guillaume Alain and Yoshua Bengio},
      year={2018},
      eprint={1610.01644},
      archivePrefix={arXiv},
      primaryClass={stat.ML},
      url={https://arxiv.org/abs/1610.01644}, 
}

@inproceedings{
meng2022locating,
title={Locating and Editing Factual Associations in {GPT}},
author={Kevin Meng and David Bau and Alex J Andonian and Yonatan Belinkov},
booktitle={Advances in Neural Information Processing Systems},
editor={Alice H. Oh and Alekh Agarwal and Danielle Belgrave and Kyunghyun Cho},
year={2022},
url={https://openreview.net/forum?id=-h6WAS6eE4}
}
